\documentclass[acmsmall]{acmart}

\usepackage{multicol}
\usepackage{array}
\usepackage{multirow}
\usepackage{tabularx}
\usepackage{graphicx,rotating}
\usepackage{pdflscape}

\usepackage{diagbox}
\usepackage{slashbox}

\usepackage{indentfirst}

\newcommand{\blue}[1]{\color{black} #1}

\usepackage{ulem}
\usepackage{threeparttable}
\usepackage{amsmath}
\usepackage{amsthm}
\usepackage{mathtools}
\usepackage{tikz}
\usepackage{enumitem}

\DeclareMathOperator*{\argmax}{arg\,max}

\DeclareMathOperator*{\argmin}{arg\,min}

\newtheorem{remark}{Remark}

\newcommand*{\circled}[1]{\lower.7ex\hbox{\tikz\draw (0pt, 0pt)%
    circle (.5em) node {\makebox[1em][c]{\small #1}};}}
\robustify{\circled}

\AtBeginDocument{%
  \providecommand\BibTeX{{%
    \normalfont B\kern-0.5em{\scshape i\kern-0.25em b}\kern-0.8em\TeX}}}

\setcopyright{acmcopyright}
\copyrightyear{2018}
\acmYear{2018}

\acmJournal{JACM}
\acmVolume{37}
\acmNumber{4}
\acmArticle{111}
\acmMonth{8}

\begin{document}

\title{Grounding Foundation Models through Federated Transfer Learning: A General Framework}


\author{Yan Kang}
\email{yangkang@webank.com}
\affiliation{%
  \institution{WeBank}
  \country{China}
}

\author{Tao Fan}
\email{tfanac@cse.ust.hk}
\affiliation{%
  \institution{WeBank and Hong Kong University of Science and Technology}
  \country{China}
}

\author{Hanlin Gu}
\affiliation{%
  \institution{WeBank}
  \country{China}
}

\author{Xiaojin Zhang}
\affiliation{%
  \institution{Huazhong University of Science and Technology}
  \country{China}
}

\author{Lixin Fan}
\affiliation{%
  \institution{WeBank}
  \country{China}
}

\author{Qiang Yang}
\affiliation{%
  \institution{WeBank and Hong Kong University of Science and Technology}
  \country{China}
}

\renewcommand{\shortauthors}{Yan and Tao, et al.}

\begin{abstract}
Foundation Models (FMs) such as GPT-4 encoded with vast knowledge and powerful emergent abilities have achieved remarkable success in various natural language processing and computer vision tasks. Grounding FMs by adapting them to domain-specific tasks or augmenting them with domain-specific knowledge enables us to exploit the full potential of FMs. However, grounding FMs faces several challenges, stemming primarily from constrained computing resources, data privacy, model heterogeneity, and model ownership. Federated Transfer Learning (FTL), the combination of federated learning and transfer learning, provides promising solutions to address these challenges. Recently, the need for grounding FMs leveraging FTL, coined FTL-FM, has arisen strongly in both academia and industry. Motivated by the strong growth in FTL-FM research and the potential impact of FTL-FM on industrial applications, we propose an FTL-FM framework that formulates problems of grounding FMs in the federated learning setting, construct a detailed taxonomy based on the FTL-FM framework to categorize state-of-the-art FTL-FM works, and comprehensively overview FTL-FM works based on the proposed taxonomy. We also establish correspondence between FTL-FM and conventional phases of adapting FM so that FM practitioners can align their research works with FTL-FM. In addition, we overview advanced efficiency-improving and privacy-preserving techniques because efficiency and privacy are critical concerns in FTL-FM. Last, we discuss opportunities and future research directions of FTL-FM.
\end{abstract}





\maketitle

\section{Introduction}

In recent years, the field of artificial intelligence (AI) has undergone a significant transformation with the advent of Foundation Models (FMs)~\cite{bommasani2021fm}.  Advanced FMs, such as GPT-4~\cite{openai2023gpt4}, PaLM~\cite{chowdhery2022palm} and LLaMA ~\cite{touvron2023llama} boasting billions of parameters, have drawn considerable attention due to their remarkable performance in various AI tasks ranging from natural language processing, content generation to more complex tasks such as planning and reasoning.

While FMs have shown great success in various natural language processing~\cite{openai2023gpt4,liu2023pre,kamalloo2023evaluating} and computer vision tasks~\cite{kirillov2023segment,rombach2022high}, FMs have limitations that prevent them from being adopted to domain-specific applications: (1) FMs typically are pre-trained on publicly available datasets and lack industry-level and domain-specific knowledge, implying the deficiency of applying FMs to applications of various industrialized domains. (2) FMs are trained only up to a certain point in time, and thus, their knowledge is often outdated, thereby making it challenging to apply FMs to scenarios that require real-time updating of knowledge, e.g., in Newspaper publications. To address these limitations and unlock the full potential of FMs in addressing complex domain-specific tasks, it is imperative to integrate FMs with domain-specific knowledge in a timely manner.

\begin{figure*}[h!]
    \centering
    \includegraphics[width=0.45 \linewidth]{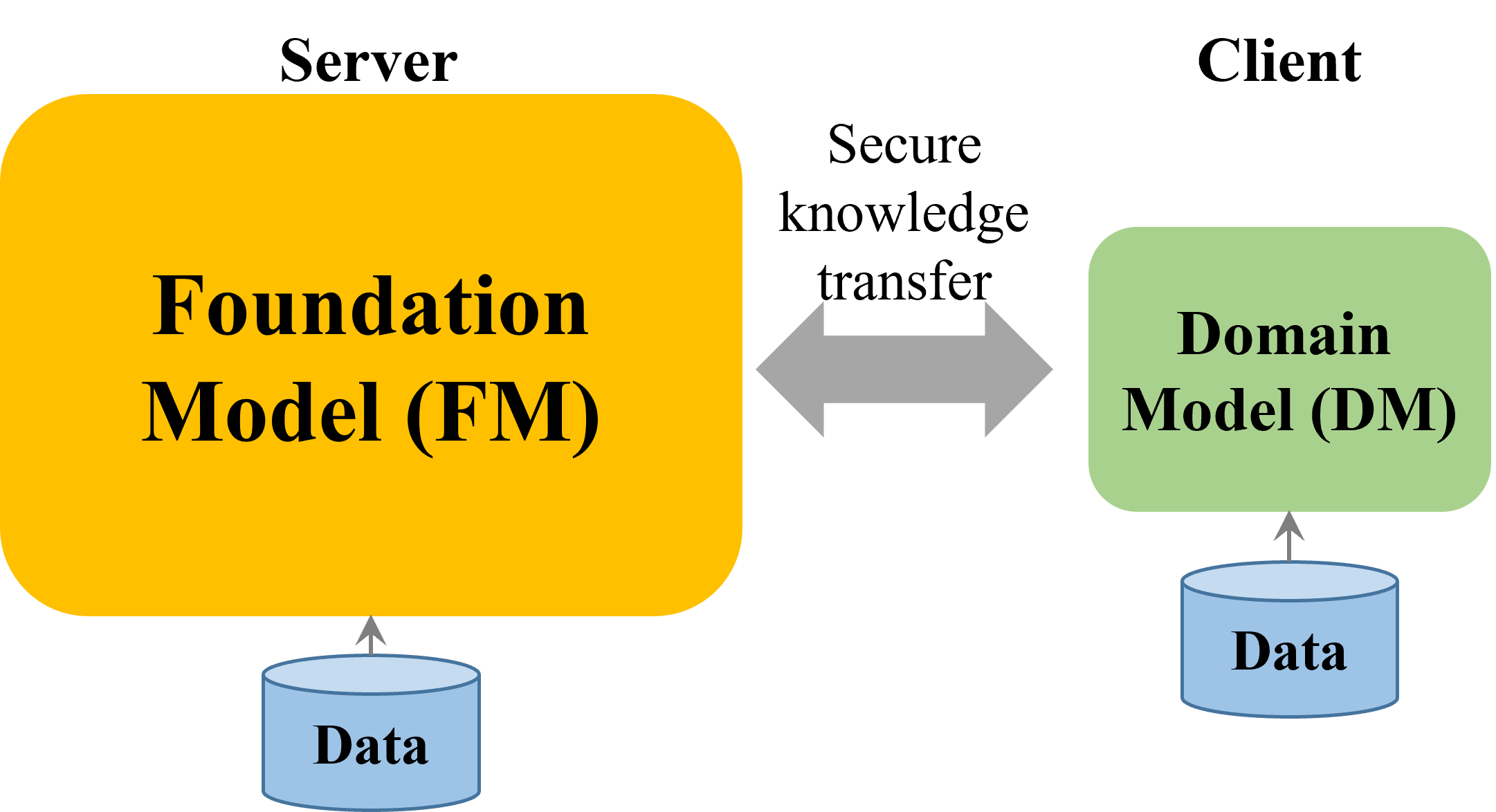}
    \caption{High-level illustration of knowledge transferring between a server hosting foundation model and a client with domain knowledge (in the form of domain models and data).}
    \label{fig:overview}
\end{figure*}

Nevertheless, some powerful FMs are closed-source (e.g., GPT-4 and PaLM), and valuable domain knowledge is often distributed across different enterprises, organizations, or edge devices, stored in their local data warehouses, or encoded in private machine learning models. Consequently, grounding FMs requires transferring knowledge between FM providers and domain knowledge holders. We refer to the former as the "server" and the latter as the "client," as depicted in Figure \ref{fig:overview}. 


The distributed nature of the foundation models and domain knowledge presents several challenges for grounding FMs:
\begin{itemize}
    \item \textbf{Constrained Resources}. FMs comprise a colossal amount of parameters and thus require massive amounts of computational resources to train, which prevents them from being deployed in companies or edge devices with constrained storage and computing resources. 
    \item \textbf{Data Privacy}. Clients with constrained computing resources (e.g., small-sized companies and edge devices) can send their data to the FM server (e.g., OpenAI) to fine-tune the FM for them. However, sharing private data with the FM server may compromise data privacy.
    \item \textbf{Model Ownership}. The FM server can send the full FM model to clients for further fine-tuning. However, directly sharing private FMs threatens the ownership of the FMs.
    \item \textbf{Model Heterogeneity}. The models held by the server and clients often have heterogeneous architectures and sizes. Thus, advanced knowledge transfer techniques are required to co-train models of all parties than the traditional FedAvg algorithm~\cite{mcmahan2017communication}.
\end{itemize}

Federated Transfer Learning (FTL) provides promising solutions to address these obstacles for grounding FMs as it provides servers of FMs and clients of DMs with privacy and security defense tools~\cite{li2023privacy}, federated learning algorithms~\cite{yang2019federated,mcmahan2017communication}, and transfer learning approaches~\cite{pan2010stf} to collaboratively adapt FMs to domain-specific models or augment FMs with domain-specific knowledge. We coin the concept of grounding FMs through FTL techniques as \textbf{FTL-FM}. In recent years, a considerable amount of methods across the spectrum of FTL-FM have been proposed to ground FMs. Subsequently, several surveys and vision papers were proposed to overview these works, summarize application scenarios, and discuss potential challenges and solutions~\cite{gou2021knowledge,ling2023beyond,zhu2023survey,zhuang2023foundation,chen2023federated}. However, these works either focus on specific techniques of customizing FMs without covering the technology landscape related to FTL-FM~\cite{gou2021knowledge,ling2023beyond,zhu2023survey} or they focus on federated learning but lack a comprehensive review of related FTL-FM approaches~\cite{zhuang2023foundation,chen2023federated}. Moreover, these works lack a unified framework that encompasses and formulates existing FTL-FM approaches.


{\blue{

This work fills these gaps by answering the following questions that are critical to ground foundation models when the foundation models and domain knowledge are distributed on different sites: (1) What are the typical settings in which foundation models interact with domain knowledge to ground foundation models? (2) What are the specific approaches for transferring knowledge across federated participants in grounding foundation models? (3) What are the privacy and efficiency considerations in federated transfer learning for grounding foundation models, and how are they addressed?

To answer these critical questions, we first propose a general FTL-FM framework that formulates FTL-FM settings, corresponding objectives, representative knowledge transfer approaches, and privacy measurements. According to this FTL-FM framework, we break down the aforementioned questions into concrete research issues, based on which we elaborate on state-of-the-art FTL-FM works. The main contributions of this work include:

}}

\begin{itemize}
    \item \textbf{A general FTL-FM framework of grounding FMs}. We first establish high-level correspondence between FTL-FM and conventional phases of adapting FM. We then propose an FTL-FM framework that formulates three settings and corresponding objectives of grounding FMs in federated learning (see Definition \ref{def:ftl-llm}). A broad range of state-of-the-art FTL-FM works can be reduced to these FTL-FM formulations, which serve as the basis for our proposed taxonomy of FTL-FM works.
    

    \item \textbf{A detailed taxonomy to categorize state-of-the-art FTL-FM works}. According to the FTL-FM framework, we raise five specific research issues, including what knowledge to transfer, how to transfer the knowledge, what the threat model is, what information to protect, and how to protect the information (see Section \ref{sec:category}). Based on these research issues, we construct a taxonomy to categorize state-of-the-art FTL-FM works (see Table \ref{tab:ftl-fm-tree}). The FTL-FM framework and the taxonomy can serve as guidance for future FTL-FM research.
    
    \item \textbf{A systematic overview of state-of-the-art FTL-FM works}. We systematically review state-of-the-art FTL-FM works based on the proposed taxonomy. We also discuss the relationship between reviewed FTL-FM works and the formulations defined in the FTL-FM framework. In addition, we overview advanced efficiency-improving and privacy-preserving methods since efficiency and privacy are critical concerns of FTL-FM. In this paper, our main focus is to review FTL-FM works that encompass large language models (LLM) with sizes no smaller than $\text{BERT}_{\text{base}}$. Nevertheless, we also take into account representative FTL-FM works that are dedicated to other types of models, such as large vision and speech models.
    \item \textbf{A discussion on future research directions.} Based on the comprehensive investigation of application scenarios, knowledge transfer methods, and privacy-preserving techniques adopted in existing FTL-FM works, we provide a detailed discussion of the open opportunities and future directions of FTL-FM research.
\end{itemize}

\begin{figure*}[h!]
    \centering
    \includegraphics[width=0.95\linewidth]{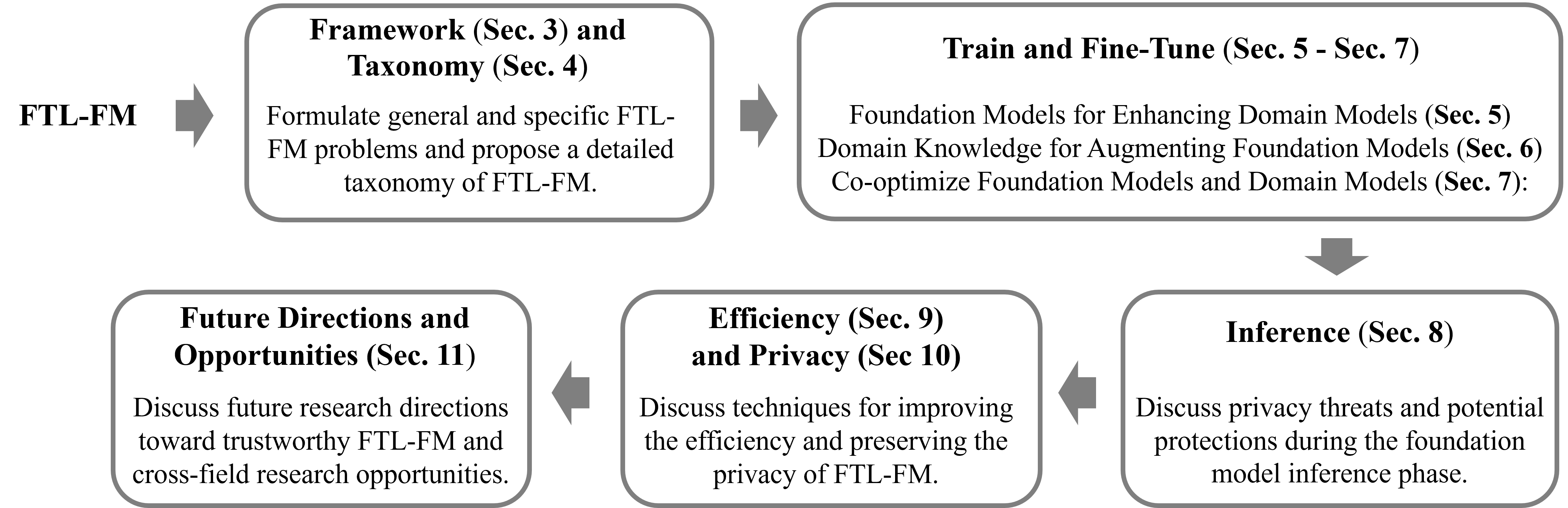}
    \caption{{\blue{Outline of core sections of this work.}}}
    \label{fig:outline}
\end{figure*}

The rest of the paper is organized as follows. Section \ref{sec:rel} overviews related works. Section \ref{sec:framework} defines a general framework of FTL-FM, including the formulations of three FTL-FM settings, general machine learning tasks, specific knowledge transfer approaches, and privacy leakage. Section \ref{sec:category} presents a taxonomy to categorize state-of-the-art FTL-FM works. Moving forward, Section \ref{sec:fm_to_dm}, Section \ref{sec:dm_to_fm}, and Section \ref{sec:fm_dm_coevolve} comprehensively review state-of-the-art works of the three FTL-FM settings, respectively. Section \ref{sec:infer} discusses privacy threats and protections during foundation model inference. Subsequently, Section \ref{sec:efficiency} and Section \ref{sec:privacy} examine efficiency-improving and privacy-preserving methods, respectively. Section \ref{sec:future} discusses opportunities and future directions. Finally, Section \ref{sec:conclusion} presents the concluding remarks for this work. The outline of the core sections of this work is pictorially illustrated in Figure \ref{fig:outline}.

\section{Related Work}\label{sec:rel}

In this section, we briefly review the existing survey and vision papers related to grounding foundation models (FMs). These papers either focused on techniques of compressing and customizing FMs without considering distributed scenarios of augmenting and adapting FMs~\cite{gou2021knowledge,ling2023beyond,zhu2023survey}, or broadly reviewed federated foundation models without formulating FTL-FM problems and comprehensively reviewing relevant FTL-FM works~\cite{zhuang2023foundation,chen2023federated}.

\citet{gou2021knowledge} comprehensively reviewed knowledge distillation techniques, which aim to reduce a large teacher model to a small student model without jeopardizing model performance substantially, facilitating the deployment of distilled but powerful deep models to resource-constrained entities, e.g., mobile phones and edge devices. More specifically, they reviewed knowledge distillation techniques from a wide range of perspectives, including knowledge categories, training schemes, teacher-student architecture, distillation algorithms, performance comparison, and applications. ~\citet{zhu2023survey} comprehensively reviewed model compression techniques tailored to large language models (LLMs). They proposed a taxonomy of model compression methods for LLMs, including pruning, knowledge distillation quantization, and low-rank factorization. In addition, They exhaustively examined metrics and benchmarks for evaluating compressed LLMs. While LLMs achieved remarkable performance across various natural language processing and understanding tasks, they may hallucinate because, among other reasons, they lack domain-specific knowledge. Therefore, domain specification has become a pivotal research area for customizing LLMs. \citet{ling2023beyond} proposed a systematic review, categorization, and taxonomy of LLM domain specification techniques, including external augmentation, prompt crafting, and model fine-tuning. 

General-purpose FMs typically lack domain knowledge. Enhancing FMs with industry-level or domain-specific knowledge requires a large amount of domain data that are often dispersed among multiple private entities (e.g., corporations). ~\citet{chen2023federated} proposed a concept of federated large language models (LLMs) and presented a framework for training LLMs in a federated manner. Technically, they focused on the federated learning scenarios in which each client owns a local LLM, and all clients collaboratively pre-train, fine-tune, or prompt engineer their LLMs. ~\citet{zhuang2023foundation} provided an overview of the combination of FMs and FL. In addition to discussing the motivation, challenges, and opportunities of federated FMs, they reviewed the motivations, challenges, opportunities, and future directions of leveraging FMs to facilitate FL.

\section{FTL-FM Framework}\label{sec:framework}


In this section, we propose a general framework of grounding foundation models (FMs) through federated transfer learning (FTL), which we term FTL-FM. Before delving into the formal definition of FTL-FM, we first establish the correspondence between phases of FTL-FM and those of adapting FMs and between FTL-FM and trustworthy federated learning.



\begin{figure*}[h!]
    \centering
    \includegraphics[width=0.98\linewidth]{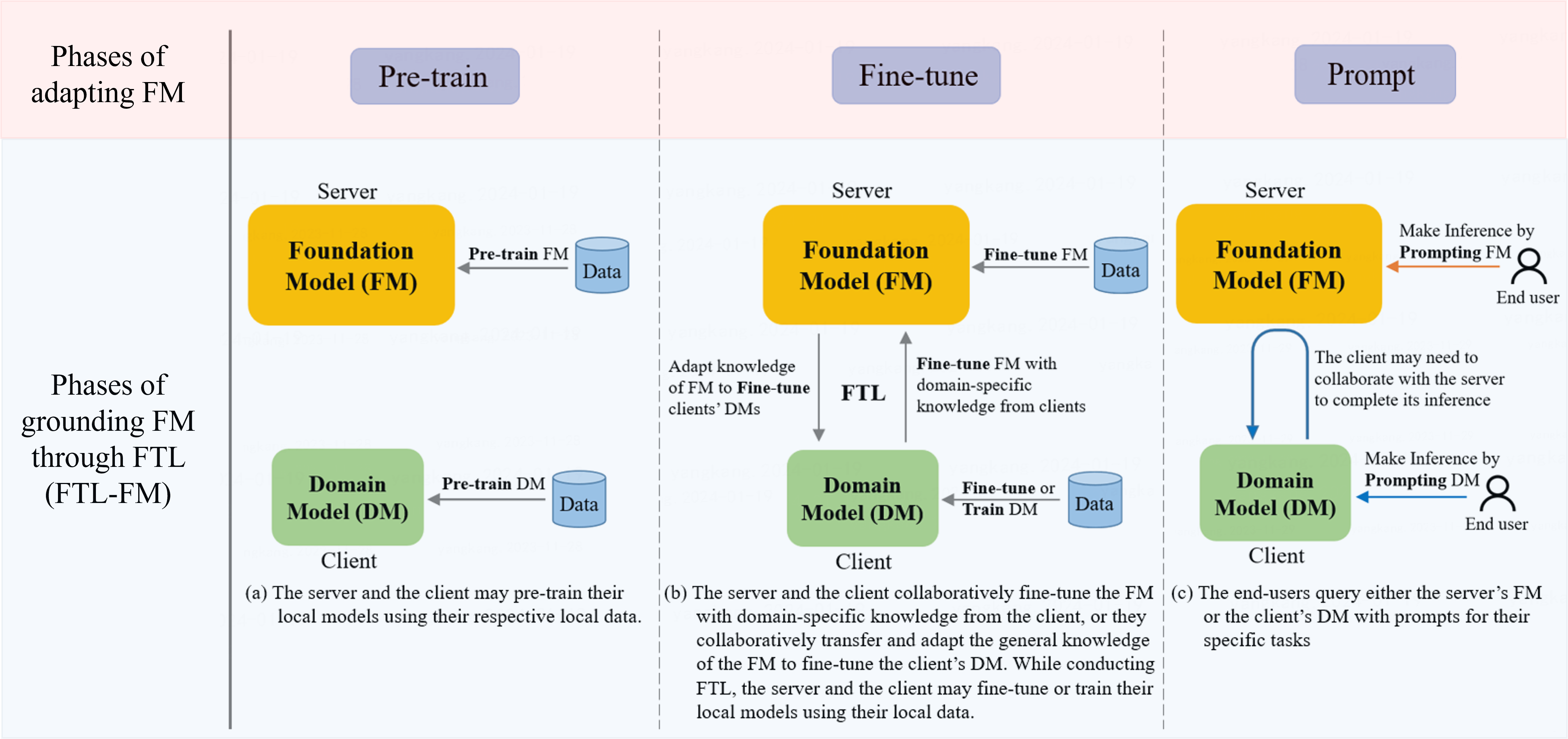}
    \caption{The correspondence between phases of FTL-FM and those of adapting FMs. We illustrate this correspondence with one server and one client. However, multiple clients can be involved in FTL-FM.}
    \label{fig:rel_tfl_fm}
\end{figure*}

Adapting an FM typically comprises three phases: pre-training, fine-tuning, and prompting. Pre-training aims to train an FM from scratch based on broad ranges of large-scale datasets so that the FM can be adaptable to a wide range of downstream tasks. Fine-tuning aims to train a pre-trained FM with specific goals. It can be instruction tuning that enables an FM to follow human instructions for solving specific tasks~\cite{ouyang2022rlhf}, alignment tuning that aligns an FM's behaviors with human intentions and values~\cite{ji2023aialign}, or domain-specific tuning that augments the FM with domain-specific knowledge. Prompting is to query an FM with prompts for solving various tasks. 

FTL-FM aims to adapt knowledge of the FM to enhance the client's domain model (DM) or transfer knowledge from the client to augment the FM through federated transfer learning. FTL-FM also involves three phases: pre-training, federated transfer learning, and inference. Herein, we establish correspondence between the phases of FTL-FM and those of adapting an FM, as illustrated in Figure \ref{fig:rel_tfl_fm}. This correspondence may help FM practitioners align their research works with FTL-FM.

\begin{itemize}

\item Pre-training: The server and the client may pre-train their local models (i.e., FM and DM) using their respective local data. Note that the client's DM can also be an FM.

\item Federated Transfer Learning (FTL): FTL involves two directions. (1) The server and the client collaboratively fine-tune the server's FM based on the client's domain-specific knowledge. (2) The server and the client collaboratively transfer and adapt general knowledge of the server's FM to fine-tune the client's DM. While conducting FTL, the server and the client may train or fine-tune their local models using their respective local data.

\begin{remark}
Because FTL-FM is typically based on foundation models that have already been pre-trained, this work focuses on the FTL phase that fine-tunes FMs and trains or fine-tunes DMs.
\end{remark}
\begin{remark}
The two knowledge transfer directions in FTL combined form a scenario where knowledge is transferred to optimize the server's FM and the client's DM simultaneously. This scenario is rarely explored in literature.
\end{remark}

\item Inference: Upon the completion of federated transfer learning, end users can query the FM of the server or the DM of the client to accomplish their specific tasks. Note that the client may need to collaborate with the server to complete the inference in some applications. 

\end{itemize}

\begin{figure*}[h!]
    \centering
    \includegraphics[width=0.95\linewidth]{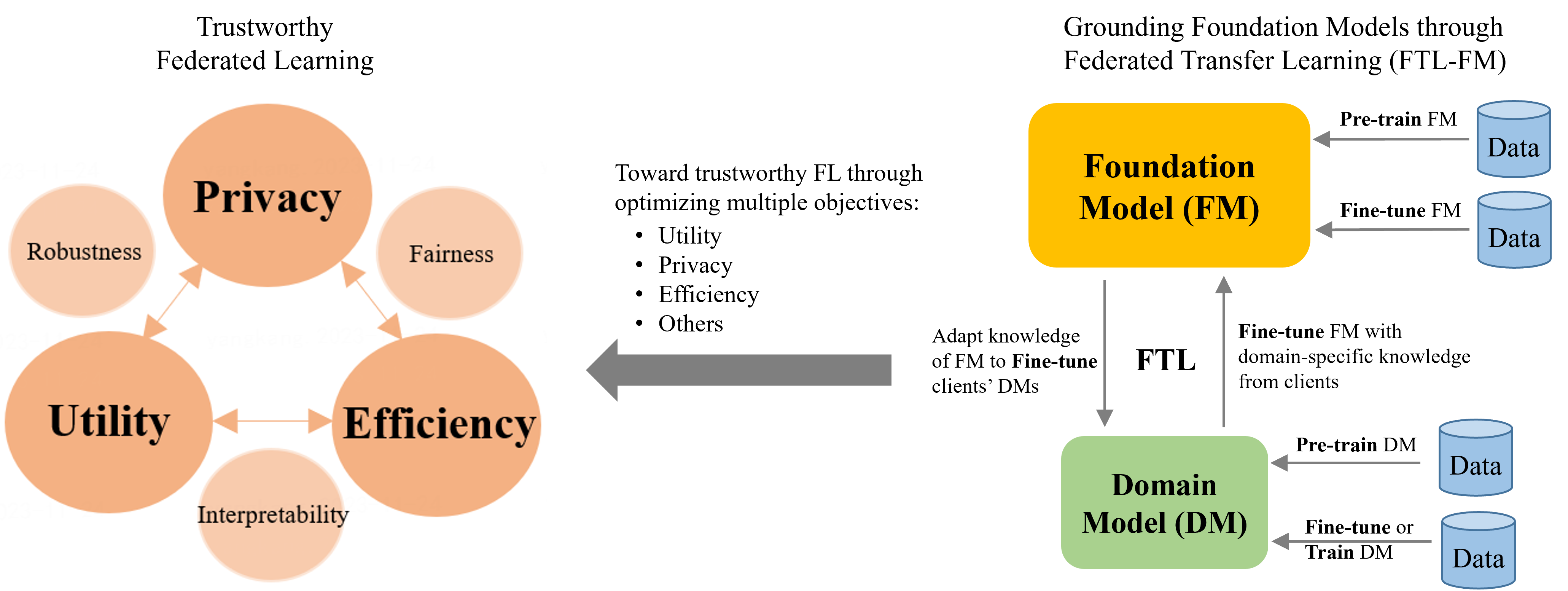}
    \caption{The correspondence between FTL-FM and trustworthy federated learning.}
    \label{fig:rel_tfl_fm_v2}
\end{figure*}

FTL-FM adheres to the principles of trustworthy federated learning (TFL)~\cite{kang2023cmofl}, as illustrated in Figure \ref{fig:rel_tfl_fm_v2}. To gain the trust of various stakeholders, such as FL participants, users, and regulators, an FTL-FM approach must simultaneously fulfill multiple objectives and optimize their trade-offs. Among these objectives, optimizing utility, efficiency, and privacy are the most critical ones to TFL. 


\subsection{The Definition of FTL-FM Framework}

We formally define the framework of grounding foundation models through federated transfer learning (FTL-FM) as follows.
\begin{figure*}[h!]
    \centering
    \includegraphics[width=0.99\linewidth]{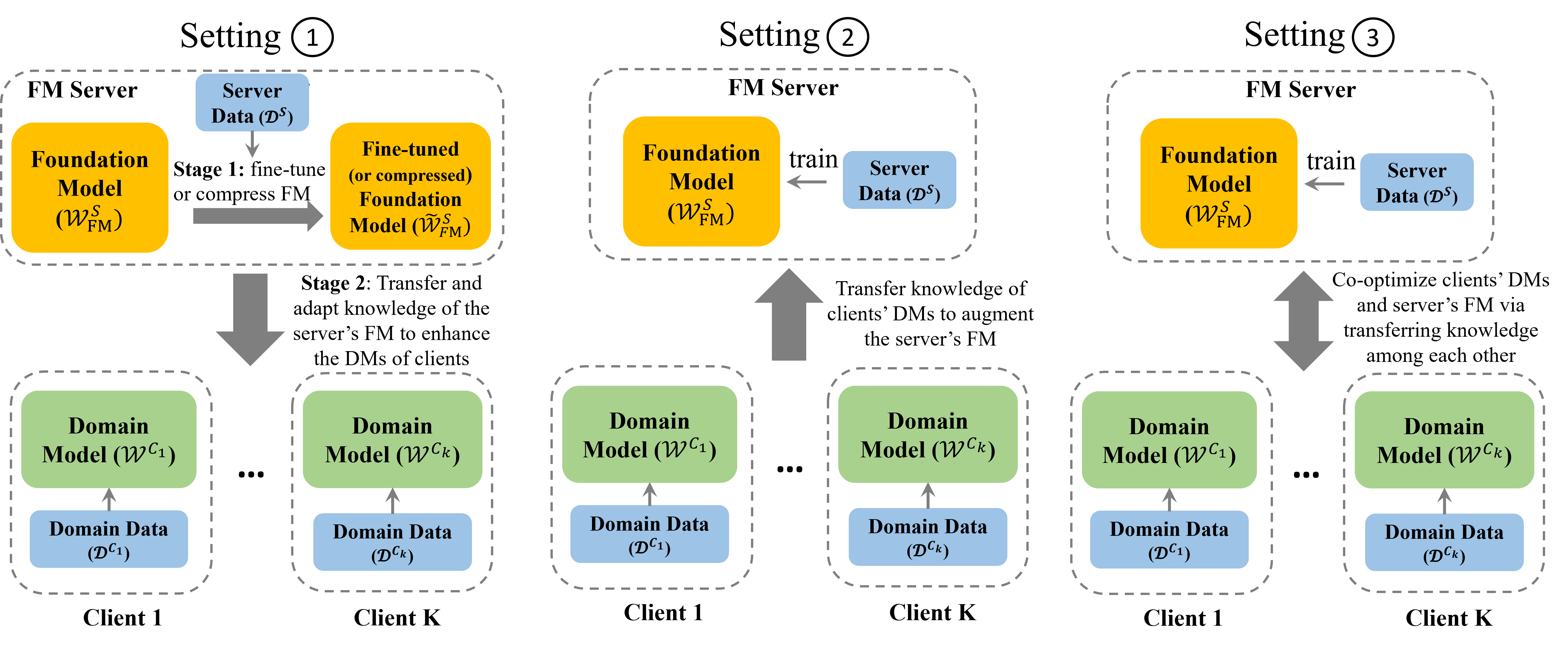}
    \caption{Illustration of the three settings formulated in the FTL-FM framework (see Definition \ref{def:ftl-llm}). The objective of setting \textcircled{1} is transferring and adapting knowledge of the server's foundation model (FM) to clients' domain models (DMs), the objective of setting \textcircled{2} is leveraging clients' domain-specific knowledge to enhance the server's FM, and the objective of setting \textcircled{3} is co-optimizing both the server's FM and clients' DMs.}
    \label{fig:overview_setting}
\end{figure*}

\begin{definition}[FTL-FM]\label{def:ftl-llm}
Given a server with a dataset $\mathcal{D}^S$ and a $\textit{foundation model}$ (FM) $\mathcal{W}^S_{\text{FM}}$ as well as $K$ clients that each client $k$ has a private dataset $\mathcal{D}^{C_k}$ and a domain model (DM) $\mathcal{W}^{C_k}$. $\mathcal{D}^S$ and $\mathcal{W}^S$ can be either public or private. 
FTL-FM can be categorized into three settings, where each setting aims to achieve a federated transfer learning objective through the collaboration between the server and clients (illustrated in Figure \ref{fig:overview_setting}):
\begin{itemize}
    \item \textbf{Setting $\circled{1}$:} Optimize clients' DMs $\mathcal{W}^{C_k}, k \in \{1,...,K\}$ by leveraging knowledge of $\mathcal{D}^S$ and $\mathcal{W}^S_{\text{FM}}$. This setting can involve two stages, which are formulated as follows. Note that stage 1 is not a prerequisite for stage 2.

    \textbf{Stage 1}: The server fine-tunes or compresses $\mathcal{W}^S_{\text{FM}}$ based on its data $\mathcal{D}^S$ while minimizing the distance between distribution $\mathcal{P}^S$ of $\mathcal{D}^S$ and distribution $\mathcal{P}^C$ of clients' data and mitigating the private leakage $\epsilon_p$ under a privacy constraint $\xi$:
    \begin{equation}\label{eq:setting_1_stage_1}
        \begin{split}
            & \min_{\mathcal{W}^{S}_{\text{FM}}} \ell^{S}(\mathcal{W}^S_{\text{FM}}) + \lambda \mathcal{M}(\mathcal{P}^S, \mathcal{P}^C) + \eta \epsilon_{p},\\
            & \text{ subject to} \quad \epsilon_{p}\le\xi,\\
            & \text{ where }\,\, \ell^{S}(\mathcal{W}^S_{\text{FM}}) = \mathbb{E}_{d \sim \mathcal{D}^{S}}[\ell^{S}(\mathcal{W}^S_{\text{FM}}; d)].
        \end{split}
    \end{equation}
    in which $\ell^{S}(\mathcal{W}^S_{\text{FM}})$ is the expected loss over dataset $\mathcal{D}^{S}$ for fine-tuning or compressing $\mathcal{W}^{S}_{\text{FM}}$; $\mathcal{M}$ is a distribution distance metric; $\lambda$ and $\eta$ are preferences toward the distribution distance and private leakage, respectively. The resulting model $\widetilde{\mathcal{W}}^S_{\text{FM}}$ fine-tuned or compressed by Eq.(\ref{eq:setting_1_stage_1}) is then used to facilitate clients to train their DMs through federated learning or local training. 
    
    \textbf{Stage 2}: Transfer and adapt the knowledge of $\mathcal{W}^S_{\text{FM}}$ (or $\widetilde{\mathcal{W}}^S_{\text{FM}}$ from stage 1) to help train clients' DMs for their specific tasks while mitigating the private leakage $\epsilon_p$ under a privacy constraint $\xi$.  
    \begin{equation}\label{eq:setting_1_stage_2}
        \begin{split}
            &\min_{\mathcal{W}^{C_1}, \dots ,\mathcal{W}^{C_K}} \mathcal{L}(\mathcal{W}^S_{\text{FM}},\mathcal{W}^{C_1}, \dots ,\mathcal{W}^{C_K}) = \sum^K_{k=1} p^{C_k} \ell^{C_k}(\mathcal{W}^S_{\text{FM}}, \mathcal{W}^{C_k}) + \eta \epsilon_{p},\\
            & \text{ subject to} \quad \epsilon_{p}\le\xi,\\
            & \text{ where }\,\, \ell^{C_k}(\mathcal{W}^S_{\text{FM}}, \mathcal{W}^{C_k}) = \mathbb{E}_{d \sim \mathcal{D}^{C_k}}[\ell^{C_k}(\mathcal{W}^S_{\text{FM}}, \mathcal{W}^{C_k}; d)], k \in \{1,...,K\}
        \end{split}
    \end{equation}
     in which $\ell^{C_k}(\mathcal{W}^S_{\text{FM}}, \mathcal{W}^{C_k})$ is the expected loss over the dataset $\mathcal{D}^{C_k}$, aiming to transfer and adapt the general knowledge and abilities of $\mathcal{W}^S_{\text{FM}}$ to client $k$'s DM $\mathcal{W}^{C_k}$ for a domain-specific task; $p^{C_k}$ and $\eta$ are preferences toward the loss of client $C_k$ and privacy leakage, respectively. 

    \item \textbf{Setting $\circled{2}$:} Optimize the server's FM $\mathcal{W}^S_{\text{FM}}$ by leveraging clients' domain-specific knowledge of $\mathcal{D}^{C_k}$ and $\mathcal{W}^{C_k}$$, k \in \{1,...,K\}$ while mitigating the private leakage $\epsilon_p$ under a privacy constraint $\xi$. The main objective is formulated as follows:
    \begin{equation}\label{eq:setting_2}
        \begin{split}
            &\min_{\mathcal{W}^{S}_{\text{FM}}} \mathcal{L}(\mathcal{W}^S_{\text{FM}},\mathcal{W}^{C_1}, \dots ,\mathcal{W}^{C_K}) = \sum^K_{k=1} p^{C_k} \ell^{C_k}(\mathcal{W}^S_{\text{FM}}, \mathcal{W}^{C_k}) + p^S \ell^{S}(\mathcal{W}^S_{\text{FM}}) + \eta \epsilon_{p},\\
            & \text{ subject to} \quad \epsilon_{p}\le\xi,\\
            & \text{ where }\,\, \ell^{C_k}(\mathcal{W}^S_{\text{FM}}, \mathcal{W}^{C_k}) = \mathbb{E}_{d \sim \mathcal{D}^{C_k}}[\ell^{C_k}(\mathcal{W}^S_{\text{FM}}, \mathcal{W}^{C_k}; d)], k \in \{1,...,K\} \text{ and optionally}\\
            & \quad\quad\quad\, \ell^{S}(\mathcal{W}^S_{\text{FM}}) = \mathbb{E}_{d \sim \mathcal{D}^{S}}[\ell^{S}(\mathcal{W}^S_{\text{FM}}; d)]
        \end{split}
    \end{equation}

    in which $\ell^{C_k}(\mathcal{W}^S_{\text{FM}}, \mathcal{W}^{C_k})$ is the expected loss over the dataset $\mathcal{D}^{C_k}$, aiming to transfer domain-specific knowledge of $\mathcal{W}^{C_k}$ to enhance $\mathcal{W}^S_{\text{FM}}$; $\ell^{S}(\mathcal{W}^S_{\text{FM}})$ is the loss over dataset $\mathcal{D}^{S}$, aiming to train $\mathcal{W}^S_{\text{FM}}$ for the server's task; $p^{S}$, $p^{C_k}$ and $\eta$ are preferences toward the loss of the server, loss of client $C_k$ and privacy leakage, respectively.
       
    \item \textbf{Setting $\circled{3}$:} Co-optimize server's FM $\mathcal{W}^S$ and clients' DMs $\mathcal{W}^{C_k}, k \in \{1,...,K\}$ by leveraging the knowledge of each other while mitigating the private leakage $\epsilon_p$ under a privacy constraint $\xi$. The main objective is formulated as follows:
    \begin{equation}\label{eq:setting_3}
        \begin{split}
            &\min_{\mathcal{W}^{S}_{\text{FM}},\mathcal{W}^{C_1}, \dots ,\mathcal{W}^{C_K}} \mathcal{L}(\mathcal{W}^S_{\text{FM}},\mathcal{W}^{C_1}, \dots ,\mathcal{W}^{C_K}) = \sum^K_{k=1} p^{C_k} \ell^{C_k}(\mathcal{W}^S_{\text{FM}}, \mathcal{W}^{C_k}) + p^S \ell^{S}(\mathcal{W}^S_{\text{FM}}) + \eta \epsilon_{p},\\
            &\quad \text{subject to} \quad \epsilon_{p}\le\xi,\\
            &\quad \text{where }\,\, \ell^{C_k}(\mathcal{W}^S_{\text{FM}}, \mathcal{W}^{C_k}) = \mathbb{E}_{d \sim \mathcal{D}^{C_k}}[\ell^{C_k}(\mathcal{W}^S_{\text{FM}}, \mathcal{W}^{C_k}; d)], k \in \{1,...,K\} \text{ and optionally}\\
            & \quad\quad\quad\quad\, \ell^{S}(\mathcal{W}^S_{\text{FM}}) = \mathbb{E}_{d \sim \mathcal{D}^{S}}[\ell^{S}(\mathcal{W}^S_{\text{FM}}; d)]
        \end{split}
    \end{equation}
    in which $\ell^{C_k}(\mathcal{W}^S_{\text{FM}}, \mathcal{W}^{C_k})$ is the expected loss over the dataset $\mathcal{D}^{C_k}$, aiming to transfer knowledge between server and clients to optimize $\mathcal{W}^S_{\text{FM}}$ and  $\mathcal{W}^{C_k}$; $\ell^{S}(\mathcal{W}^S_{\text{FM}})$ is the loss over dataset $\mathcal{D}^{S}$, aiming to train $\mathcal{W}^S_{\text{FM}}$ for the server's task; $p^{S}$, $p^{C_k}$ and $\eta$ are preferences toward the loss of the server, loss of client $C_k$ and privacy leakage, respectively. 
\end{itemize}

During the federated transfer learning, the following requirements should be satisfied: (1) The private data and models of the server and clients should not be passed to each other in plain text; (2) the intermediate training and inference results passed among FL parties should be protected; (3) the efficiency of training and inference should be guaranteed.
\end{definition}
\begin{remark}
Objectives of settings \textcircled{1} and \textcircled{2} aim to optimize clients' DMs $\mathcal{W}^{C_k},k=1,\dots,K$ and the server's FM $\mathcal{W}^{S}_{\text{FM}}$, respectively. Toward this end, other models may be trained as part of the optimization process. For instance, Objective \textcircled{1} may entail training $\mathcal{W}^{S}_{\text{FM}}$, while objective \textcircled{2} often involve training $\mathcal{W}^{C_k},k=1,\dots,K$. For the sake of clarity, these factors are not included in the formulation.
\end{remark}
\begin{remark}
A special case of the objective formulated in Eq.(\ref{eq:setting_3}) is that the server provides no model and acts as an aggregation function. The DMs of clients $1,...,K$ are initialized with FMs. Thus, the co-optimizing FMs and DMs is reduced to horizontal federated learning~\cite{mcmahan2017communication,tan2022towards} where clients collaboratively train a global domain-specific foundation model $\mathcal{W}^{G}_{\text{FM}}$. 
    \begin{equation}\label{eq:setting_3_variant_0}
        \begin{split}
        &\min_{\mathcal{W}^{G}_{\text{FM}}} \mathcal{L}(\mathcal{W}^G_{\text{FM}}) = \sum^K_{k=1} p^{C_k} \ell^{C_k}(\mathcal{W}^G_{\text{FM}}) + \eta \epsilon_{p},\\
            & \text{subject to} \quad \epsilon_{p}\le\xi,\\
            & \text{where }\,\, \ell^{C_k}(\mathcal{W}^G_{\text{FM}}) = \mathbb{E}_{d \sim \mathcal{D}^{C_k}}[\ell^{C_k}(\mathcal{W}^G_{\text{FM}}; d)], k \in \{1,...,K\}
        \end{split}
    \end{equation}

However, since FMs are expensive to train or clients may initialize their DMs with different FMs, clients typically train a global proxy model or a global Parameter-Efficient Fine-Tuning (PEFT) module $\mathcal{W}^G_{\text{a}}$ (e.g., adapter and LoRA) by minimizing their local loss $\ell^{C_k}(\mathcal{W}^G_{\text{a}}, \mathcal{W}^{C_k}_{\text{FM}})$ while keeping their local FMs $\mathcal{W}^{C_k}_{\text{FM}}$ frozen.
    \begin{equation}\label{eq:setting_3_variant}
        \begin{split}
        &\min_{\mathcal{W}^{G}_{\text{a}}} \mathcal{L}(\mathcal{W}^G_{\text{a}},\mathcal{W}^{C_1}_{\text{FM}}, \dots ,\mathcal{W}^{C_K}_{\text{FM}}) = \sum^K_{k=1} p^{C_k} \ell^{C_k}(\mathcal{W}^G_{\text{a}}, \mathcal{W}^{C_k}_{\text{FM}}) + \eta \epsilon_{p},\\
            & \text{subject to} \quad \epsilon_{p}\le\xi,\\
            & \text{where }\,\, \ell^{C_k}(\mathcal{W}^G_{\text{a}}, \mathcal{W}^{C_k}_{\text{FM}}) = \mathbb{E}_{d \sim \mathcal{D}^{C_k}}[\ell^{C_k}(\mathcal{W}^G_{\text{a}}, \mathcal{W}^{C_k}_{\text{FM}}; d)], k \in \{1,...,K\}
        \end{split}
    \end{equation}
\end{remark}
\begin{remark}
While formulations (i.e., Eqs.(\ref{eq:setting_1_stage_2}),(\ref{eq:setting_2}), (\ref{eq:setting_3}), and (\ref{eq:setting_3_variant})) are derived from horizontal federated learning, they can be generalized to vertical federated learning (VFL)~\cite{yang2019federated,liu2022vertical} by replacing the weighted average aggregation function with aggregation approaches dedicated to VFL scenarios.
\end{remark}

Definition ~\ref{def:ftl-llm} defines a general FTL-FM framework that formulates three settings of grounding FMs through FTL. These formulations reconcile a broad range of machine learning tasks, knowledge transfer approaches, and privacy leakage measures. Next, we provide detailed formulations based on existing works to illustrate each aspect. 

\subsection{Formulations of Machine Learning Tasks}

Federated transfer learning can be applied to address any machine learning tasks. Herein, we provide auto-regressive Language Modeling (LM)~\cite{radford2018gpt}, classification, and regression, which are widely adopted in real-world applications, as examples.
\begin{equation*}
        \begin{aligned}
            \textit{Auto-regressive LM:}\quad & \ell^{C_k}(\mathcal{W}^S_{\text{FM}}, \mathcal{W}^{C_k}; \mathcal{D}^{C_k}) = -\sum_{i=1}^{N} \sum_{j=1}^{|x|}\text{log}(\mathbb{P}(x_{i,j}|x_{i,<j}) \\ & \text{ where } \mathbb{P}(x_{i,j}|x_{i,<j})=h_{\mathcal{W}^S_{\text{FM}},\mathcal{W}^{C_k}}(x_i), \\
            \textit{Classification:}\quad & \ell^{C_k}(\mathcal{W}^S_{\text{FM}}, \mathcal{W}^{C_k}; \mathcal{D}^{C_k}) = - \sum_{i=1}^{N}\sum_{j=1}^{M}y_{i,j}\text{log}(p_{i,j}) \text{ where } p_i=h_{\mathcal{W}^S_{\text{FM}},\mathcal{W}^{C_k}}(x_i),\\
            \textit{Regression:}\quad & \ell^{C_k}(\mathcal{W}^S_{\text{FM}}, \mathcal{W}^{C_k}; \mathcal{D}^{C_k}) = \sum_{i=1}^{N}(y_i - p_i)^2 \text{ where } p_i=h_{\mathcal{W}^S_{\text{FM}},\mathcal{W}^{C_k}}(x_i).
        \end{aligned}
\end{equation*}
where $M$ is the number of classes, $\mathcal{D}^{C_k} = \{x_i,y_i\}_{i=1}^N$ is the dataset of client $k$, a sentence $x_i=\{x_{i,0},x_{i,1},...,x_{i,|x_i|-1}\}$, and $y_i$ is the ground truth label; $h_{\mathcal{W}^S_{\text{FM}},\mathcal{W}^{C_k}}(\cdot)$ parameterized by $\mathcal{W}^S_{\text{FM}}$ and $\mathcal{W}^{C_k}$ denotes the general federated transfer learning procedure between the server and client $k$, and $p_i$ is the prediction of $h_{\mathcal{W}^S_{\text{FM}},\mathcal{W}^{C_k}}$ given $x_i$.

\subsection{Formulations of Knowledge Transfer Approaches.}

Various knowledge transfer approaches have been proposed to achieve each of the three objectives defined in the FTL-FM framework (see Definition \ref{def:ftl-llm}). In this section, we review and formulate representative approaches. These formulations are summarized in Table \ref{tab:summary-ftl-fm-kt-loss}. To better illustrate these formulations, we denote the FM hosted by the server as $f_{\mathcal{W}^S_{\text{FM}}}(\cdot)$ parameterized by $\mathcal{W}^S_{\text{FM}}$, the DM owned by client $k$ as $g_{\mathcal{W}^{C_k}}(\cdot)$ parameterized by $\mathcal{W}^{C_k}$, the data owned by the server as $\mathcal{D}^S=\{x^{S},y^{S}\}$, and the data owned by client $k$ as $\mathcal{D}^k=\{x^{C_k},y^{C_k}\}$.

\begin{table*}[!h]
 	\centering
	\footnotesize
	\caption{An overview of formulations of representative federated knowledge transfer approaches.} 
\bgroup
\def\arraystretch{1.3}
  \setlength{\tabcolsep}{1.3pt}
	\begin{tabular}{c||c|c|c}
	    \hline
		\multirow{2}{*}{Objective}  & \multirow{2}{*}{\shortstack{Formulation of Specific Federated Transfer\\Learning Loss $\ell^{C_k}$ of each client $k$}} & \multirow{2}{*}{\shortstack{Knowledge Transfer \\ Approach}} &  \multirow{2}{*}{\shortstack{Representative \\ Work}} \\
      ~  & ~  &  ~  \\
	    \hline
	    \hline
        \multirow{14}{*}{\shortstack{Optimize \\ Domain Models \\ $\mathcal{W}^{C_k}$ $k=1,\dots,K$}} & \multirow{4}{*}{\shortstack{$\ell_{\text{TA}}^{C_k}(g_{\mathcal{W}^{C_k}}(x^{C_k}), y^{C_k}) \text{ where } $ \\ $\mathcal{W}^{C_k} = \argmin_{\mathcal{W}^{S}_{\text{SM}}} \ell^{S}_{\text{KD}}(f'_{\mathcal{W}_{\text{SM}}^S}(x^S), f_{\mathcal{W}^S_{\text{FM}}}(x^S))$ \\ + $\lambda \mathcal{M}(\mathcal{P}^S, \mathcal{P}^C)$}} & \multirow{4}{*}{domain adaptation} & \multirow{4}{*}{\cite{wang2023can,hou2023FreD}} \\
       ~  & ~  &  ~  \\
       ~  & ~  &  ~  \\
       ~  & ~  &  ~  \\
        \cline{2-4}
        ~ & \multirow{3}{*}{\shortstack{$\ell_{\text{TA}}^{C_k}(f_{\mathcal{W}^S_{\text{FM}}}([x^{C_k},g_{\mathcal{W}^{C_k}}(x^{C_k})]), y^{C_k})$\\where $\mathcal{W}^{C_k}$ can be parameters of prompt generator \\or discrete prompts.}} & \multirow{3}{*}{\shortstack{federated \\prompt optimization}} & \multirow{3}{*}{\shortstack{\cite{sun2022bbt,diao2023black,li2023dsp}\\\cite{lin2023fedbbpt,zhou2023ape,sordoni2023dln}\\\cite{guo2023evoprompt}}} \\
        ~  & ~  &  ~  \\
        ~  & ~  &  ~  \\
        \cline{2-4}
        ~ & \multirow{4}{*}{\shortstack{$\ell_{\text{TA}}^{C_k}(h_{\mathcal{W}^{\diamond}_\text{HE}} \circ f_{\mathcal{W}^S_{\text{BA}}} \circ g_{\mathcal{W}^{C_k}_\text{BO}}(x^{C_k}), y^{C_k})$\\where $h_{\mathcal{W}^{\diamond}_\text{HE}},\diamond \in \{S, C_k\}$, $f_{\mathcal{W}^S_{\text{BA}}}$, and $g_{\mathcal{W}^{C_k}_\text{BO}}$ are the head, \\backbone, and bottom models split from the FM $f_{\mathcal{W}^S_{\text{FM}}}$.}} &  \multirow{4}{*}{\shortstack{federated \\split learning}} & \multirow{4}{*}{\shortstack{~\cite{zhou2023textobfuscator,shen2023sap,li2023privacy}\\~\cite{xu2023shuffled,Tian2022fedbert}}}\\
        ~  & ~  & ~ \\
        ~  & ~  & ~ \\
        ~  & ~  & ~ \\
        \cline{2-4}
        ~  & \multirow{4}{*}{\shortstack{ $\ell_{\text{TA}}^{C_k}(g_{\mathcal{W}^{C_k}}(x^{C_k}), y^{C_k}) + \ell_{\text{KD}}^{C_k}(g_{\mathcal{W}^{C_k}}(z^{C_k}); z^S)$ \\$\text{ where } z^S = f_{\mathcal{W}^S_{\text{FM}}}(z^{C_k})$ and $z^{C_k}$ is data used by client $k$ \\to extract knowledge from $f_{\mathcal{W}^S_{\text{FM}}}$}}  &  \multirow{4}{*}{\shortstack{federated \\ knowledge distillation }} & \multirow{4}{*}{\cite{ho2023large,hsieh2023distilling,he2020fedgkt}}\\
        ~  & ~  &  ~  \\
        ~  & ~  &  ~  \\
        ~  & ~  &  ~  \\
	\hline
     \multirow{8}{*}{\shortstack{Optimize \\Foundation Model \\ $\mathcal{W}^S_{\text{FM}}$}} & \multirow{4}{*}{\shortstack{$\ell_{\text{TA}}^{C_k}(\phi_{\mathcal{W}^S_{a}} \circ \tilde{f}_{\widetilde{\mathcal{W}}^S_{\text{FM}}} \circ \psi_{\mathcal{W}^S_{b}}(x^{C_k}), y^{C_k})$ \\ where $\phi_{\mathcal{W}^S_{a}}$ and $\psi_{\mathcal{W}^S_{b}}$ are adapters selected from $f_{\mathcal{W}^S_{\text{FM}}}$, \\and $\tilde{f}_{\widetilde{\mathcal{W}}^S_{\text{FM}}}$ is emulator compressed from $f_{\mathcal{W}^S_{\text{FM}}}$.}} & \multirow{4}{*}{\shortstack{federated \\offsite tuning}}  & \multirow{4}{*}{\shortstack{\cite{xiao2023offsite,fan2023fate,fedost}\\\cite{chua2023fedpeat,khalid2023cefhri}}} \\
    ~  & ~  &  ~  \\
    ~  & ~  &  ~  \\
    ~  & ~  &  ~  \\

        
      \cline{2-4}
      ~ & \multirow{3}{*}{\shortstack{$\ell^{S}_{\text{TA}}(f_{\mathcal{W}^S_{\text{FM}}}(x^S),y^S) + \ell_{\text{KD}}^{S}(f_{\mathcal{W}^S_{\text{FM}}}(z^{S}), z^{C_k})$ \\ where $z^{C_k} = g_{\mathcal{W}^{C_k}}(z^S)$ and $z^S$ is data used by server $S$ \\to extract knowledge from $g_{\mathcal{W}^{C_k}}$}}. & \multirow{3}{*}{\shortstack{federated \\ knowledge distillation}} & \multirow{3}{*}{\shortstack{\cite{yu2023multimodal}}}\\
      ~  & ~  &  ~  \\
      ~  & ~  &  ~  \\
	\hline
    
    \multirow{12}{*}{\shortstack{Optimize $\mathcal{W}^S_{\text{FM}}$ \\ and \\ $\mathcal{W}^{C_k}$, $k=1,\dots,K$ \\ (or $\mathcal{W}^G)$}} &  \multirow{3}{*}{\shortstack{$\ell_{\text{TA}}^{C_k}(g_{\mathcal{W}^{\diamond}}(x^{C_k}), y^{C_k}) \quad\quad\quad\quad\quad\quad\quad\quad$ \\$\quad\quad\quad\quad + \ell_{\text{FT}}(f_{\mathcal{W}_{\text{FM}}^{S}}(x^a), y^a) + \ell_{\text{FT}}(g_{\mathcal{W}^{\diamond}}(x^b), y^b)$ \\ where $\diamond \in \{G, C_k\}$, $\{x^a, y^a\}$ and $\{x^b, y^b\}$ are fine-tuning \\ data, and their specific forms are algorithm-dependent.}}  &  \multirow{4}{*}{\shortstack{federated \\co-optimization}} & \multirow{4}{*}{\shortstack{\cite{fan2023fate,deng2023crosslm}}} \\
    ~  & ~ & ~ \\
    ~  & ~ & ~ \\
    ~  & ~ & ~ \\
     \cline{2-4}
      ~  & \multirow{3}{*}{\shortstack{$\ell_{\text{TA}}^{C_k}(g_{\mathcal{W}^{G}_{\text{FM}}}(x^{C_k}), y^{C_k})$ \\ where $g_{\mathcal{W}^{G}_{\text{FM}}}$ is the global domain-specific foundation \\model shared by all clients.}} &  \multirow{3}{*}{\shortstack{federated full-model\\ training or fine-tuning}} & \multirow{3}{*}{\shortstack{~\cite{lin2022fednlp,woi2023fededge}}} \\
      ~  & ~  &  ~  \\
     ~  & ~  &  ~  \\
     \cline{2-4}
      ~  & \multirow{4}{*}{\shortstack{$\ell_{\text{TA}}^{C_k}(\phi_{\mathcal{W}^{G}_a} \oplus g_{\mathcal{W}^{C_k}_{\text{FM}}}(x^{C_k}), y^{C_k})$ \\ where $\phi_{\mathcal{W}^{G}_a}$ is the global PEFT module shared by all clients, \\and $g_{\mathcal{W}^{C_k}_{\text{FM}}}$ is client $k$'s FM and is frozen during training.}} &  \multirow{4}{*}{\shortstack{federated parameter-\\efficient fine-tuning}} & \multirow{4}{*}{\shortstack{~\cite{zhang2023fedpetuning,Zhao2023fedprompt}\\~\cite{cai2023fedadapter,zhang2023fedit}}} \\
      ~  & ~  &  ~  \\
      ~  & ~  &  ~  \\
       ~  & ~  &  ~  \\
	\hline
	\end{tabular}
 \egroup
\label{tab:summary-ftl-fm-kt-loss}
\end{table*}

\textbf{To achieve the objective of setting \textcircled{1}}, a straightforward way is through domain adaptation: the server compresses the FM into a relatively smaller model that is adapted to the client's task and sends the distilled model to the downstream client for further fine-tuning ~\cite{   mire2022dpkd,yu2022differentially,wang2023can}. We formulate this approach as follows.
    \begin{equation}
        \begin{aligned}
            \textit{domain adaptation:}\quad & \min_{\mathcal{W}^{C_k}} \ell_{\text{TA}}^{C_k}(g_{\mathcal{W}^{C_k}}(x^{C_k}), y^{C_k}),\\
            & \text{ where } \mathcal{W}^{C_k} = \argmin_{\mathcal{W}^{S}_{\text{SM}}} \ell^{S}_{\text{KD}}(f'_{\mathcal{W}_{\text{SM}}^S}(x^S), f_{\mathcal{W}^S_{\text{FM}}}(x^S)) + \lambda \mathcal{M}(\mathcal{P}^S, \mathcal{P}^C).
        \end{aligned}
    \end{equation}
where $\ell^S_{\text{KD}}$ is the knowledge distillation loss for distilling knowledge from the teacher model $f_{\mathcal{W}^S_{\text{FM}}}$ to the student model $f'_{\mathcal{W}_{\text{SM}}^S}$, which is used to initialize downstream client $k$'s domain model $g_{\mathcal{W}^{C_k}}$; $\ell_{\text{TA}}^{C_k}$ is the task loss to fine-tune local model $g_{\mathcal{W}^{C_k}}$ of client $k$.

As a case in point, ~\citet{wang2023can} proposed an approach that involves sampling data from a public dataset that closely resembles the distribution of private data owned by clients. These selected samples are utilized to distill an FM into an on-device language model. Then, clients employ these distilled models as initialization for subsequent federated training. ~\citet{hou2023FreD} proposed a similar approach, but they fine-tune an FM for clients instead of distillation.

In addition to model transfer, the server can adapt the general knowledge of the FM to the clients' domain models by transferring data-level and representation-level knowledge. Federated prompt tuning~\cite{lin2023fedbbpt,sun2023fedbpt}, federated split learning~\cite{li2023privacy,Tian2022fedbert} and federated knowledge distillation~\cite{he2020fedgkt} are three representative approaches. We formulate their objectives of transferring knowledge from the server's FM to the client $k$'s DM as follows:
\begin{align}
    \textit{federated prompt optimization:}\quad & \min_{\mathcal{W}^{C_k}} \ell_{\text{TA}}^{C_k}(f_{\mathcal{W}^S_{\text{FM}}}([x^{C_k},g_{\mathcal{W}^{C_k}}(x^{C_k})]), y^{C_k}), \label{eq:fedpt}\\
    \textit{federated split learning:}\quad & \min_{\mathcal{W}^{C_k}_{\text{HE}},\mathcal{W}^{C_k}_{\text{BO}}} \ell_{\text{TA}}^{C_k}(h_{\mathcal{W}^{\diamond}_\text{HE}} \circ f_{\mathcal{W}^S_{\text{BA}}} \circ g_{\mathcal{W}^{C_k}_\text{BO}}(x^{C_k}), y^{C_k}), \label{eq:fedsplit} \\ &
    \text{ where } \diamond \in \{S, C_k\}, \nonumber\\
    \textit{federated knowledge distillation:}\quad & \min_{\mathcal{W}^{C_k}} \ell_{\text{TA}}^{C_k}(g_{\mathcal{W}^{C_k}}(x^{C_k}), y^{C_k}) + \ell_{\text{KD}}^{C_k}(g_{\mathcal{W}^{C_k}}(z^{C_k}), z^S) \label{eq:fedkd_1} \\ & \text{ where } z^S = f_{\mathcal{W}^S_{\text{FM}}}(z^{C_k}). \nonumber
\end{align}
where $h_{\mathcal{W}^{\diamond}_\text{HE}}$, $f_{\mathcal{W}^S_{\text{BA}}}$, and $g_{\mathcal{W}^{C_k}_\text{BO}}$ are the head, backbone, and bottom models split from the FM $f_{\mathcal{W}^S_{\text{FM}}}$, respectively; [·, ·] stands for concatenation; $f \circ g$ is the function composition such that the outputs of the function $g$ are fed into the function $f$; $\ell_{\text{TA}}^{C_k}$ is the task loss for training client $k$'s DM  using local data $\{x^{C_k}, y^{C_k}\}$; $\ell_{\text{KD}}^{C_k}$ is the knowledge distillation loss for distilling FM's knowledge in the form of $z^S$ to client $k$'s DM based on $z^{C_k}$. The particular form of $z^{C_k}$ varies according to the specific application. It could potentially be representations or synthetic data generated by the local model of client $C_k$, or it may even be the local data of client $C_k$.


Federated prompt optimization can be prompt generator optimization or discrete prompt optimization. For the former~\cite{sun2022bbt,diao2023black,li2023dsp,lin2023fedbbpt}, a client $k$ trains its local prompt generator $g_{\mathcal{W}^{C_k}}$ based on its own local data and the responses made by $f_{\mathcal{W}^S_{\text{FM}}}$. For the latter~\cite{zhou2023ape,sordoni2023dln,guo2023evoprompt}, a client $k$ optimizes discrete prompts with the guidance of the FM. In federated split learning~\cite{zhou2023textobfuscator,shen2023sap,li2023privacy,xu2023shuffled,Tian2022fedbert}, the FM $f_{\mathcal{W}^S_{\text{FM}}}$ typically is split into a head model $h_{\mathcal{W}^{\diamond}_\text{HE}},\diamond \in \{S, C_k\}$, a backbone model $f_{\mathcal{W}^S_{\text{BA}}}$, and a bottom model $g_{\mathcal{W}^{C_k}_\text{BO}}$. The backbone model is deployed on the server side, and the bottom model is deployed on the client side, while the head model can be deployed on either side. The server and clients collaborate to train the dispersed FM. The core idea of federated knowledge distillation~\cite{ho2023large,hsieh2023distilling,he2020fedgkt} is to train each client's domain model $g_{\mathcal{W}^{C_k}}$ using knowledge $z^S$ generated by the FM $f_{\mathcal{W}^S_{\text{FM}}}$ and optionally client's local datasets $\mathcal{D}^{C_k}$.

\textbf{To achieve the objective of setting \textcircled{2}}, federated offsite tuning~\cite{xiao2023offsite,fan2023fate,fedost,chua2023fedpeat,khalid2023cefhri} and federated knowledge distillation~\cite{yu2023multimodal} are two representative approaches. We formulate their objectives as follows:
\begin{align}
    \textit{federated offsite tuning:}\quad &  \min_{\mathcal{W}^S_{a},\mathcal{W}^S_{b}} \ell_{\text{TA}}^{C_k}(\phi_{\mathcal{W}^S_{a}} \circ \tilde{f}_{\widetilde{\mathcal{W}}^S_{\text{FM}}} \circ \psi_{\mathcal{W}^S_{b}}(x^{C_k}), y^{C_k}), \label{eq:fedost}\\
    \textit{federated knowledge distillation:}\quad & \min_{\mathcal{W}^{S}_{\text{FM}}} \ell^{S}_{\text{TA}}(f_{\mathcal{W}^S_{\text{FM}}}(x^S),y^S) + \ell_{\text{KD}}^{S}(f_{\mathcal{W}^S_{\text{FM}}}(z^{S}), z^{C_k})\label{eq:fedkd_2}\\&\text{ where } z^{C_k} = g_{\mathcal{W}^{C_k}}(z^{S}).\nonumber
\end{align}
where $\phi_{\mathcal{W}^S_{a}}$ and $\psi_{\mathcal{W}^S_{b}}$ are the two adapters selected from $f_{\mathcal{W}^S_{\text{FM}}}$, and $\tilde{f}_{\widetilde{\mathcal{W}}^S_{\text{FM}}}$ is emulator compressed from $f_{\mathcal{W}^S_{\text{FM}}}$; $\ell_{\text{TA}}^{C_k}$ is the task loss for training $\phi_{\mathcal{W}^S_{a}}$ and $\psi_{\mathcal{W}^S_{b}}$ with the help of emulator $\tilde{f}_{\widetilde{\mathcal{W}}^S_{\text{FM}}}$; $\ell_{\text{TA}}^{S}$ is the task loss for fine-tuning the server's $f_{\mathcal{W}^S_{\text{FM}}}$ using local data $\{x^{S}, y^{S}\}$; $\ell_{\text{KD}}^{S}$ is the knowledge distillation loss for fine-tuning $f_{\mathcal{W}^S_{\text{FM}}}$ using representations $z^{C_k}$ generated by client $k$'s local DM $g_{\mathcal{W}^{C_k}}$ based on $z^S$. The particular form of $z^{S}$ varies according to the specific application. It could potentially be representations or synthetic data generated by the model of the server $S$, or it may be the local data of the server $S$.

In federated offsite tuning~\cite{fan2023fate,fedost,chua2023fedpeat}, the server first sends $\phi_{\mathcal{W}^S_{a}}$, $\psi_{\mathcal{W}^S_{b}}$, and $\tilde{f}_{\widetilde{\mathcal{W}}^S_{\text{FM}}}$ to all clients before training. Then, each client trains the adaptors $\phi_{\mathcal{W}^S_{a}}$ and $\psi_{\mathcal{W}^S_{b}}$ with the help of the emulator $\tilde{f}_{\widetilde{\mathcal{W}}^S_{\text{FM}}}$ based on its local data. Next, all clients send their adaptors back to the server, which in turn aggregates clients' adaptors and plugs the aggregated adaptors back to the FM. In federated knowledge distillation~\cite{yu2023multimodal}, the $f_{\mathcal{W}^S_{\text{FM}}}$ at the server is fine-tuned by the representations generated by each client's $g_{\mathcal{W}^{C_k}}$. In both approaches, $f_{\mathcal{W}^S_{\text{FM}}}$ can be optionally fine-tuned by the server's local data if available. 

\textbf{To achieve the objective of setting \textcircled{3}}, the server's FM and clients' DMs are co-optimized through sharing knowledge of each other. This specific co-optimization objective of each client $k$ can be formulated as follows. 
\begin{equation}\label{eq:fedcotuning}
\begin{aligned}
    \textit{federated co-optimization:}\quad \min_{\mathcal{W}^{\diamond},\mathcal{W}^{S}_{\text{FM}}} &\ell_{\text{TA}}^{C_k}(g_{\mathcal{W}^{\diamond}}(x^{C_k}), y^{C_k}) \\
    & + \ell_{\text{FT}}(f_{\mathcal{W}_{\text{FM}}^{S}}(x^a), y^a) + \ell_{\text{FT}}(g_{\mathcal{W}^{\diamond}}(x^b), y^b),\\ &
    \text{ where } \diamond \in \{G, C_k\}\\
\end{aligned}
\end{equation}
where $g_{\mathcal{W}^{G}}$ is the global model shared by all clients when $\diamond$ is $G$, and $g_{\mathcal{W}^{C_k}}$ is the personalized model of client $k$ when $\diamond$ is $C_k$; $\ell_{\text{TA}}^{C_k}$ is the task loss for training client $k$'s DM $g_{\mathcal{W}^{\diamond}},\diamond \in \{G, C_k\}$ based on local data $\{x^{C_k},y^{C_k}\}$; $\ell_{\text{FT}}$ is the loss for fine-tuning $f_{\mathcal{W}^S_{\text{FM}}}$ and $g_{\mathcal{W}^{\diamond}}$ using knowledge in the form of $\{x^a, y^a\}$ and $\{x^b, y^b\}$, respectively. The specific forms of $\{x^a, y^a\}$ and $\{x^b, y^b\}$ are algorithm-dependent.


The federated co-optimization can be implemented in various forms. \citet{fan2023fate} proposed FedCoLLM~\cite{fan2023fate} that accomplishes the co-optimization by distilling knowledge between the server's FM  $f_{\mathcal{W}^S_{\text{FM}}}$ and the global DM $g_{\mathcal{W}^{G}}$ aggregated from client's local DMs. In FedCoLLM, $x^a=x^b=x^S$, $y^a=g_{W^{G}}(x^S)$, and $y^b=f_{\mathcal{W}^{S}_{\text{FM}}}(x^S)$ where $x^S$ is public data. \citet{deng2023crosslm} proposed CrossLM that accomplishes the co-optimization of $f_{\mathcal{W}^{S}_{\text{FM}}}$ and $g_{\mathcal{W}^{C_k}}, k \in \{1,\dots,K\}$ based on a synthetic dataset generated by $f_{\mathcal{W}^{S}_{\text{FM}}}$. In CrossLM, $y^a=y^b=y^S$ and $x^a=x^b=f_{\mathcal{W}^{S}_{\text{FM}}}(\mathcal{F}(y^S))$ where $y^S$ is the given labels and $\mathcal{F}(y^S)$ is the corresponding label-descriptive prompts.

A special case of objective \textcircled{3} is that the server serves no FM while clients' DMs are initialized with FMs, and client's domain-specific FMs are trained with full model fine-tuning (see Eq.(\ref{eq:setting_3_variant_0})) or parameter-efficient fine-tuning techniques (see Eq.(\ref{eq:setting_3_variant})). We formulate the specific objectives of each client $k$ as follows.
\begin{align}
    \textit{federated full-model training or fine-tuning:}\quad & \min_{\mathcal{W}^{G}_{\text{FM}}} \ell_{\text{TA}}^{C_k}(g_{\mathcal{W}^{G}_{\text{FM}}}(x^{C_k}), y^{C_k}).\label{eq:fedft}\\
    \textit{federated parameter-efficient fine-tuning:}\quad & \min_{\mathcal{W}^{G}_a} \ell_{\text{TA}}^{C_k}(\phi_{\mathcal{W}^{G}_a} \oplus g_{\mathcal{W}^{C_k}_{\text{FM}}}(x^{C_k}), y^{C_k}).\label{eq:fedpeft}
\end{align}
where $g_{\mathcal{W}^{G}_{\text{FM}}}$ is the global FM, $g_{\mathcal{W}^{C_k}_{\text{FM}}}$ is the local FM of client $k$, and $\phi_{\mathcal{W}^{G}_a}$ is the global PEFT module parameterized by $\mathcal{W}^{G}_a$ (e.g., adaptor and LoRA) and shared by all clients; $\oplus$ denotes an operator that composes $g_{\mathcal{W}^{C_k}_{\text{FM}}}$ with PEFT module $\phi_{\mathcal{W}^{G}_a}$; $\ell_{\text{TA}}^{C_k}$ is the task loss for optimizing the global FM or PEFT module $\phi_{\mathcal{W}^{G}_a}$ based on client $k$'s local data $x^{C_k}$ and $y^{C_k}$.


In federated full-model training or fine-tuning~\cite{lin2022fednlp,woi2023fededge}, clients initialize their DMs with FMs having the same architecture and conduct knowledge transfer by sharing their full DMs with the server for aggregation. In federated parameter-efficient fine-tuning~\cite{zhang2023fedpetuning,Zhao2023fedprompt,cai2023fedadapter,zhang2023fedit}, clients initialize their DMs with FMs that may have different architectures and conduct knowledge transfer by sharing partial or proxies of their DMs. 

\subsection{Formulations of Privacy Leakage}

The privacy leakage $\epsilon_p$ measures the privacy loss incurred by a computation (e.g., deep learning) when certain privacy protection mechanisms are applied. It can be defined in various forms. Herein, we introduce Differential privacy leakage~\cite{abadi2016dpsgd,sommer2018privacy} and Bayesian privacy leakage~\cite{Zhang2022fednfl}. 


\textit{Differential privacy leakage.} Let $\epsilon_{p}^{C_k}$ represent the privacy leakage of client $k$. Given a randomization mechanism $\mathcal{M}$, neighboring databases $\mathcal{D}^{C_k},\hat{\mathcal{D}}^{C_k}$ of client $k$, the privacy leakage of the federated learning is defined as follows.
\begin{equation}\label{eq:dp}
\epsilon_p = \frac{1}{K}\sum_{k=1}^K \epsilon_{p}^{C_k},  \text{ where } \epsilon_{p}^{C_k}(o) = \text{log}\frac{Pr[\mathcal{M}(\mathcal{D}^{C_k})=o]}{Pr[\mathcal{M}(\hat{\mathcal{D}}^{C_k})=o]}
\end{equation}
where $\epsilon_{p}^{C_k}(o)$ is the privacy leakage of outcome $o$ at client $k$.


The $\epsilon_p$ in formulation Eq.(\ref{eq:dp}) quantifies the privacy loss that an individual faces when its data is included in a computation and establishes the correspondence between the maximum amount of privacy leakage and the level of privacy protection. A smaller value of $\epsilon_p$ implies a stronger privacy guarantee. The exact relationship between the privacy leakage and the level of privacy protection may vary depending on the specific differentially private mechanism used. Different mechanisms have different privacy guarantees and may incur privacy leakage differently. The widely accepted differentially private mechanism in deep learning is DP-SGD~\cite{abadi2016dpsgd}.

Although DP-SGD is a commonly used protection mechanism, it has limitations. Firstly, DP-SGD and its variations~\cite{kairouz2021dpftrl,shi2022sdp} generally do not consider the prior distributions of privacy data, thereby being prior-independent. Consequently, they are inadequate for modeling recent Bayesian privacy attacks, for example, DLG~\cite{zhu2020deep}. Secondly, differential privacy protects data privacy by introducing random noise during computation, and it fails to consider other protection mechanisms, such as sparsification. To mitigate these drawbacks, \citet{Zhang2022fednfl} introduced Bayesian privacy leakage to federated learning.

\textit{Bayesian privacy leakage.} Let $\epsilon_{p}^{C_k}$ represent the privacy leakage of client $k$. The privacy leakage of the federated learning is defined as follows.
\begin{equation}\label{eq: def_of_pl}
\epsilon_p = \frac{1}{K}\sum_{k=1}^K \epsilon_{p}^{C_k}, \text{ where } \epsilon_{p}^{C_k} = \sqrt{{\text{JS}}(\mathcal{P}_{A}^{C_k} || \mathcal{P}_{O}^{C_k})} 
\end{equation}
where $\mathcal{P}_{A}^{C_k}$ and $\mathcal{P}_{O}^{C_k}$ represent the attacker's belief distribution about private data $\mathcal{D}^{C_k}$ of client $k$ upon observing the protected information and without observing any information, respectively, and $\text{JS}(\cdot||\cdot)$ denotes Jensen-Shannon divergence between two distributions~\cite{Zhang2022fednfl}. 

Bayesian privacy leakage measures the variation between the adversary’s prior and posterior beliefs on private data. Based on Bayesian privacy leakage,  \citet{Zhang2022fednfl} provided lower bounds of $\epsilon_p$ for three privacy-preserving mechanisms, including Randomization, Sparsification, and Homomorphic Encryption.

\section{FTL-FM Taxonomy}\label{sec:category}

We break down the general questions we raised in the introduction into the following five research issues: (1) What to transfer; (2) How to transfer; (3) What to protect; (4) How to protect; (5) How to attack. Based on the three FTL-FM settings and the five research issues, we construct a taxonomy to categorize state-of-the-art FTL-FM works, as shown in Table \ref{tab:ftl-fm-tree}.

\begin{itemize}
\item "What to Transfer (WT)" asks what kind of knowledge is transferred between participating parties of federated learning. Motivated by ~\cite{pan2010stf}, we consider three kinds of knowledge that can be transferred between parties: \textit{data-level knowledge}, \textit{representation-level knowledge}, and \textit{model-level knowledge}. Specifically, \textit{data-level knowledge} refers to synthetic or original data from a dataset as well as prompts manually or automatically generated. \textit{representation-level knowledge} refers to intermediate training results such as representations, logits, and gradients. \textit{model-level knowledge} refers to model parameters.

\item "How to Transfer (HT)" asks about the specific transfer method leveraged by an FTL-LLM work. We provide five categories of transferring methods: \textit{data transfer} for transferring data-level knowledge; \textit{split learning} and \textit{representation transfer} for transferring representation-level knowledge; \textit{homo-model transfer} and \textit{hetero-model transfer} for transferring model-level knowledge. From Section \ref{sec:fm_to_dm} to Section \ref{sec:fm_dm_coevolve}, we will discuss the specific transfer method used by each FTL-FM work.

\item "What to Protect (WP)" asks what information is protected during knowledge transfer. We consider four kinds of information that can be protected during knowledge transfer: data ($\mathcal{D}^S$) and model ($\mathcal{W}^S$) of the server as well as data ($\mathcal{D^C}$) and model ($\mathcal{W^C}$) of a client. Particularly, we want to protect data privacy as well as model ownership and performance. 

\item "How to Protect (HP)" asks about the specific protection method (e.g., differential privacy and secure aggregation) used to protect the target information. 

\item "How to Attack (HA)" asks the way to attack the protected information. For conciseness, we present \textit{semi-honest} and \textit{malicious} attacks in the taxonomy for categorization. Semi-honest attackers follow federated learning training and inference protocols but may try to infer the private data of participating parties based on observed information, while malicious attackers may update intermediate training results or model architecture maliciously to extract private information or jeopardize the model performance of participating parties. 

\end{itemize}

The taxonomy (see Table \ref{tab:ftl-fm-tree}) elucidates settings and technical details of state-of-the-art FTL-FM works. Note that Table \ref{tab:ftl-fm-tree} presents only FTL-FM works that consider all the research issues discussed aforementioned and focus on the \textit{training} phase\footnote{We are aware that many FTL-FM works are dedicated to knowledge transfer techniques without explicitly protecting data privacy, model ownership, or defending against backdoor attacks. We will discuss these works in detail from Section \ref{sec:fm_to_dm} to Section \ref{sec:fm_dm_coevolve}}.

From Table \ref{tab:ftl-fm-tree}, we can see that most of the FTL-FM works focus on the setting of transferring and adapting knowledge of the FM to DMs (i.e., the first setting), and these works involve a wide range of knowledge transfer techniques, including data transfer, split learning, homo-model transfer, and hetero-model transfer. This is expected because adapting or customizing FMs to downstream domain-specific tasks is a pivotal research area. The setting of augmenting FMs with domain-specific knowledge (i.e., the second setting) is rarely studied, involving only offsite-tuning~\cite{xiao2023offsite} and its federated learning counterparts~\cite{fan2023fate}. The third setting focuses on investigating model transfer techniques to co-optimize FMs and DMs.



From the perspective of data and model protection, most of the FTL-FM works (superscripted by $*$) focus on protecting data privacy under semi-honest attacks. Backdoor (superscripted by $\ddagger$), including both attack and defense, is also an active research area. While the model ownership protection (superscripted by $\dagger$) is under-explored. 

Next, we will review FTL-FM works presented in Table 2 and other relevant works that fall within the spectrum of FTL-FM in Sections \ref{sec:fm_to_dm}, \ref{sec:dm_to_fm}, and \ref{sec:fm_dm_coevolve}, respectively. These methods can be reduced to the formulations defined under the FTL-FM framework (see Definition \ref{def:ftl-llm}). We will indicate their corresponding relationships in the discussion.

\begin{landscape}
\begin{table*}[!h]
 	\centering
\includegraphics[width=0.99\linewidth]{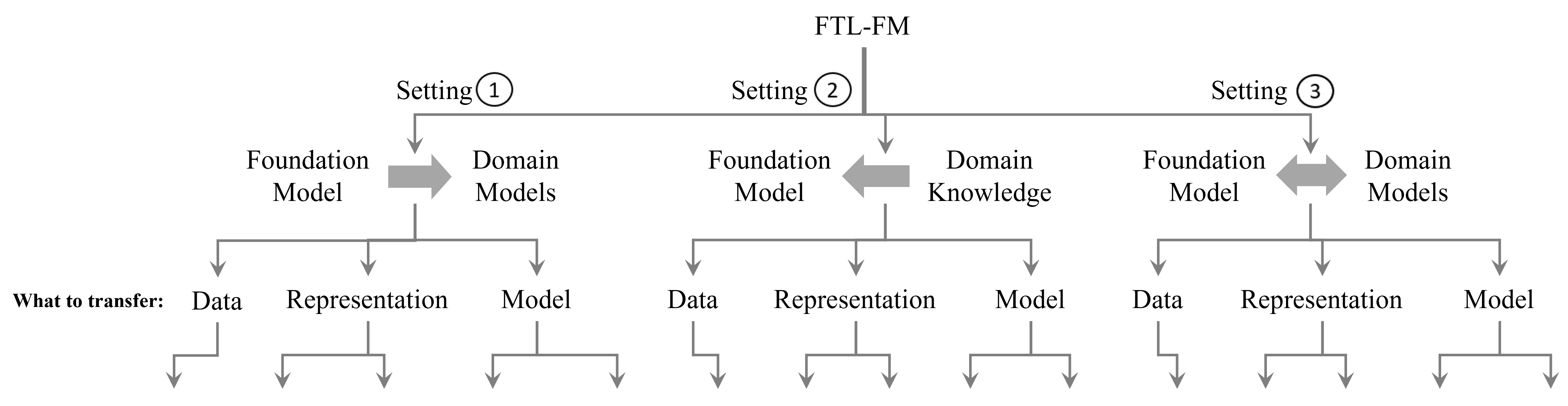}
\begin{threeparttable}
 \tiny
\setlength{\tabcolsep}{4.5pt}
	\begin{tabular}{c|c||c|c|c|c|c||c|c|c|c|c||c|c|c|c|c}
		\multirow{3}{*}{HA} & \multirow{3}{*}{\backslashbox[8mm]{WP}{HT}}
        & \multirow{3}{*}{\shortstack{Data\\Transfer}} & \multirow{3}{*}{\shortstack{Split \\Learning}} & \multirow{3}{*}{\shortstack{Repr\\Transfer}} & \multirow{3}{*}{\shortstack{Homo-\\Model \\Transfer}} & \multirow{3}{*}{\shortstack{Hetero-\\Model\\Transfer}}
        & \multirow{3}{*}{\shortstack{Data\\Transfer}} & \multirow{3}{*}{\shortstack{Split \\Learning}} & \multirow{3}{*}{\shortstack{Repr\\Transfer}} & \multirow{3}{*}{\shortstack{Homo-\\Model \\Transfer}} & \multirow{3}{*}{\shortstack{Hetero-\\Model\\Transfer}}
        & \multirow{3}{*}{\shortstack{Data\\Transfer}} & \multirow{3}{*}{\shortstack{Split \\Learning}} & \multirow{3}{*}{\shortstack{Repr\\Transfer}} & \multirow{3}{*}{\shortstack{Homo-\\Model \\Transfer}} & \multirow{3}{*}{\shortstack{Hetero-\\Model\\Transfer}}\\
       ~ &&&&&&&&&&&&&&&&~\\
       ~ &&&&&&&&&&&&&&&&~\\
	    \hline
	    \hline
		\multirow{8}{*}{\shortstack{\begin{turn}{90}Semi-Honest\end{turn}}} & \multirow{4}{*}{\shortstack{$\mathcal{D}^S$ \\($\mathcal{W}^S$)}} & \multirow{4}{*}{\shortstack{PPDG\\~\cite{carranza2023privacy,kurakin2023harnessing}$^*$\\ ~\cite{mattern2022sds,yue2023synthetic}$^*$}} & \multirow{4}{*}{\shortstack{}} &  \multirow{4}{*}{\shortstack{}} & \multirow{4}{*}{\shortstack{DP\\~\cite{behnia2022ew,bu2022dptitfif,shi2022jft}$^*$,\\MP~\cite{youssef2023proxydata}$^*$}} & \multirow{4}{*}{\shortstack{DP\\~\cite{yu2022differentially, mire2022dpkd}$^*$}} & & & & & \multirow{4}{*}{\shortstack{CP~\cite{xiao2023offsite}\\~\cite{fan2023fate,chua2023fedpeat,khalid2023cefhri}$^\dagger$}} & & & &\\
       ~ &&&&&&&&&&&&&&&&~\\
       ~ &&&&&&&&&&&&&&&&~\\
       ~ &&&&&&&&&&&&&&&&~\\
	 \cline{2-17}
		~  & \multirow{6}{*}{\shortstack{$\mathcal{D}^C$ \\($\mathcal{W}^C$)}} & & \multirow{6}{*}{\shortstack{DP~\cite{li2023privacy,shen2023sap}$^*$,\\WPE~\cite{xu2023shuffled}$^*$,\\PTB~\cite{zhou2023textobfuscator}$^*$}} & & \multirow{6}{*}{\shortstack{PPDS\\ \cite{hou2023FreD, yu2023selective}$^*$}} & \multirow{6}{*}{\shortstack{DP~\cite{wang2023can}$^*$}} & ~
  & ~ &  \multirow{6}{*}{\shortstack{}} & ~ & \multirow{6}{*}{\shortstack{SA~\cite{fan2023fate,fedost}$^*$}} & & & & \multirow{6}{*}{\shortstack{DP~\cite{azam2023fl4asr,balunovic2022lamp}$^*$,\\ FE~\cite{gupta2022recovering}$^*$ \\ HE~\cite{jin2023fedhe}$^*$}} & \multirow{6}{*}{\shortstack{DP\\~\cite{xu2023lvnlm,Zhao2023fedprompt}$^*$,\\ PEFT~\cite{zhang2023fedpetuning}$^*$,\\SA~\cite{fan2023fate}$^*$\\TEE~\cite{huang2024fast}$^{*\dagger}$}}\\
        ~ &&&&&&&&&&&&&&&&~ \\
        ~ &&&&&&&&&&&&&&&&~\\
        ~ &&&&&&&&&&&&&&&&~\\
       ~ &&&&&&&&&&&&&&&&~\\
        ~ &&&&&&&&&&&&&&&&~\\
		\cline{1-17}

	 \multirow{7}{*}{\shortstack{\begin{turn}{90}Malicious\end{turn}}} & \multirow{3}{*}{\shortstack{$\mathcal{D}^S$ \\($\mathcal{W}^S$)}} &  & & & & & ~ & & & & & & & & \multirow{3}{*}{\shortstack{NC, DP\\ \cite{zhang2022neurotoxin, yoo2022backdoor}$^\ddagger$}} & \\
      ~  & ~  &  ~  &  ~ & & & & & & & & & & & &  \\
      ~  & ~  &  ~  &  ~ & & & & & & & & & & & &  \\
	  \cline{2-17}

	~ & \multirow{5}{*}{\shortstack{$\mathcal{D}^C$ \\($\mathcal{W}^C$)}} &  &  &  & \multirow{5}{*}{\shortstack{FP~\cite{Zhang2021neuba,Shen2021por}$^\ddagger$,\\ OWD~\cite{chen2022badpre,mei2023notable}$^\ddagger$,\\MDP~\cite{xi2023mdp}$^\ddagger$}} & \multirow{5}{*}{\shortstack{OWD~\cite{cai2022badprompt}$^\ddagger$}}& & & & & & & & & \multirow{5}{*}{\shortstack{DP~\cite{chu2023panning}$^*$\\~\cite{fowl2023decepticons,rashid2023fltrojan}$^*$\\SCR~\cite{rashid2023fltrojan}$^*$}} \\
        ~  & ~  &  ~  &  ~ & & & & & & & & & & & & \\
       ~  & ~  &  ~  &  ~ & & & & & & & & & & & & \\
        ~  & ~  &  ~  &  ~ & & & & & & & & & & & & \\
     ~ & ~ & ~ & ~ & & & & & & & & & & & & \\
		\hline
	\end{tabular}


\end{threeparttable}

\begin{flushleft}
\scriptsize

\begin{tabular}{l}
Hetero-Model Transfer means the model transferred to other parties has a \textit{different} architecture from the original one; If multiple clients are involved, their model architectures are different.\\
\end{tabular}

\begin{tabular}{lll}
$*$ signifies works defending against data privacy attacks.  $\quad$ & $\dagger$ signifies works defending against model stealing attacks. $\quad$ & $\ddagger$ signifies works defending against backdoor attacks.  \\
\end{tabular}


\begin{tabular}{lllll}
{\textbf{Item to be protected:}} & $\mathcal{D}^S$: data of server & $\mathcal{D}^C$: data of client & $\mathcal{W}^S$: model of server & $\mathcal{W}^C$: model of client \\
\end{tabular}

\begin{tabular}{l l l l l }
\textbf{Protection Methods:} & & & &\\
 DP: Differential Privacy & PPDG: Privacy-Preserving Data Generation & MP: Model Trained via Proxy Data & MDP: Masking-Differential Prompting & TEE: Trusted Execution Environment\\

 SA: Secure Aggregation & PPDS: Privacy-Preserving Data Selection & PEFT: Parameter-Efficient Fine-Tuning & HE: Homomorphic Encryption\\

 NC: Norm Clipping & FE: Freezing Embeddings & PTB: Perturbation & CP: Compression\\

 FP: Fine Pruning & WPE: Weight Permutation Equivalence & OWD: Outlier Word Detection & SCR: Scrubbing\\
  
\end{tabular}
\end{flushleft}
	\caption{Taxonomy of existing FTL-FM works. This taxonomy only presents works focusing on the \textit{learning} (e.g., training, fine-tuning, optimization) phase.}\label{tab:ftl-fm-tree}
\end{table*}
\end{landscape}

\section{Foundation Models for Enhancing Domain Models}\label{sec:fm_to_dm}

In this section, we overview FTL-FM methods that fall into setting \textcircled{1} (see Definition \ref{def:ftl-llm}), the objective of which is to transfer and adapt the knowledge of foundation models (FMs) to downstream clients' domain models (DMs) for exploiting the power of FMs. To this end, the literature proposed a variety of methods involving data-level knowledge transfer, representation-level knowledge transfer, and model-level knowledge transfer.

\begin{table*}[!ht]
	\caption{\textbf{Summary of data-level knowledge transfer methods of setting \textcircled{1}}. $\bigcirc$ denotes that a reference work intends to protect data or model against privacy attacks using certain protection presented in the "How To Protect" columns. DP: Differential Privacy; $S_{\text{FM}}$: FM server; $S_{\text{Fed}}$: federated server; $X$, $Y$, $\hat{Y}$, and $P$ are input, ground truth response, response generated by server's FM, and prompt, respectively. Fr.: From.} 
	\centering
\scriptsize
\bgroup
\def\arraystretch{1.3}
  \setlength{\tabcolsep}{1.2pt}
	\begin{tabular}{c|c||c|c|c|c|c|c|c|c}
	    \hline

    	\multirow{3}{*}{\shortstack{Transfer\\Method\\Category}} & \multirow{3}{*}{\shortstack{Reference}} & \multicolumn{4}{c|}{\shortstack{What To Protect}} & \multicolumn{2}{c|}{\shortstack{Exchanged Information}} & \multicolumn{2}{c}{How To Protect}  \\
 
       \cline{3-10}
        
		~ & ~ & \multicolumn{2}{c|}{\shortstack{Server(s)}} & \multicolumn{2}{c|}{\shortstack{Client(s)}} & \multirow{2}{*}{\shortstack{Server(s)\\To Client(s)}} & \multirow{2}{*}{\shortstack{Client(s)\\To Server(s)}} & \multirow{2}{*}{Server} & \multirow{2}{*}{\shortstack{Client}}\\
		\cline{3-6}

		~ & ~ & \tiny\multirow{1}{*}{$\mathcal{D}^S$} & \tiny\multirow{1}{*}{$\mathcal{W}^S_{\text{FM}}$} & \tiny\multirow{1}{*}{$\mathcal{D}^C$} & \tiny\multirow{1}{*}{$\mathcal{W}^C$} & ~ & ~ & ~ & ~ \\
	  \hline
        \hline
         
     \multirow{10}{*}{\shortstack{Prompt \\Generator\\ Optimization}} & BBT~\cite{sun2022bbt,sun2022bbtv2} & & & &  & \multirow{2}{*}{\shortstack{$\hat{Y}$ from $\mathcal{W}^S_{\text{FM}}$}} & \multirow{2}{*}{\shortstack{$X$ + soft $P$}} & & \\
    	~ & BlackVIP~\cite{oh2023blackvip}, & & & & & & & &  \\
         \cline{2-10}
        ~ & \multirow{1}{*}{BDPL~\cite{diao2023black}} & & & &  & $\hat{Y}$ from $\mathcal{W}^S_{\text{FM}}$ & $X$ + hard $P$ & & \\
         \cline{2-10}
        ~ & \multirow{1}{*}{DSP~\cite{li2023dsp}} & & & &  & Reward from $\mathcal{W}^S_{\text{FM}}$ & $X$ + hard $P$ & & \\
       \cline{2-10}
          ~ & \multirow{1}{*}{Fed-BBPT~\cite{lin2023fedbbpt}} & & & &  &  
          

         \begin{tabular}[c]{p{0.8cm}p{1.7cm}}
         {\tiny Fr. $S_{\text{FM}}$}: &$\hat{Y}$ from $\mathcal{W}^S_{\text{FM}}$ \\
         {\tiny Fr. $S_{\text{Fed}}$}: &Global model \\
         \end{tabular}

          & 
          

         \begin{tabular}{p{0.8cm}p{1.7cm}}
         {\tiny To $S_{\text{FM}}$}: &$X$ + soft $P$ \\
         {\tiny To $S_{\text{Fed}}$}: &$\mathcal{W}^C$ \\
         \end{tabular}
          
          & & \\
      \cline{2-10}
         ~ & \multirow{1}{*}{FedBPT~\cite{sun2023fedbpt}} & & & & & 
         

         \begin{tabular}[l]{p{0.8cm}p{1.7cm}}
         {\tiny Fr. $S_{\text{FM}}$}: &$\hat{Y}$ from $\mathcal{W}^S_{\text{FM}}$ \\
         {\tiny Fr. $S_{\text{Fed}}$}: &Global distr. of $P$\\
         \end{tabular}
         
         & 
         

        \begin{tabular}[l]{p{0.8cm}p{1.7cm}}
         {\tiny To $S_{\text{FM}}$}: &$X$ + soft $P$ \\
         {\tiny To $S_{\text{Fed}}$}: &Local distr. of $P$  \\
         \end{tabular}

         & & \\
      \cline{2-10}
        ~ & \multirow{1}{*}{ZooPFL~\cite{lu2023zoopfl}} & & & &  & 
        

      \begin{tabular}[l]{p{0.8cm}p{1.7cm}}
         {\tiny Fr. $S_{\text{FM}}$}: &$\hat{Y}$ from $\mathcal{W}^S_{\text{FM}}$ \\
         {\tiny Fr. $S_{\text{Fed}}$}: &Global encoder  \\
         \end{tabular}
         
        & 
        

      \begin{tabular}[l]{p{0.8cm}p{1.7cm}}
         {\tiny To $S_{\text{FM}}$}: &Perturbed $X$ \\
         {\tiny To $S_{\text{Fed}}$}: &Local encoder  \\
         \end{tabular}
    
         & & \\
         \cline{1-10}
         \multirow{6}{*}{\shortstack{Discrete \\Prompt \\ Optimization}} & \multirow{1}{*}{APE~\cite{zhou2023ape}} & & & &   & \multirow{2}{*}{\shortstack{$\hat{Y}$ from $\mathcal{W}^S_{\text{FM}}$}} & \multirow{2}{*}{\shortstack{Queries to initialize, \\evaluate, and generate $P$}} & & \\
         ~ & DLN~\cite{sordoni2023dln} & & & &  &  & & \\
         
         \cline{2-10}
        ~ & \multirow{2}{*}{EvoPrompt~\cite{hong2023dpopt}} & & & &  & \multirow{2}{*}{\shortstack{$\hat{Y}$ from $\mathcal{W}^S_{\text{FM}}$}} & \multirow{2}{*}{\shortstack{Queries to crossover,\\ mutate, and evaluate $P$}} & &\\
        ~ & & & & & & & & &  \\
        
        \cline{2-10}
        ~ & \multirow{3}{*}{PromptAgent~\cite{wang2023promptagent}} & & & & & \multirow{3}{*}{\shortstack{$\hat{Y}$ from $\mathcal{W}^S_{\text{FM}}$}} & \multirow{3}{*}{\shortstack{Queries to retrieve error\\ and generate error feedback; \\Queries to update $P$}} & &\\
        ~ & & & & & & & & &  \\
        ~ & & & & & & & & &  \\

         \cline{1-10}
        \multirow{6}{*}{\shortstack{LLM-generated \\Knowledge \\ Transfer}} & \multirow{1}{*}{Fine-tune-CoT~\cite{ho2023large}} & & & &  & \multirow{3}{*}{\shortstack{Reasoning explanations \\from $\mathcal{W}^S_{\text{FM}}$}} & \multirow{3}{*}{\shortstack{$X$ + hard $P$}} & & \\
        ~ & \multirow{1}{*}{Prompting~\cite{magister2023teaching}} & & & & & &  & & \\
        ~ & \multirow{1}{*}{MT-COT~\cite{li2022mtcot}} & & & &  & & & & \\
       \cline{2-10}
        ~ & \multirow{1}{*}{D-SBS~\cite{hsieh2023distilling}} & & & &  & Rationals for $\hat{Y}$ from $\mathcal{W}^S_{\text{FM}}$ & $X$ + hard $P$ & & \\
       \cline{2-10}
        ~ & \multirow{1}{*}{Sci-CoT~\cite{ma2023sci}} & & & &  & Rationals for $X$ from $\mathcal{W}^S_{\text{FM}}$ & $X$ + hard $P$ & & \\
       \cline{2-10}
        ~ & \multirow{1}{*}{PaD~\cite{zhu2023pad}} & & & &  & Reasoning Programs from $\mathcal{W}^S_{\text{FM}}$ & $X$ + hard $P$ & & \\
       \cline{2-10}
        ~ & \multirow{1}{*}{LaMini-LM~\cite{wu2023lamini}} & & & & & Instructions from $\mathcal{W}^S_{\text{FM}}$ & $X$ + hard $P$ & & \\
        
     \cline{1-10}
    \multirow{2}{*}{\shortstack{Synthetic Data \\ Transfer}} & \multirow{2}{*}{\shortstack{\cite{yue2023synthetic,mattern2022sds,kurakin2023harnessing,carranza2023privacy}}} & \multirow{2}{*}{$\bigcirc$} & & & & \multirow{2}{*}{\shortstack{Data generated via\\ DP-tuned $\mathcal{W}^S_{\text{FM}}$}} & & \multirow{2}{*}{DP} &  \\
    	 ~ & & & & & & & & &  \\
       \hline
	\end{tabular}
 \egroup
\label{table:server_llm_to_clients_data_transfer}
\end{table*}

\subsection{FM adapted to DMs through data-level knowledge transfer} 

The literature has explored mainly four categories of data-level knowledge transfer methods that adapt the knowledge of FMs to a client of a specific domain: (1) the client prompts tuning a closed-source FM (e.g., ChatGPT) by optimizing a local prompt generator based on its domain-specific data and predictions generated by the FM. (2) the client optimizes discrete prompts with the guidance of FMs. (3) the client fine-tunes its domain model with knowledge generated by FMs. (4) An FM server transfers data generated by general-purpose FMs or FMs augmented with industry-level knowledge (e.g., LawGPT~\cite{nguyen2023lawgpt}, FinGPT~\cite{yang2023fingpt}, BloombergGPT~\cite{wu2023bloomberggpt}, and Med-PaLM2~\cite{singhal2023medPaLM2}) to clients of specific domains. We summarize data-level knowledge transfer methods that fall into setting \textcircled{1} in Table \ref{table:server_llm_to_clients_data_transfer}\footnote{While we consider an FTL-FM approach should incorporate certain protection mechanisms to protect data or models, we still review highly relevant works that prioritize the development of knowledge transfer methods without considering data or model protection and include them in the Table to ensure thoroughness. (same for Section \ref{sec:dm_to_fm} and Section \ref{sec:fm_dm_coevolve}). These methods leave "blank" in the "What To Protect" and "How To Protect" columns.} and elaborate on these methods in this subsection.

\begin{figure*}[!t]
    \centering
    \includegraphics[width=0.95\linewidth]{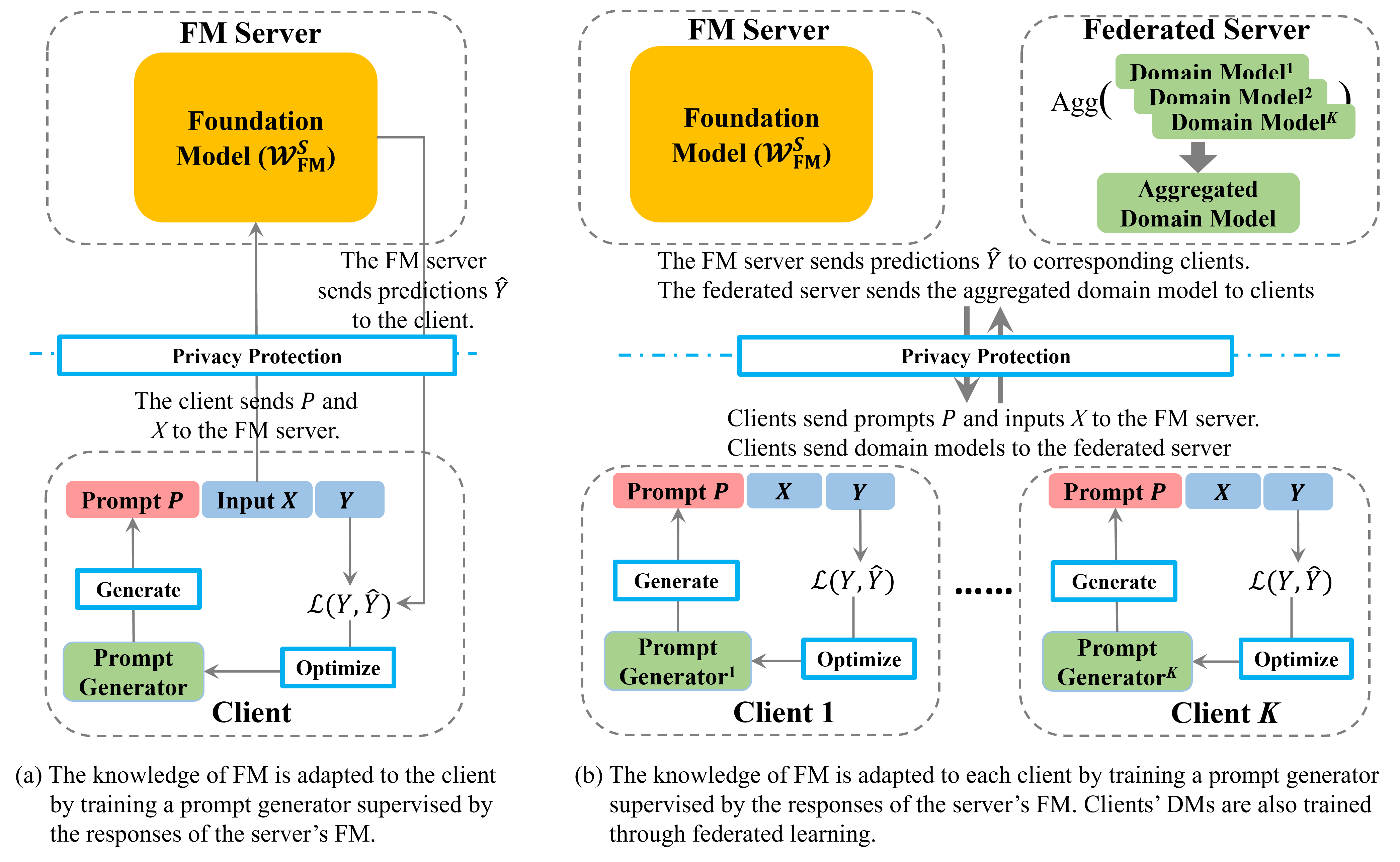}
    \caption{Illustration of adapting knowledge of the server's foundation model (FM) to clients' domain models (DMs) through prompt generator tuning. (a) illustrates that a client can adapt the knowledge of the server's FM to its local task via training a local prompt generator. (b) extends the scenario of (a) to multiple clients. (b) illustrates that each client trains a local prompt generator supervised by predictions of the FM, and these clients also leverage the FedAvg algorithm or its variants with the help of the federated server to train their local domain models.}
    \label{fig:prompt_generator_tuning}
\end{figure*}

\subsubsection{Prompt Generator Optimization}
Alternatively, clients can adapt knowledge of the FM to their specific domains through federated prompt optimization. A typical setting of federated prompt optimization involves a client that learns a local prompt generator to generate prompts that can guide the FM toward the desired responses (illustrated in Figure \ref{fig:prompt_generator_tuning}(a)). 

BBT~\cite{sun2022bbt}, BBTv2~\cite{sun2022bbtv2}, and BlackVIP~\cite{oh2023blackvip} proposed that a client can customize a remote FM (e.g., ${\text{RoBERTa}}_{\text{large}}$~\cite{Liu2019RoBERTaAR}, ${\text{BART}}_{\text{large}}$~\cite{lewis2020bart}, and $\text{ViT-B}$~\cite{radford2021vit}) by optimizing a domain-specific prompt generator based on its local data together with the predictions (e.g., logits) generated by the remote FM. BBT and BBTv2 employ a gradient-free method called CMA-ES (Covariance Matrix Adaptation Evolution Strategy)~\cite{hansen2016cma} to optimize the client's local prompt generator, whereas BlackVIP leverages SPSA (Simultaneous Perturbation Stochastic Approximation)~\cite{spall1992sasp} to efficiently approximates the high-dimensional gradients, which then are used to optimize the prompt generator. BBT, BBTv2, and BlackVIP require the FM API to take continuous (i.e., soft) prompts as input, and thus, they cannot be applied to FM APIs that only accept discrete inputs. To eliminate this constraint, BDPL~\cite{diao2023black} and DSP~\cite{li2023dsp} were proposed to learn a client side's discrete (i.e., hard) prompt generator (DPG) for customizing FM's responses. BDPL adopts a variance-reduced policy gradient algorithm~\cite{zhou2021efficient} to optimize the DPG, while DSP optimizes the DPG through supervised fine-tuning and reinforcement learning. BBT, BBTv2, BlackVIP, BDPL, and DSP involve only one client, and thus, they correspond to stage 2 of the first objective formulated in Eq.(\ref{eq:setting_1_stage_2}) with $K=1, p^{C_1}=1$, $\eta=0$, the domain model $\mathcal{W}^{C_1}$ is the client's local prompt generator and the specific form of the loss $\ell^{C_1}(\mathcal{W}^S_{\text{FM}}, \mathcal{W}^{C_1};d)$ is formulated in Eq.(\ref{eq:fedpt}).

The aforementioned federated prompt tuning methods involve an FM server and only one client. Fed-BBPT~\cite{lin2023fedbbpt} and FedBPT~\cite{sun2023fedbpt} extend the federated prompt tuning to multiple clients, the workflow of which is illustrated in Figure ~\ref{fig:prompt_generator_tuning}(b). More specifically, Fed-BBPT and FedBPT involve two servers: an FM server, denoted as $S_{\text{FM}}$, and a federated server, denoted as $S_{\text{Fed}}$. $S_{\text{FM}}$ hosts an FM to be prompted by clients, whereas $S_{\text{Fed}}$ is for federated tuning clients' local prompt generators. In Fed-BBPT, each client leverages the Simultaneous Perturbation Stochastic Approximation~\cite{SPALL1997109} to train its prompt generators based on local data and predictions generated by the FM, and the federated server aggregates clients' local prompt generators to form the global prompt generator. FedBPT differs from Fed-BBPT in that each client in FedBPT exploits CMA-ES to optimize the distribution of prompts generated by its prompt generator, and the server derives the global prompt generator by aggregating local distributions of clients' prompts. Generally, Fed-BBPT and FedBPT adhere to the objective formulated in Eq.(\ref{eq:setting_1_stage_2}) with $K>1$, $\eta=0$, the domain model $\mathcal{W}^{C_k}$ of each client $k$ is a prompt generator and the specific loss of federated prompt tuning $\ell^{C_k}(\mathcal{W}^S_{\text{FM}}, \mathcal{W}^{C_k};d)$ between the FM server and a client $k$ is formulated in Eq.(\ref{eq:fedpt}).

ZooPFL~\cite{lu2023zoopfl} adapts the knowledge of FM to downstream clients' tasks by training a client-specific embedding and a semantic re-mapping module for each client. More specifically, each client appends the client-specific embedding to the embedding generated by a pre-trained auto-encoder, which aims to make the client's local input consistent with the FM. Clients also collaboratively fine-tune encoders of their auto-encoders through FedAVG~\cite{mcmahan2017communication} to share knowledge. The semantic re-mapping module aligns the logits output from the FM with the semantic space of clients' local tasks. ZooPFL adheres to the objective formulated in Eq. (\ref{eq:setting_1_stage_2}) with $K>1$, $\eta=0$. The domain model $\mathcal{W}^{Ck}$ of each client $k$ in ZooPFL is the client-specific embedding and semantic re-mapping module.

\begin{figure*}[!ht]
    \centering
    \includegraphics[width=0.99\linewidth]{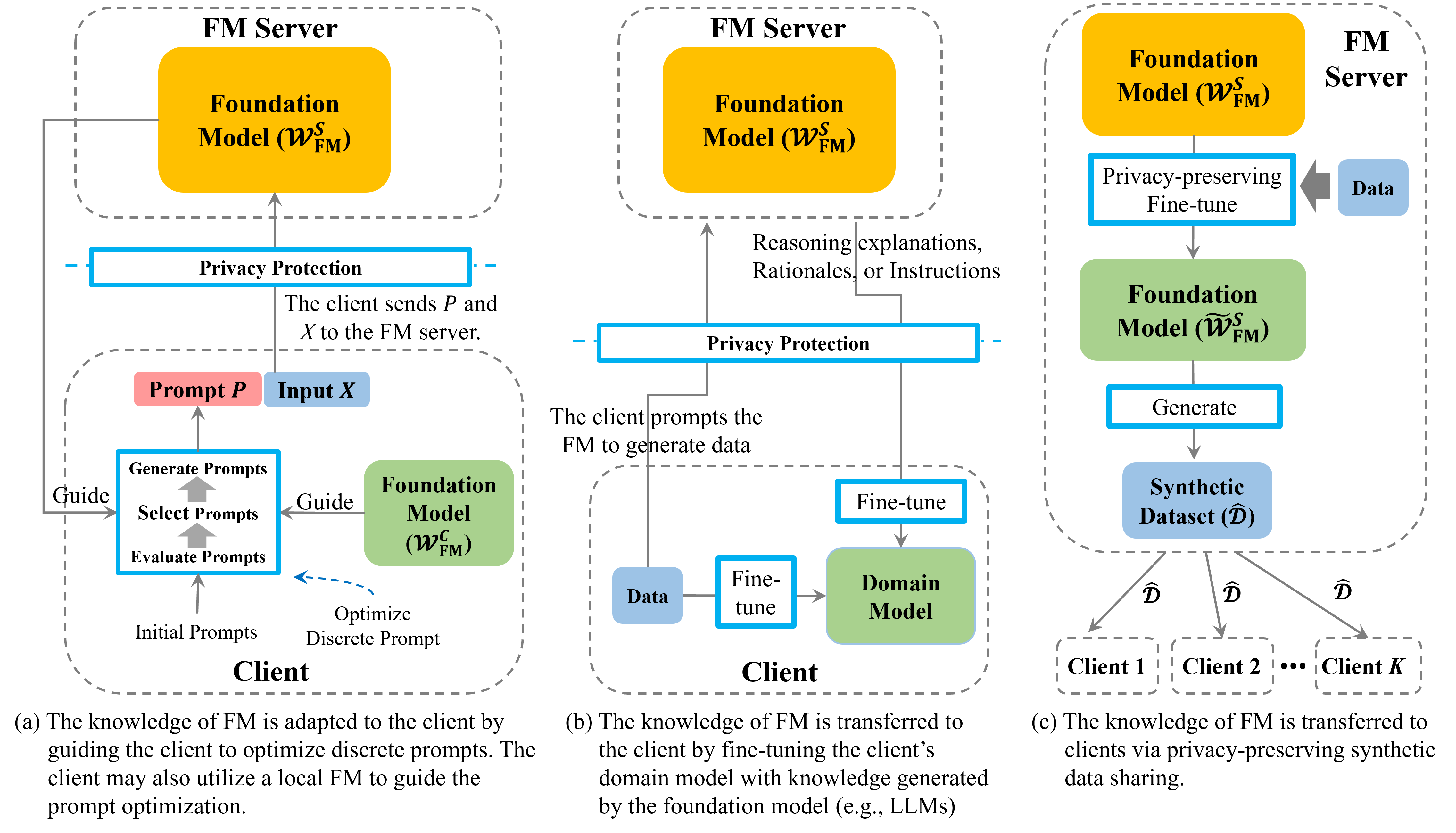}
    \caption{Illustration of federated discrete prompt tuning, LLM-generated knowledge transfer, and synthetic data transfer. (a) illustrates that a client adapts the knowledge of the FM to its local task by querying the FM to guide the discrete prompt optimization. The client may also turn to a local FM to provide guidance. (b) illustrates that a client queries an LLM to generate emergent ability knowledge and leverages this knowledge to fine-tune its local domain model. (c) illustrates that the knowledge of FM is transferred to downstream clients via privacy-preserving synthetic data sharing.}
    \label{fig:data_transfer_2}
\end{figure*}

\subsubsection{Discrete Prompt Optimization}
Another line of research on federated prompt tuning is federated discrete prompt optimization (FedDPO), in which the client aims to find the optimal discrete prompt that maximizes the performance of the queried FM $\mathcal{W}^S_{\text{FM}}$ towards a score function $\mathcal{R}$ (e.g., accuracy) on certain validation data $\mathcal{D}_{\text{val}}$. The objective of the FedDPO is a variant of Eq.(\ref{eq:fedpt}). We write this objective as follows:
\begin{equation}\label{eq:ape}
\begin{split}
    P^{*}=\argmax_{P} \mathcal{R} (f_{\mathcal{W}^S_{\text{FM}}}([X,P])|\mathcal{D}_{\text{val}})
\end{split}
\end{equation}
where $X$ is the input, $P$ is the prompt and $[X,P]$ forms the query to the $f_{\mathcal{W}^S_{\text{FM}}}$ hosted by a server.

Figure \ref{fig:data_transfer_2}(a) provides a high-level illustration of FedDPO. Generally, the discrete prompt optimization procedure involves three steps: evaluate existing prompts, select prompts with the best scores, and generate candidate prompts. The three steps iterate until the algorithm converges. During the iteration, the client keeps querying the server's FM and possibly an auxiliary local FM to guide its prompt optimization. Various methods have been proposed to implement Eq.(\ref{eq:ape}). 
In particular, APE~\cite{zhou2023ape} adopts Iterative Monte Carlo Search to explore the prompt search space. During the optimization, APE queries a black-box FM (i.e., GPT-3) to score candidate prompts and generate new prompts. DLN~\cite{sordoni2023dln} extended APE and proposed Stacked LLMs to guide the prompt optimization. EvoPrompt~\cite{guo2023evoprompt} utilizes evolutionary algorithms to optimize prompts and leverage a black-box FM (i.e., GPT-3.5) to perform evolutionary operators such as crossover and mutation in the Genetic Algorithm. PromptAgent~\cite{wang2023promptagent} proposed an agent-based framework for optimizing prompts, aiming to produce expert-level task prompts via strategic planning and reflecting with error feedback during the optimization. Specifically, PromptAgent applies Monte Carlo Tree Search to orchestrate the prompt optimization: given a current prompt, the agent first queries an FM (i.e., GPT-3.5) to collect errors from the task dataset; next, it queries an FM (i.e., GPT-4) to provide error feedback, and then updates the prompt according to the feedback and transits to the next iteration.

While these federated prompt optimization methods (including prompt generator tuning and discrete prompt tuning) work well in adapting knowledge of FMs to clients' domain-specific tasks, they put the privacy of clients' local data at high risk since they send (soft or hard) prompts and input text directly to the FM server without any privacy protection. For applications where (training and inference) data contains sensitive and private information of users, privacy-preserving techniques such as DP-rewrite~\cite{timour2022dprewrite} should be applied to data transmitted between parties. 

\subsubsection{LLM-generated Knowledge Transfer}\label{sec:llm_kd}
Large-scale FMs (i.e., LLMs) have demonstrated remarkable abilities, coined as emergent abilities~\cite{wei2022emergent}, which encompass multiple captivating aspects, such as In-Context Learning (ICL), Chain-of-Thought (CoT), and Instruction Following (IF). LLM-generated Knowledge Distillation aims to efficiently fine-tune small LLMs for a client using knowledge generated by a powerful LLM. Figure \ref{fig:data_transfer_2}(b) provides a high-level illustration of LLM-generated Knowledge Distillation. 

Fine-tune-CoT~\cite{ho2023large}, CoT Prompting~\cite{magister2023teaching} and MT-COT~\cite{li2022mtcot} prompt an LLM (e.g., GPT-3 or PaLM) to generate reasoning explanations and leverage these reasoning explanations to fine-tune a student model (e.g., T5). Distilling Step-by-Step (D-SBS)~\cite{hsieh2023distilling} prompts an LLM (e.g., PaLM) to generate labels along with rationales that justify the labels. D-SBS then leverages these rationales to fine-tune a student model (e.g., T5). In addition to rationales, D-SBS also exploits labels of the fine-tuning dataset as additional guidance to fine-tune the student model. Sci-CoT~\cite{ma2023sci} proposed a similar approach as D-SBS, but it conducts the distillation through two small models: The first small model (i.e., Flan-T5-small) is fine-tuned by rationales generated by an LLM (i.e., GPT-3.5), aiming to generate rationales for questions. Subsequently, the second model (i.e., Flan-T5-small) is fine-tuned by the rationales generated by the first model, aiming to answer questions. PaD~\cite{zhu2023pad} queries GPT-3.5 to obtain reasoning programs for input questions and leverages the reasoning programs to fine-tune student models. PaD also leverages self-distillation as an additional supervision term during fine-tuning. To facilitate the deployment of language models in resource-constrained environments, LaMini-LM~\cite{wu2023lamini} generates a large-scale offline distillation dataset comprising 2.58M instructions collected from various existing datasets and generated by ChatGPT (i.e., GPT-3.5), and fine-tune a collection of smaller language models (e.g., T5 and GPT2), which achieve comparable performance to Alpaca. 

These methods involve only one client. They adheres to the objective formulated in Eq.(\ref{eq:setting_1_stage_2}) with $K=1$, $p^{C_1}=1$, $p^S=0$, and $\eta=0$. The specific loss $\ell^{C_1}(\mathcal{W}^S_{\text{FM}}, \mathcal{W}^{C_1};d)$ of training the client $1$' domain model $\mathcal{W}^{C_1}$ using local data and knowledge transferred from server's $\mathcal{W}^S_{\text{FM}}$ is formulated in Eq.(\ref{eq:fedkd_1}). Currently, existing LLM-generated Knowledge Distillation methods primarily emphasize enhancing model performance and training efficiency, ignoring potential privacy concerns that may arise during the knowledge transfer process (as reported in Table \ref{table:server_llm_to_clients_data_transfer}). Nonetheless, privacy holds paramount importance in numerous real-life applications, particularly within the domains of finance and healthcare. Therefore, it is imperative to delve into the realm of privacy-preserving LLM-generated knowledge distillation, as it holds significant practical and research values.

\subsubsection{Synthetic Data Transfer}
In some proprietary domains, such as the audit department in a bank, data resources are typically limited for deep learning tasks. One potential solution to address data scarcity is leveraging FMs pre-trained on a large corpus of domain-specific data to generate synthetic training and testing data (e.g., LawGPT~\cite{nguyen2023lawgpt}). However, domain-specific data (e.g., financial and legal data) typically contain sensitive and private information. Thus, directly releasing FMs trained on these data presents potential privacy issues because sensitive information encoded in FMs can possibly be recovered by adversaries. In order to share synthetic data generated by FMs without compromising the privacy of training data, privacy-preserving data generation methods have been proposed~\cite{carranza2023privacy,yue2023synthetic,mattern2022sds,kurakin2023harnessing}. Essentially, these methods first fine-tune a pretrained public FM using differential privacy (DP) on private domain data and then exploit the DP-tuned FM to generate new data, which can be distributed to downstream clients to augment their training data (illustrated in Figure \ref{fig:data_transfer_2}(c)). These methods correspond to stage 1 of objective \textcircled{1} formulated in Eq.(\ref{eq:setting_1_stage_1}) but with an additional step of generating data using fine-tuned FMs.

\subsection{FM adapted to DMs through representation-level knowledge transfer} 

Representation-level knowledge transfer (RLKT) refers to transferring and adapting knowledge of FMs to domain-specific models through passing representation-level knowledge, such as intermediate forward outputs and backward gradients, between FL participating parties. In literature, \textit{split learning}~\cite{liu2022vertical} and \textit{representation transfer} are two main categories of RLKT explored in the setting \textcircled{1} of FTL-FM. We summarize the two categories of RLKT methods in Table \ref{table:server_llm_to_clients_repr} and elaborate on these methods in this subsection.

\begin{table*}[!ht]
	\caption{\textbf{Summary of representation-level knowledge transfer methods of setting \textcircled{1}}. $\bigcirc$ denotes that a reference work intends to protect data or models against privacy attacks using certain protection presented in the "How To Protect" columns. DP: Differential Privacy; PTB: Perturbation; WPE: Weight Permutation Equivalence; repr: Representation.} 
	\centering
\scriptsize
\bgroup
\def\arraystretch{1.3}
	\begin{tabular}{c|c||c|c|c|c|c|c|c|c}
	    \hline

    	\multirow{3}{*}{\shortstack{Transfer\\Method\\Category}} & \multirow{3}{*}{\shortstack{Reference}} & \multicolumn{4}{c|}{\shortstack{What To Protect}} & \multicolumn{2}{c|}{\shortstack{Exchanged Information}} & \multicolumn{2}{c}{How To Protect}  \\
 
       \cline{3-10}
        
		~ & ~ & \multicolumn{2}{c|}{\shortstack{Server(s)}} & \multicolumn{2}{c|}{\shortstack{Client(s)}} & \multirow{2}{*}{\shortstack{Server(s)\\To Client(s)}} & \multirow{2}{*}{\shortstack{Client(s)\\To Server(s)}} & \multirow{2}{*}{Server} & \multirow{2}{*}{\shortstack{Client}}\\
		\cline{3-6}

		~ & ~ & \tiny\multirow{1}{*}{$\mathcal{D}^S$} & \tiny\multirow{1}{*}{$\mathcal{W}^S_{\text{FM}}$} & \tiny\multirow{1}{*}{$\mathcal{D}^C$} & \tiny\multirow{1}{*}{$\mathcal{W}^C$} & ~ & ~ & ~ & ~ \\
	  \hline
        \hline

        \multirow{10}{*}{\shortstack{Split \\Learning}} & \tiny TextObfuscator~\cite{zhou2023textobfuscator} &  & & $\bigcirc$ & & Backward gradient & Forward repr $+Y$ & & PTB\\
        \cline{2-10}
	   ~ &  \multirow{2}{*}{\shortstack{SAP~\cite{shen2023sap}}} &  & & \multirow{2}{*}{$\bigcirc$} &  & \multirow{2}{*}{\shortstack{Prediction}} & \multirow{2}{*}{\shortstack{Forward repr, \\ Backward gradient}} & & \multirow{2}{*}{PTB} \\
       ~ & & & & & & & & &  \\
        \cline{2-10}
	   ~ &  \multirow{2}{*}{\shortstack{RAPT~\cite{li2023privacy}}} &  & & \multirow{2}{*}{\shortstack{$\bigcirc$}} &  & \multirow{2}{*}{\shortstack{Prediction \\ Backward gradient}} & \multirow{2}{*}{\shortstack{Forward Repr, \\ Backward gradient}} & & \multirow{2}{*}{\shortstack{DP}} \\
       ~ & & & & & & & & &  \\
        \cline{2-10}
	   ~ &  \multirow{2}{*}{\shortstack{Shuffled \\Transformer~\cite{xu2023shuffled}}} & & & \multirow{2}{*}{$\bigcirc$} & & \multirow{2}{*}{\shortstack{Forward Repr, \\ Backward gradient}} & \multirow{2}{*}{\shortstack{Forward Repr, \\ Backward gradient}} & & \multirow{2}{*}{WPE} \\
        ~ & & & & & & & & &  \\
        \cline{2-10}
	   ~ &  \multirow{2}{*}{\shortstack{PrivateLoRA~\cite{wang2023prilora}}} & & &  & & \multirow{2}{*}{\shortstack{Forward Repr, \\ Backward gradient}} & \multirow{2}{*}{\shortstack{Forward Repr, \\ Backward gradient}} & & \\
        ~ & & & & & & & & &  \\
        \cline{2-10}
        
	  ~ &  \multirow{2}{*}{FedBERT~\cite{Tian2022fedbert}} & & & & &  \multirow{2}{*}{\shortstack{Forward Repr, \\ Backward gradient}} & \multirow{2}{*}{\shortstack{Forward Repr, \\ Backward gradient}}  & & \\
      ~ & & & & & & & & &  \\
       \hline

        \multirow{2}{*}{\shortstack{Repr \\Transfer}} &  \multirow{2}{*}{\shortstack{FedGKT~\cite{he2020fedgkt}}} &  & & & & \multirow{2}{*}{\shortstack{Global logit}} & \multirow{2}{*}{\shortstack{Repr + Logit}} & & \\
        ~ & & & & & & & & &  \\
       \hline
	\end{tabular}
 \egroup
\label{table:server_llm_to_clients_repr}
\end{table*}

\begin{figure*}[!ht]
    \centering
    \includegraphics[width=0.95\linewidth]{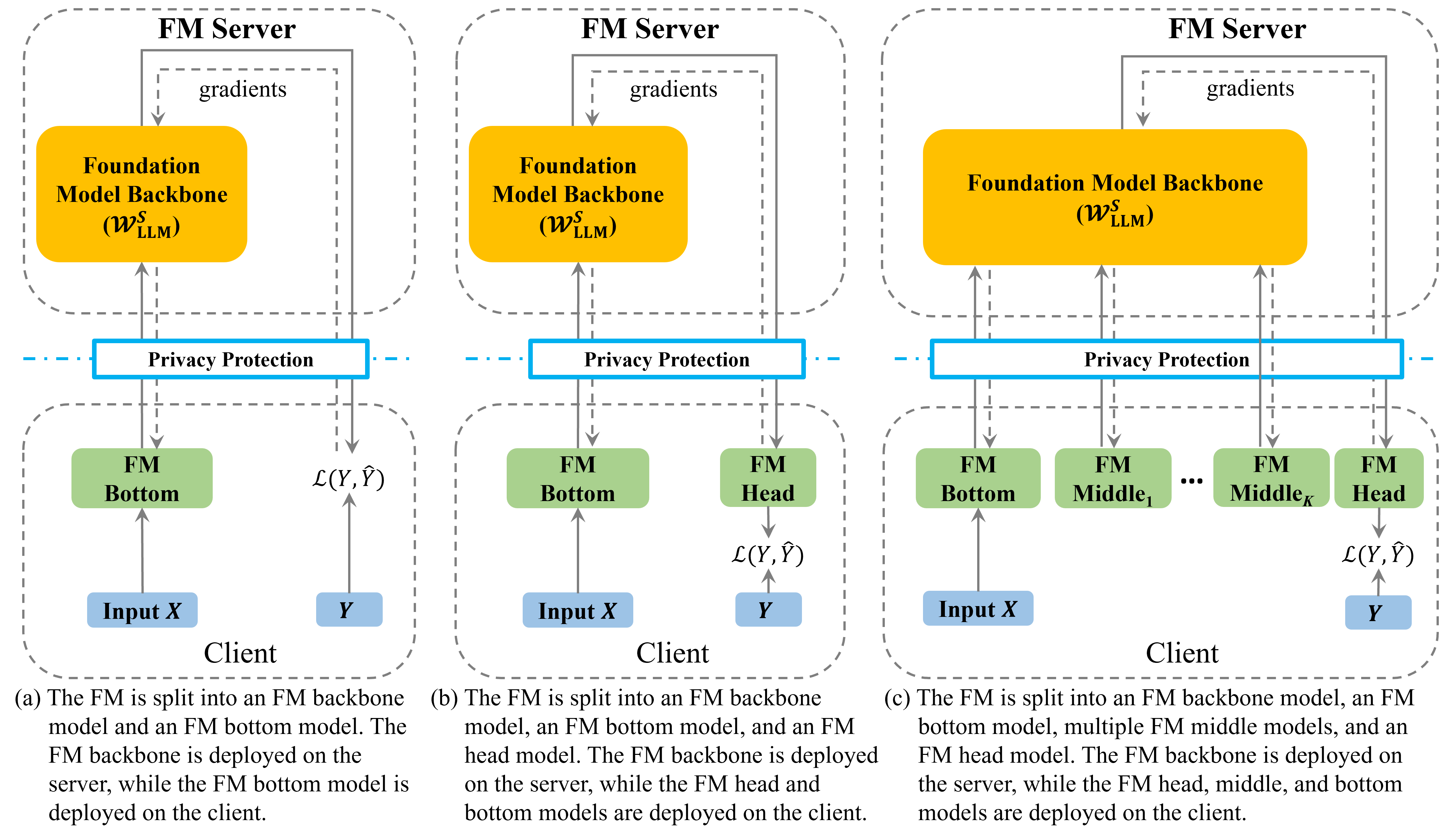}
    \caption{Illustration of adapting knowledge of the server's foundation model (FM) to the client's domain model (DM) through federated split learning that involves one server and one client. The two-party federated split learning generally has three scenarios: (a) The FM is split into an FM backbone model and an FM bottom model. The FM backbone is deployed on the server, while the FM bottom model is deployed on the client; (b) The FM is split into an FM backbone model, an FM bottom model, and an FM head model. The FM backbone is deployed on the server, while the FM head and bottom models are deployed on the client; (c) The FM is split into an FM backbone model, an FM bottom model, multiple FM middle models, and an FM head model. The FM backbone is deployed on the server, while the FM head, middle, and bottom models are deployed on the client.}
    \label{fig:setting_1_sl_1_client}
\end{figure*}

In a two-party federated split learning setting involving a client and a server, the client typically owns a private dataset and a small portion of an FM, while the server owns the rest larger portion. The objective of FM-based split learning is to adapt the knowledge of FM to the client's specific domain by fine-tuning the distributed FM using the client's local data. The FM is often split into an FM backbone and multiple other smaller parts. The FM backbone is the most computationally expensive part and is typically deployed on a server with sufficient computing resources, while other parts can be deployed on either the client or server.
The training procedure of distributed FM follows the conventional split learning procedure using stochastic gradient descent, but each portion of FM is trained by a different party (i.e., client and server). Figure \ref{fig:setting_1_sl_1_client} illustrates three two-party federated split learning settings studied in the literature.  

{\blue{

TextObfuscator~\cite{zhou2023textobfuscator}, RAPT~\cite{li2023privacy}, and Shuffled Transformer \cite{xu2023shuffled} investigated the first two settings of two-party federated split learning, as illustrated in Figure \ref{fig:setting_1_sl_1_client}(a) and Figure \ref{fig:setting_1_sl_1_client}(b). In TextObfuscator~\cite{zhou2023textobfuscator}, a ${\text{RoBERTa}}_{\text{base}}$ model~\cite{Liu2019RoBERTaAR} is divided into a large server model and a smaller client model. The client and server collaboratively train the two models following the split learning protocol. To protect privacy, the client perturbs the representations to be shared with the server with semantically similar alternatives based on clustering. However, the client in TextObfuscator sends labels directly to the server to compute the loss, thereby leaking the private label information. Figure \ref{fig:setting_1_sl_1_client}(a) provides a high-level overview of the workflow of TextObfuscator. RAPT~\cite{li2023privacy} deploys the head and bottom of an FM (i.e., $\text{BERT}_{\text{base}}$~\cite{devlin2019bert} or $\text{T5}_{\text{base}}$~\cite{Raffel2020t5}) on the client side while the FM backbone on the server side, and thus the client in RAPT retains both input texts and labels at local. Besides, RAPT leverages $d_{\chi}$-privacy~\cite{Chatzikokolakis2013dx} to protect the client's private data. Shuffled Transformer investigated the federated split learning setting similar to RAPT, but they focus on designing transformer-based networks (e.g., ViT) that are weight permutation equivalent (WPE). Such WPE networks can prevent the server from inferring the private data of the client without compromising model performance. Figure \ref{fig:setting_1_sl_1_client}(b) overviews the workflow of RAPT and Shuffled Transformer. TextObfuscator, RAPT, and Shuffled Transformer \cite{xu2023shuffled} adhere to the objective formulated in Eq.(\ref{eq:setting_1_stage_2}) with $K=1, p^{C_1}=1$, and $\eta > 0$. Their domain models $\mathcal{W}^{C_1}$ are composed of an FM bottom and, optionally, an FM head. The specific loss of federated split learning $\ell^{C_1}(\mathcal{W}^S_{\text{FM}}, \mathcal{W}^{C_1};d)$ between the server and the client is formulated in Eq.(\ref{eq:fedsplit}).

SAP~\cite{shen2023sap} takes a similar setting as TextObfuscator, but it adopts a different training strategy: the client model is frozen while the server model is being fine-tuned using LoRA. To protect its data privacy, the client applies $d_{\chi}$-privacy~\cite{Chatzikokolakis2013dx} to obfuscate representation vectors before transmitting them to the server. Furthermore, SAP studied the impact of the FM's split position on the privacy-preserving capacity and the model performance of the SAP framework.  It empirically demonstrated that as the number of layers in the bottom model increases, the utility diminishes while the privacy-preserving capability strengthens, highlighting the existence of a trade-off between utility and privacy when partitioning the FM. SAP is a variant of the typical federated split learning we formulated in Eq.(\ref{eq:fedsplit}) as it conducts the FM adaptation through fine-tuning the model on the server side, as opposed to the client model.


PrivateLoRA~\cite{wang2023prilora} explored the third setting of federated split learning, as illustrated in Figure \ref{fig:setting_1_sl_1_client}(c). Inspired by LoRA, PrivateLoRA represents the update $\Delta W$ on each linear projection of query, key, or value in self-attention with three sequential low-rank matrices $\Delta W = A \times M \times B$. Consequently, there are $\{A\}_{i=1}^K$, $\{B\}_{i=1}^K$, and $\{M\}_{i=1}^K$ for $K$ target linear projections. PrivateLoRA deploys $\{A\}_{i=1}^K$ and $\{B\}_{i=1}^K$ on the server, whereas $\{M\}_{i=1}^K$ on the client in addition to deploying the FM backbone on the server and FM head and bottom models on the client. The $k$th FM middle model in Figure \ref{fig:setting_1_sl_1_client}(c) corresponds to $M_k$. During training, $\{A\}_{i=1}^K$ $\{B\}_{i=1}^K$, and the FM backbone are frozen, while the $\{M\}_{i=1}^K$, FM head, and FM bottom models are optimized. The benefits of PrivateLoRA lie in enhancing the performance of the distributed FM on the client's local task while significantly diminishing communication overheads.

}}

\begin{figure*}[!h]
    \centering
    \includegraphics[width=0.93\linewidth]{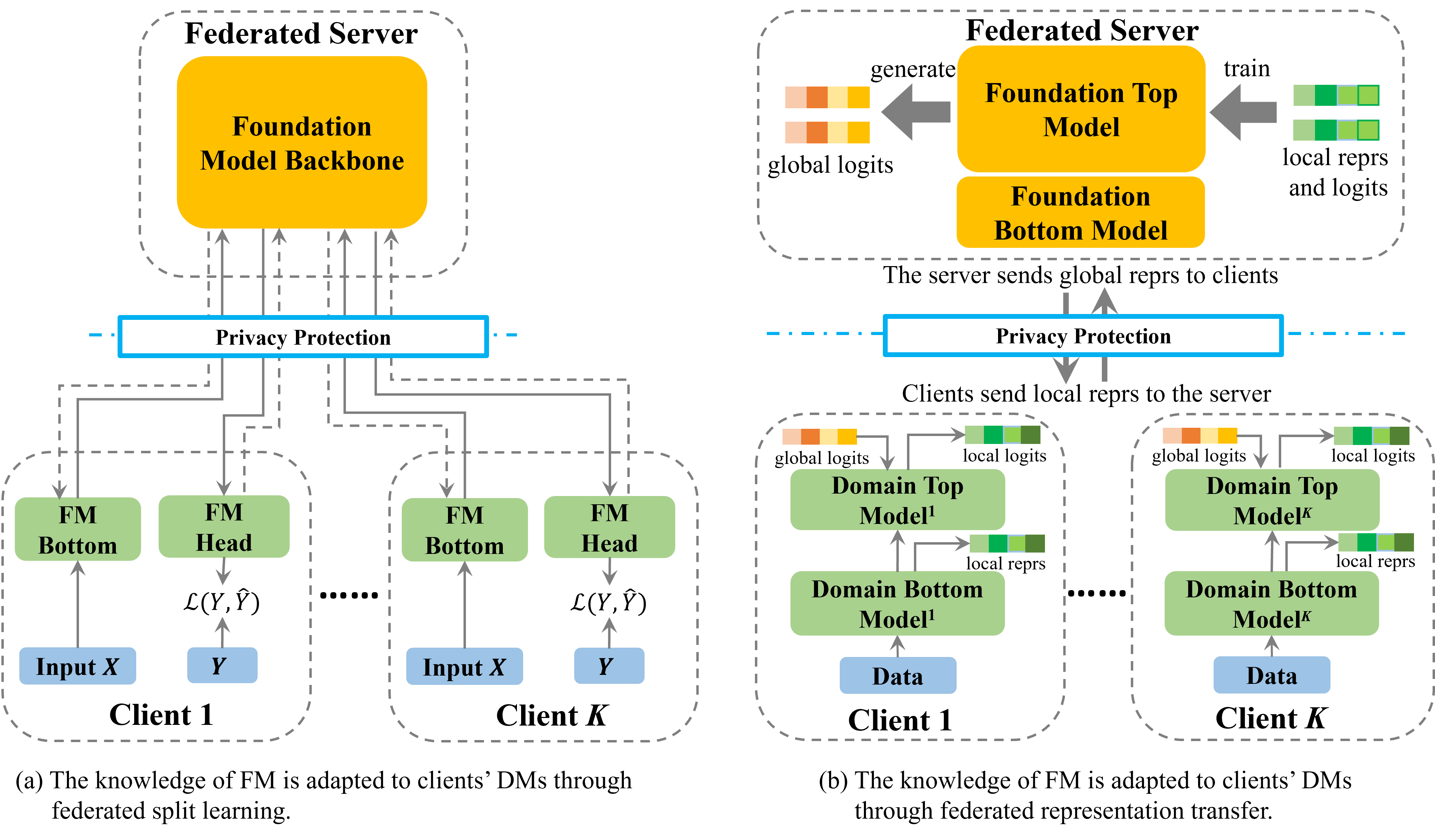}
    \caption{Illustration of adapting knowledge of the server's foundation model (FM) to clients' domain models (DMs) through representation-level knowledge transfer. (a) illustrates that the knowledge of FM is adapted to clients' DMs through federated split learning that involves multiple clients, while (b) through federated representation transfer.}
    \label{fig:setting_1_sl_rt}
\end{figure*}

FedBERT~\cite{Tian2022fedbert} extends the federated split learning with one client to multiple clients, the workflow of which is illustrated in Figure \ref{fig:setting_1_sl_rt}(a). In FedBERT, a pre-specified FM (e.g., ${\text{roberta}}_{\text{base}}$ and GPT2~\cite{radford2019gpt2}) is split into three parts: FM backbone, FM head, and FM bottom. The FM backbone is deployed on the server, while the FM head and FM bottom are deployed on each client. Clients in FedBERT collaborate with the server to train the FM backbone and their corresponding FM bottoms and heads following conventional split learning protocol~\cite{liu2022vertical}. FedBERT adheres to the objective formulation Eq.(\ref{eq:setting_1_stage_2}) with $K>1$, and $\eta=0$. The domain model $\mathcal{W}^{C_k}$ of each client $k$ is composed of an FM head and an FM bottom, and the specific loss of federated split learning $\ell^{C_k}(\mathcal{W}^S_{\text{FM}}, \mathcal{W}^{C_k};d)$ between the server and a client $k$ is formulated in Eq.(\ref{eq:fedsplit}).

Federated split learning enables clients to adapt the knowledge of an FM to their domain-specific tasks while consuming only a small amount of computing resources (i.e., a client only needs to train a small portion of an FM). However, clients have to collaborate with the server to make inferences, which may not apply to scenarios where efficient online inference is critical.

FedGKT~\cite{he2020fedgkt} offers an alternative representation-level transfer approach that adapts the server's large model to clients' DMs. Although FedGKT used ResNet-110 as the server's model, the same wisdom can apply to FMs. According to FedGKT, each DM and the FM are split into a top part and a bottom part. Upon receiving global logits, each client trains its DM based on its local data and global logits. Then, each client generates local representations and logits using its DM and sends them to the server. The server, in turn, trains its foundation top model using received local representations and logits. Next, the server sends global logits generated from the foundation top model to clients for the next round of training. The training process of FedGKT is illustrated in Figure \ref{fig:setting_1_sl_rt}(b). In general, FedGKT adheres to the objective formulated in Eq.(\ref{eq:setting_1_stage_2}) with $K>1$, $\sum^K_{k=1}p^{C_k}=1$, $p^S>0$, and $\eta=0$. The specific loss $\ell^{C_k}(\mathcal{W}^S_{\text{FM}}, \mathcal{W}^{C_k};d)$ of training client $k$'s domain model $\mathcal{W}^{C_k}$ using local data and logits transferred from server's $\mathcal{W}^S_{\text{FM}}$ is formulated in Eq.(\ref{eq:fedkd_1}). 

\subsection{FM adapted to DMs through model-level knowledge transfer} 

\begin{figure*}[!ht]
    \centering
    \includegraphics[width=0.72\linewidth]{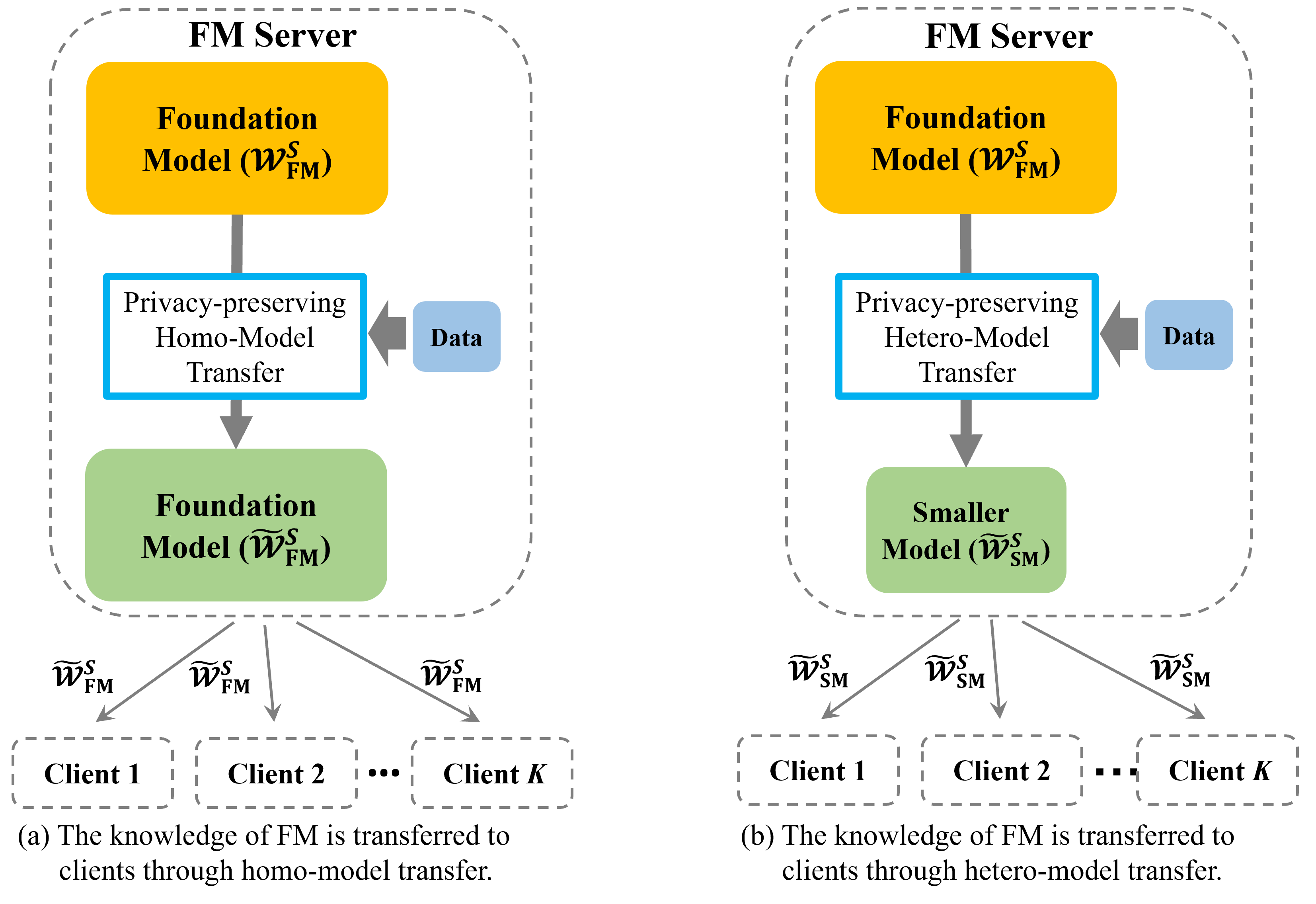}
    \caption{Illustration of transferring knowledge of the server's foundation model (FM) to clients through model transfer. (a) illustrates the knowledge of FM is transferred to downstream clients through homo-model transfer, while (b) through hetero-model transfer.}
    \label{fig:setting_1_model_transfer}
\end{figure*}

Transferring knowledge from FMs in the form of models to downstream clients is an active research direction for grounding FMs. Model transferring methods can be reduced to the formulation of Eq.(\ref{eq:setting_1_stage_1}). We group model transferring methods into two categories. The first category is \textit{homo-model transfer}, which first minimizes loss $\ell^{S}(\mathcal{W}^S_{\text{FM}})$ by fine-tuning or pre-training the FM $\mathcal{W}^S_{\text{FM}}$ and then transfers the resulting model $\mathcal{\widetilde{W}}^S_{\text{FM}}$ to downstream clients. $\mathcal{\widetilde{W}}^S_{\text{FM}}$ has the same architecture as the original FM $\mathcal{W}^S_{\text{FM}}$ (illustrated in Figure \ref{fig:setting_1_model_transfer}(a)). The second category is \textit{hetero-model transfer}, which first minimizes loss $\ell^{S}(\mathcal{W}^S_{\text{FM}})$ by compressing the FM $\mathcal{W}^S_{\text{FM}}$ and then transfer the resulting model $\mathcal{\widetilde{W}}^S_{\text{SM}}$ to downstream clients. $\mathcal{\widetilde{W}}^S_{\text{SM}}$ has a different architecture as the original FM $\mathcal{W}^S_{\text{FM}}$ (illustrated in Figure \ref{fig:setting_1_model_transfer}(b)). A model transferring method involves domain adaptation if $\lambda>0$ and privacy protection if $\eta>0$. We summarize the two categories of model-level knowledge transfer methods in Table \ref{table:server_llm_to_clients_model_transfer} and illustrate them in Figure \ref{fig:setting_1_model_transfer}(a) and Figure \ref{fig:setting_1_model_transfer}(b), respectively. In the following, we will elaborate on these methods of the two categories.

\begin{table*}[!ht]
	\caption{\textbf{Summary of model-level knowledge transfer methods of setting \textcircled{1}}. $\bigcirc$ ($\bigcirc^\ddagger$) denotes that a reference work intends to protect data or model against privacy (backdoor) attacks using certain protection presented in the "How To Protect" columns. DP: Differential Privacy; MP: Model Trained via Proxy Data; PPDS: Privacy-Preserving Data Selection; OWD: Outlier Word Detection; MDP: Masking-Differential Prompting; FP: Fine Pruning. BD: Backdoored; FT: Fine-Tuned; PT: Pre-Trained; $X$, $Y$, and $P$ are input, ground truth response, and prompt, respectively.} 
	\centering
\scriptsize
\bgroup
\def\arraystretch{1.3}
	\begin{tabular}{c|c||c|c|c|c|c|c|c|c}
	    \hline

    	\multirow{3}{*}{\shortstack{Transfer\\Method\\Category}} & \multirow{3}{*}{\shortstack{Reference}} & \multicolumn{4}{c|}{\shortstack{What To Protect}} & \multicolumn{2}{c|}{\shortstack{Exchanged Information}} & \multicolumn{2}{c}{How To Protect}  \\
 
       \cline{3-10}
        
		~ & ~ & \multicolumn{2}{c|}{\shortstack{Server(s)}} & \multicolumn{2}{c|}{\shortstack{Client(s)}} & \multirow{2}{*}{\shortstack{Server(s)\\To Client(s)}} & \multirow{2}{*}{\shortstack{Client(s)\\To Server(s)}} & \multirow{2}{*}{Server} & \multirow{2}{*}{\shortstack{Client}}\\
		\cline{3-6}

		~ & ~ & \tiny\multirow{1}{*}{$\mathcal{D}^S$} & \tiny\multirow{1}{*}{$\mathcal{W}^S_{\text{FM}}$} & \tiny\multirow{1}{*}{$\mathcal{D}^C$} & \tiny\multirow{1}{*}{$\mathcal{W}^C$} & ~ & ~ & ~ & ~ \\
	  \hline
        \hline
       
	\multirow{11}{*}{\shortstack{Homo-\\Model\\Transfer }} &  \multirow{1}{*}{\shortstack{FreD~\cite{hou2023FreD}}} & & & \multirow{1}{*}{$\bigcirc$} & & \multirow{1}{*}{\shortstack{FT $\mathcal{W}^S_{\text{FM}}$}} & Data statistics &  & \multirow{1}{*}{PPDS}  \\
       \cline{2-10}
        ~ & SPT~\cite{yu2023selective} & & & $\bigcirc$ & & FT $\mathcal{W}^S_{\text{FM}}$ & Domain classier & & PPDS \\

         \cline{2-10}
        ~ & EW-Tune~\cite{behnia2022ew}, & \multirow{2}{*}{$\bigcirc$} & & & &  \multirow{2}{*}{\shortstack{FT $\mathcal{W}^S_{\text{FM}}$}} &  &  \multirow{2}{*}{DP} &  \\
        ~ & JFT~\cite{shi2022jft} & & & & & & & & \\

        \cline{2-10}
	   ~ &  DP-BiTFiT~\cite{bu2022dptitfif} & $\bigcirc$ & & & & $\mathcal{W}^S_{\text{FM}}$ with FT bias & & DP &  \\

       \cline{2-10}
        ~ & Youssef et al.~\cite{youssef2023proxydata} & $\bigcirc$ & & &  & Partially PT $\mathcal{W}^S_{\text{FM}}$ & & MP &  \\
       \cline{2-10}
        ~ & NeuBA~\cite{Zhang2021neuba}, & & & & \multirow{2}{*}{$\bigcirc^\ddagger$} & \multirow{2}{*}{BD $\mathcal{W}^S_{\text{FM}}$} & & & \multirow{2}{*}{FP} \\
        ~ & POR~\cite{Shen2021por} & & &  & & & & &  \\
        \cline{2-10}
        ~ & BadPre~\cite{chen2022badpre} & & & & $\bigcirc^\ddagger$ & BD $\mathcal{W}^S_{\text{FM}}$ & & & OWD \\
        \cline{2-10}
	  ~ & Notable~\cite{mei2023notable} & & & & \multirow{1}{*}{\shortstack{$\bigcirc^\ddagger$}} & \multirow{1}{*}{BD $\mathcal{W}^S_{\text{FM}}$} &  & & \multirow{1}{*}{\shortstack{OWD}}   \\
        \cline{2-10}
	  ~ & UOR~\cite{Du2023UORUB} & & & & \multirow{1}{*}{\shortstack{$\bigcirc^\ddagger$}} & \multirow{1}{*}{BD $\mathcal{W}^S_{\text{FM}}$} & & &  \\
   \cline{2-10}
	  ~ & MDP~\cite{xi2023mdp} & & & & \multirow{1}{*}{\shortstack{$\bigcirc^\ddagger$}} & \multirow{1}{*}{BD $\mathcal{W}^S_{\text{FM}}$} & & & MDP  \\

          \hline
        \multirow{9}{*}{\shortstack{Hetero-\\Model\\Transfer}} & DP-Tune~\cite{yu2022differentially} & $\bigcirc$ & & & & Adapter of $\mathcal{W}^S_{\text{FM}}$ & & DP &  \\
    \cline{2-10}
	  ~ & DP-KD~\cite{mire2022dpkd} & $\bigcirc$ & & & & Distilled $\mathcal{W}^S_{\text{FM}}$ & & DP &  \\
   \cline{2-10}
	  ~ & DPIMP~\cite{mire2022dpkd} & $\bigcirc$ & & & & Compressed $\mathcal{W}^S_{\text{FM}}$ & & DP &  \\

      \cline{2-10}
	  ~ &~\citet{wang2023can} & & & $\bigcirc$ & & Distilled $\mathcal{W}^S_{\text{FM}}$ & $\mathcal{W}^C$ & & DP  \\

       \cline{2-10}
        ~ &~\citet{yuan2023m4} & & & & & PEFT-tuned $\mathcal{W}^S_{\text{FM}}$ &  & &  \\

      \cline{2-10}
	  ~ & GPT-FL~\cite{zhang2023gpt} & & & & & \multirow{2}{*}{\shortstack{Model trained on data\\ generated by $\mathcal{W}^S_{\text{FM}}$}} & & &  \\
	  ~ & Feng et al.~\cite{feng2023unlocking} & & & & & & & &   \\
       \cline{2-10}
        ~ & \multirow{1}{*}{\shortstack{BD-FMFL~\cite{li2023backdoor}}} & & & & & \multirow{1}{*}{\shortstack{BD global model}} & & &  \\
       \cline{2-10}
    ~ & BadPrompt~\cite{cai2022badprompt} & & & & \multirow{1}{*}{\shortstack{$\bigcirc^\ddagger$}} & \multirow{1}{*}{BD prompt model} & \multirow{1}{*}{\shortstack{$X+Y+P$}} & & \multirow{1}{*}{\shortstack{FP}}   \\
  \cline{2-10}
        ~ & PPT~\cite{du2022ppt} & & & & \multirow{1}{*}{\shortstack{$\bigcirc^\ddagger$}} & \multirow{1}{*}{BD prompt model} & \multirow{1}{*}{\shortstack{$X+Y+P$}} & &  \\
          \hline   
	\end{tabular}
 \egroup
\label{table:server_llm_to_clients_model_transfer}
\end{table*}

For homo-model transfer without adaptation, JFT~\cite{shi2022jft} proposed a selective differential privacy approach to improve the utility of DP-tuned FMs (i.e., Roberta-base and GPT2-small), which can be released to downstream clients. More specifically, JFT first fine-tunes an FM with the redacted version of a private dataset and then fine-tunes the FM again with the original
private data using an efficient DP-SGD~\cite{li2022ghostclip}. DP-BiTFiT~\cite{bu2022dptitfif} aims to improve the efficiency of fine-tining LLMs with DP-SGD by training only bias terms in an FM. ~\citet{youssef2023proxydata} proposed a privacy-preserving approach to share an FM (i.e., $\text{BERT}_{\text{base}}$) pre-trained on a private dataset. More specifically, they proposed to protect the privacy of the private data by sharing only parts of the FM's pretrained parameters with the rest parameters randomly initialized and releasing a proxy of the private data for downstream clients to fine-tune the FM for complementary knowledge transfer. 

For transferring homo-models adaptive to downstream clients' data, Yu et al.~\cite{yu2023selective} proposed a selective pre-training method (SPT) to pre-train an FM (e.g., $\text{BERT}_{\text{base}}$) based on training data selected by a domain classifier, which is differential-privately trained on the downstream client's private data. Consequently, the client can perform better when initialized with the FM pretrained on adaptively selected data than on randomly selected data. FreD~\cite{hou2023FreD} was proposed to select the data used to fine-tune an FM (i.e., DistilGPT-2) for initializing models of clients. The data selection is performed based on the statistical information of clients' private data that is collected leveraging a DP-based federated learning algorithm. In addition to adaptation by data selection, ~\citet{yuan2023m4} proposed to build a one-size-fits-all mobile foundation model termed M4 for diversified mobile AI tasks. Specifically, M4 is built on pre-trained and off-the-shelf FMs (e.g., LLaMA-7B) and exposes three learnable parts (e.g., LoRA) to be fine-tuned for downstream mobile AI tasks. Clients have the same task, say image captioning, can deploy the M4 model with learnable parameters trained on image captioning datasets.  

For hetero-model transfer without adaptation, DP-Tune~\cite{yu2022differentially} was proposed to differential-privately fine-tune an additional small set of parameters on top of pre-trained FMs (e.g., GPT-2-XL) using a private dataset so that the fine-tuned new parameters can preserve the privacy of the private dataset when releasing to downstream tasks. ~\citet{mire2022dpkd} proposed privacy-preserving knowledge distillation (DP-KD) and pruning (DPIMP) methods based on DP-SGD to compress a pre-trained FM (e.g., $\text{BERT}_{\text{base}}$) to a smaller model, which can meet the memory and latency requirements of specific downstream applications. 

For transferring hetero-models adaptive to downstream tasks, ~\citet{wang2023can} proposed an approach that leverages public data and pre-trained FM (i.e., LaMDA~\cite{Aaron2022lamda}) to enhance downstream cross-device federated learning. More specifically, they first leverage their proposed distribution matching algorithm to sample public data closely resembling the distribution of FL clients' private data, and then they use the sampled data to distill a pre-trained FM (i.e., LaMDA) into an on-device language model, which serves as an initialization for FL clients. GPT-FL~\cite{zhang2023gpt} leverages FMs (i.e., Stable Diffusion~\cite{rombach2022high}, SpeechT5~\cite{ao2022speecht5}, and AudioLDM~\cite{liu2023audioldm}) to generate diversified synthetic data to train a downstream model, which is then distributed to FL clients for model initialization.

\subsection{Privacy and Backdoor Attacks}

In the setting \textcircled{1}, the server is the potential adversary, who can launch privacy attacks with the intention of inferring clients' private information. The server can also carry out backdoor attacks to compromise the utility of clients' domain models.

Privacy attacks focus on inferring private data from knowledge at the representation level. More specifically, TextObfuscator~\cite{zhou2023textobfuscator}, RAPT~\cite{li2023privacy} and \citet{xu2023shuffled} investigated model inversion attacks in federated split learning scenarios, in which the server aims to reconstruct the client's private data based on the representations output from the client's local model. The three studies proposed representation perturbation, $d_{\chi}$-privacy, and weight permutation equivalence, respectively, to thwart model inversion attacks. The rationale behind these defenses is to reduce the correlation between private data and the representations shared with the server. As a result, the chance for the adversary to recover private data can be reduced. 

The backdoor attacks are typically conducted through model transfer. In the case of homo-model transfer, the server injects backdoors into clean foundation models to be released. This line of works, including BadPre~\cite{chen2022badpre}, POR~\cite{Shen2021por}, Notable~\cite{mei2023notable}, NeuBA~\cite{Zhang2021neuba}, UOR~\cite{Du2023UORUB}, and MDP~\cite{xi2023mdp}, follows a similar pipeline: first, a poisoning dataset is prepared; next, a clean foundation model is fine-tuned using a combination of clean and poisoning data; finally, the backdoored foundation model is distributed to downstream clients. During inference, the server can activate the backdoor embedded in the client's local model by making specific queries. NeuBA and POR leverage Fine-Pruning (FP)~\cite{liu2018fp}, while BadPre and Notable utilize Outlier Word Detection (OWD)~\cite{qi2021onion} to defend against the backdoor attack, showing a trade-off between the backdoor’ effectiveness and the model’s accuracy. MDP, short for Masking-Differential Prompting, is a backdoor defense proposed by \citet{xi2023mdp}. It identifies poisoned samples by comparing the representations of given samples under varying masking. The samples with significant variations are poisoned.

Regarding hetero-model transfer,  \citet{li2023backdoor} proposed BD-FMFL, a backdoor attack following the framework proposed in GPT-FL~\cite{zhang2023gpt}. BD-FMFL works as follows: the attacker inserts backdoors into a Language Model with Large Memory (LLM) by leveraging in-context learning. Subsequently, the server utilizes the backdoored LLM to generate prompts, which are then used to synthesize data using other FMs. The server proceeds to train a global model based on the synthetic data, thereby transferring the backdoor to the global model. Once accomplished, the backdoored global model is fine-tuned using private datasets provided by clients within the standard Federated Learning (FL) framework. Another avenue of research, explored by BadPrompt~\cite{cai2022badprompt} and PPT~\cite{du2022ppt}, focuses on prompt-based backdoor attacks, in which the attacker injects backdoors into the prompt model of an FM and subsequently releases the poisoned prompt model to the public. Upon a client downloading and utilizing the compromised prompt model for downstream tasks, the server gains the ability to activate the backdoor. BadPrompt suggests that methods such as Fine-Pruning (FP) and knowledge distillation could potentially defend against prompt-based backdoor attacks.

\section{Domain Knowledge for Augmenting Foundation Models}\label{sec:dm_to_fm}

In this section, we overview FTL-FM methods that fall into setting \textcircled{2} (see Definition \ref{def:ftl-llm}), the objective of which is to augment the FM with industry-level and domain-specific knowledge (e.g., healthcare, medicine, finance, banking, and law) transferred from clients.

\begin{table*}[!ht]
	\caption{\textbf{Summary of FTL-FM works of setting \textcircled{2}}. $\bigcirc$ ($\bigcirc^\dagger$) denotes that a reference work intends to protect data or model against privacy (backdoor) attacks using certain protection methods presented in the "How To Protect" columns. CP: Compression; SA: Secure Aggregation; repr: Representation.} 
	\centering
\scriptsize
\bgroup
\def\arraystretch{1.3}
	\begin{tabular}{c|c||c|c|c|c|c|c|c|c}
	    \hline

    	\multirow{3}{*}{\shortstack{Transfer\\Method\\Category}} & \multirow{3}{*}{\shortstack{Reference}} & \multicolumn{4}{c|}{\shortstack{What To Protect}} & \multicolumn{2}{c|}{\shortstack{Exchanged Information}} & \multicolumn{2}{c}{How To Protect}  \\
 
       \cline{3-10}
        
		~ & ~ & \multicolumn{2}{c|}{\shortstack{Client(s)}} & \multicolumn{2}{c|}{\shortstack{Server(s)}} & \multirow{2}{*}{\shortstack{Client(s)\\To Server(s)}} & \multirow{2}{*}{\shortstack{Server(s)\\To Client(s)}} & \multirow{2}{*}{ Client} & \multirow{2}{*}{\shortstack{Server}}\\
		\cline{3-6}

		~ & ~ & \tiny\multirow{1}{*}{$\mathcal{D}^C$} &  \tiny\multirow{1}{*}{$\mathcal{W}^C$} &  \tiny\multirow{1}{*}{$\mathcal{D}^S$} &  \tiny\multirow{1}{*}{$\mathcal{W}^S_{\text{FM}}$} & ~ & ~ & ~ & ~ \\
	  \hline
        \hline


        \multirow{2}{*}{\shortstack{Repr \\Transfer}} & \multirow{2}{*}{CreamFL~\cite{yu2023multimodal}} &  &  & & & \multirow{2}{*}{Local repr} & \multirow{2}{*}{\shortstack{Global repr generated\\ by $\mathcal{W}^S_{\text{FM}}$}} &  & \\
       & & & & & & & & & \\
       \hline

	   \multirow{7}{*}{\shortstack{Hetero-\\Model\\Transfer}} &  Offsite-Tuning~\cite{xiao2023offsite} & & & & $\bigcirc$$^\dagger$ & Adapter & Emulator of $\mathcal{W}^S_{\text{FM}}$ & & CP \\
       \cline{2-10}
        ~ & \multirow{2}{*}{FedOST~\cite{fedost,fan2023fate}} & \multirow{2}{*}{$\bigcirc$} & & & \multirow{2}{*}{$\bigcirc$$^\dagger$} & \multirow{2}{*}{Adapter} & \multirow{2}{*}{\shortstack{Emulator of $\mathcal{W}^S_{\text{FM}}$,\\Initial adapter}} & \multirow{2}{*}{SA} & \multirow{2}{*}{CP}  \\
        & & & & & & & & &\\
       \cline{2-10}
        ~ & \multirow{2}{*}{FedOT~\cite{kuang2023fs-llm}}& & & & \multirow{2}{*}{$\bigcirc$$^\dagger$} & \multirow{2}{*}{Adapter} & \multirow{2}{*}{\shortstack{Emulator of $\mathcal{W}^S_{\text{FM}}$,\\ Initial adapter}} & & \multirow{2}{*}{CP}  \\
        &  & & & & & & & &\\
       \cline{2-10}
        ~ & \multirow{1}{*}{FedPEAT~\cite{chua2023fedpeat}}& & & & \multirow{2}{*}{$\bigcirc$$^\dagger$} & \multirow{2}{*}{Adapter} & \multirow{2}{*}{\shortstack{Emulator of $\mathcal{W}^S_{\text{FM}}$,\\ Global adapter}} & & \multirow{2}{*}{CP}  \\
        ~ & \multirow{1}{*}{CEFHRI~\cite{khalid2023cefhri}}& & & & & & &  \\
        \hline
	\end{tabular}
 \egroup
\label{table:clients_to_server_llm}
\end{table*}

The key to augmenting FMs to be more knowledgeable, accurate, and reliable lies in fine-tuning these FMs using high-quality domain-specific datasets. These high-quality datasets may contain private and sensitive information and are often scattered among isolated entities (devices and companies). FTL-FM methods enable the server of FM and clients with domain knowledge to augment the FM collaboratively while preserving the privacy of clients' knowledge. In literature, \textit{representation-level knowledge transfer} and \textit{model-level knowledge transfer} are the main research directions explored to augment an FM hosted by a server with knowledge transferred from domain-specific clients. We summarize FTL-FM works that fall into this setting in Table \ref{table:clients_to_server_llm}, and in the following subsections, we will elaborate on these works.

\begin{figure*}[!ht]
    \centering
    \includegraphics[width=0.92\linewidth]{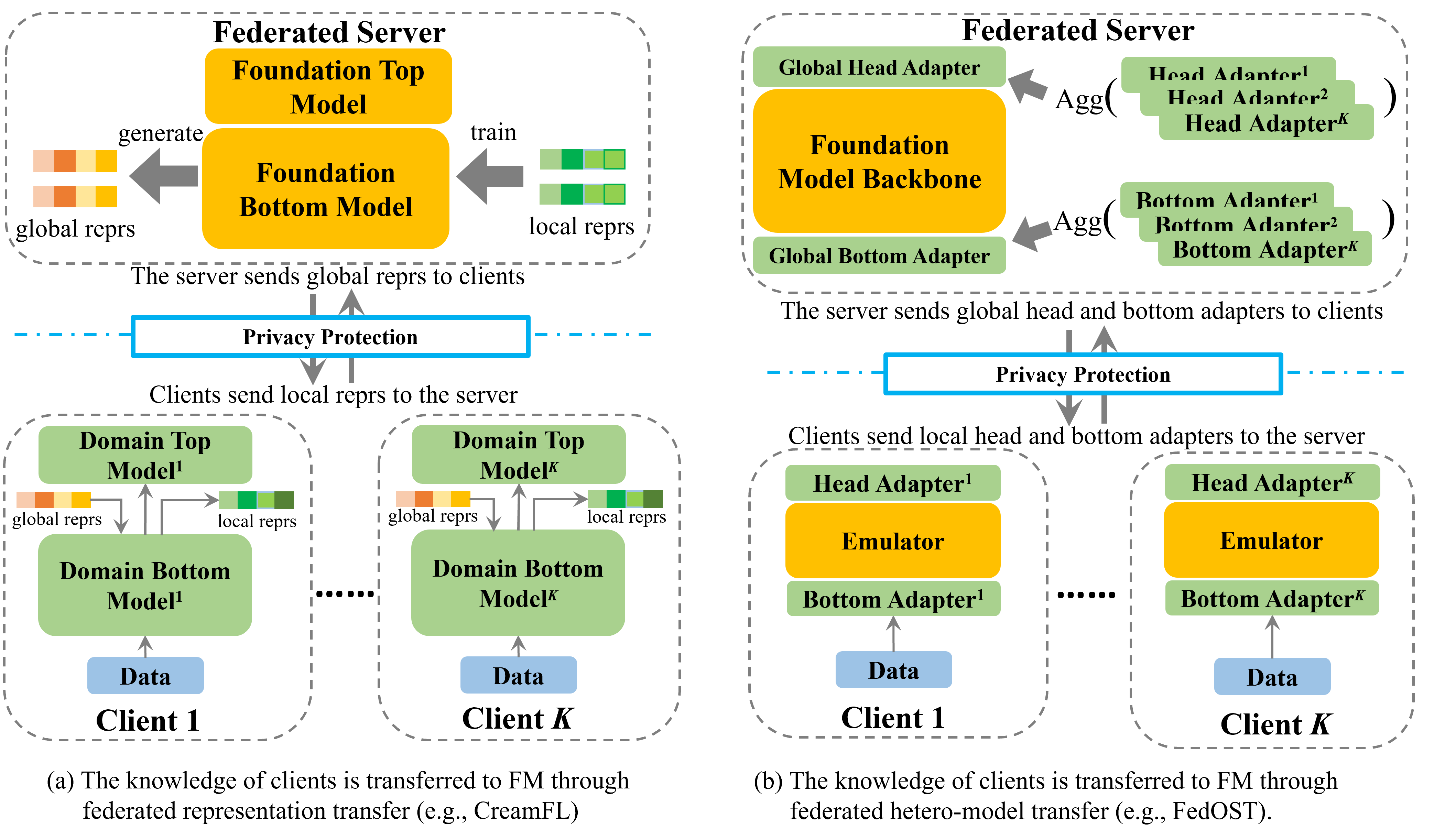}
    \caption{Illustration of transferring domain-specific knowledge of clients to the server's foundation model (FM) through (a) representation transfer and (b) model transfer.}
    \label{fig:clients_to_server_llm}
\end{figure*}

\subsection{Domain knowledge transferred to FM through representation-level knowledge transfer}

The main goal of fine-tuning FMs with clients' domain-specific knowledge through federated learning is to obtain FMs that can generalize to (multiple or many) domain-specific tasks with high accuracy and reliability. 

Toward this end, CreamFL~\cite{yu2023multimodal} was proposed to learn an FM (i.e., $\text{BERT}_{\text{base}}$) by transferring representations from diverse clients that own uni- and multi-modal data. In CreamFL, each DM and the FM are split into top and bottom parts. Upon receiving global representations, each client first trains its DM based on its local data and global representations. Then, each client generates local representations using its domain bottom model based on a public dataset and sends local representations to the server. The server, in turn, aggregates local representations sent from clients with global representations and trains the foundation bottom model through knowledge distillation based on aggregated representations. Next, the server sends global representations generated from the FM to clients for the next round of training. The workflow of CreamFL is illustrated in Figure \ref{fig:clients_to_server_llm}(a). CreamFL generally adheres to the objective formulated in Eq.(\ref{eq:setting_2}) with $K>1$, $\sum^K_{k=1}p^{C_k}=1$, $p^S>0$, and $\eta=0$. The specific form of the loss $\ell^{C_k}(\mathcal{W}^S_{\text{FM}}, \mathcal{W}^{C_k};d)$ for transferring representations generated by $\mathcal{W}^{C_k}$ to train $\mathcal{W}^S_{\text{FM}}$ is formulated in Eq.(\ref{eq:fedkd_2}). 

\subsection{Domain knowledge transferred to FM through model-level knowledge transfer}

While representation transfer mitigates privacy leakage by retaining private data locally, it still suffers from privacy risks because it sends sample-wise representations of private data to the server. Therefore, the risks of leaking data privacy by revealing these representations need to be carefully investigated. An alternative way to transfer clients' domain knowledge to the server's FM is through model transfer.

Offsite-Tuning~\cite{xiao2023offsite} is the representative work that leverages model transfer combined with parameter-efficient fine-tuning to transfer the knowledge of a downstream client to the FMs (i.e., BLOOM~\cite{scao2022bloom} and OPT~\cite{zhang2022opt}) owned by a server while protecting both the ownership of the server's FM and the privacy of the client's private data. In Offsite-Tuning, the server first selects a small subset of its FM as the adapter and compresses its FM into an emulator, and then it sends the adapter and the emulator to the downstream client. The client, in turn, fine-tunes the adapter on its private data with the assistance of the emulator. The fine-tuned adapter is then returned and plugged into the FM to create an adapted FM. As the server and the client do not share their full FM and private data, respectively, Offsite-Tuning has the potential to protect both the privacy of the client's data and ownership of the server's model. Offsite-Tuning~\cite{xiao2023offsite} corresponds to the objective formulated in Eq.(\ref{eq:setting_2}) with $K=1, p^{C_1}=1, p^S=0,$ and $\eta=0$. The client has no DM but trains two adaptors sent from the server with the help of an emulator compressed from $\mathcal{W}^S_{\text{FM}}$. The specific form of the Offsite-Tuning loss is formulated in Eq.(\ref{eq:fedost}). 


While effective, Offsite-Tuning exhibits two limitations: (1) Offsite-Tuning does not provide protection methods to explicitly protect data privacy, and thus, the client's private data can still be reconstructed by an adversary when observing the adapter fine-tuned using the private data~\cite{morris2023text}; (2) Offsite-Tuning involves only one client.

To alleviate these limitations, Fan et al.~\cite{fan2023fate,fedost} proposed a federated learning version of Offsite-Tuning named FedOST. In FedOST, the server first distributes the emulator and two adapters of its FM to all clients at the beginning of FL. Then, each client fine-tunes the two adapters on its private data with the assistance of the emulator. Next, all clients send their adapters to the server, which aggregates received adapters through secure aggregation~\cite{Bonawitz2016secagg} and plugs the aggregated adapter into the FM. FedOST protects the privacy of clients' private data and can build a fine-tuned FM with enhanced generalization capability. With similar motivations as FedOST, Kuang et al. integrated Offsite-Tuning into FederatedScope-LLM~\cite{kuang2023federatedscope} and termed it FedOT. FedPEAT~\cite{chua2023fedpeat} generalized the Offsite-Tuning approach to federated learning with multiple communication rounds and proposed an adaptive control mechanism to facilitate the adoption of FedPEAT in a dynamic environment. CEFHRI~\cite{khalid2023cefhri} adopted Offsite-Tuning to the human-robot interaction domain. 

Similar to Offsite-Tuning, FedOST, FedOT, and  FedPEAT adhere to the objective formulated in Eq.(\ref{eq:setting_2}) but with $K>1$, $\sum^K_{k=1}p^{C_k}=1$, and $p^S=0$. FedOST has $\eta>0$ since it explicitly leverages secure aggregation to protect data privacy. In general, the workflow of the three methods can be illustrated in Figure \ref{fig:clients_to_server_llm}(b).

\subsection{Privacy and Backdoor Attacks}

During the process of transferring domain-specific knowledge from clients to the server, the server can potentially infer the private data of clients from its observed information, while a client can poison the server's FM by injecting backdoors into the information transmitted to the server. Nonetheless, privacy and backdoor attacks are rarely explored in this setting. In this subsection, we discuss potential threats to data privacy and model utility as well as possible countermeasures based on existing works. 

In the Offsite-Tuning~\cite{xiao2023offsite}, the server can potentially reconstruct the client's private data through the model inversion attack or infer sensitive attributes through the attribute inference attack based on the adapters sent from the client. DP-SGD~\cite{abadi2016dpsgd} and its variants~\cite{kairouz2021dpftrl} are straightforward protection mechanisms against these attacks. When more than two clients are involved in the offsite tuning, secure aggregation can be employed to mitigate the privacy vulnerability of clients' private data, as studied in FedOST~\cite{fedost}. 

In CreamFL~\cite{wang2022multimodal}, the server can possibly infer the private data of clients by investigating representations sent from clients. \citet{takahashi2023pli} undertook a preliminary investigation into the privacy vulnerabilities of federated model distillation approaches that transfer knowledge through logits (i.e., unnormalized probabilities of an instance belonging to a certain class). They proposed a Paired-Logits Inversion (PLI) attack to infer clients' private data based on logits from a server model and those from the client model. While PLI relies on logits to carry out the attack, its rationale is worth verifying in inferring private data from representations. To mitigate the chance for the adversary to infer private data from logits or representations, a promising research direction is to reduce the dependence between the private data and the logits or representations shared with the adversary while maintaining the utility. Along this direction, obfuscation mechanisms, such as Adaptive Obfuscation~\cite{gu2023fedpass} and Mutual Information Regularization~\cite{zou2023mid}, are worth exploring.

On the other hand, the server is subject to backdoor attacks mounted by clients. For example, the client in the Offsite-Tuning~\cite{xiao2023offsite} can possibly poison the server's FM by encoding backdoors into the adapters that will be plugged into the server's FM. The client in CreamFL~\cite{wang2022multimodal} can embed backdoor triggers into the representations, which in turn transfer the triggers to the server's FM through fine-tuning. The backdoor attacks through adapters and representations are rarely explored in literature~\cite{nguyen2024bdsurvey}, thereby deserving investigation.

\section{Co-optimize Foundation Models and Domain Models}\label{sec:fm_dm_coevolve}


The FTL-FM settings we discussed in Section~\ref {sec:fm_to_dm} and Section~\ref{sec:dm_to_fm} focus on adapting general knowledge of the FM to optimize DMs, and transferring domain-specific knowledge of DMs to optimize FM, respectively. In this section, we explore the setting \textcircled{3}, which endeavors to mutually optimize FMs and DMs hosted by different parties (i.e., the server and clients). 

Optimizing both FMs and DMs helps establish a positive cycle to evolve FMs and DMs~\cite{coevolve} continuously. For example, an FM server can first deliver the general knowledge and abilities of its FM to clients' DMs, which are then trained on domain-specific data for downstream applications. Clients' DMs, in turn, transfer industry-specific knowledge to enhance the server's FMs. This loop can go beyond one training task and continues over time.  Nevertheless, mutually enhancing the server's FM and clients' DMs is rarely exploited in literature. A more commonly studied scenario involves clients collaboratively fine-tuning their domain models that are initialized with (or assisted by) FMs. We summarize the two scenarios as follows. 
\begin{enumerate}
    \item Co-optimize server's FM and clients' DMs. The server ends up with an FM augmented with domain-specific knowledge, and each client with a DM enhanced with general knowledge of the server's FM. Note that, in this scenario, clients' DM can be optionally initialized with FMs having sizes much smaller than the server's FM.
    \item Optimize clients' domain-specific FMs. The server serves no FM but acts as a model aggregation function, while clients own DMs that are initialized with or assisted by FMs. The objective of this scenario is to optimize clients' DMs through conventional horizontal federated learning. Each client ends up with a DM enhanced with the domain knowledge of all clients. 
\end{enumerate}

The FTL-FM methods proposed in the two scenarios almost focus on model-level knowledge transfer. We review FTL-FM methods of both scenarios in Section \ref{sec:co_optim} and Section \ref{sec:domain-fm}, respectively.

\subsection{Co-optimize FMs and DMs through model-level knowledge transfer} \label{sec:co_optim}

FedCoLLM~\cite{fan2023fate} and CrossLM~\cite{deng2023crosslm} recently proposed in the literature fall into the first scenario. We summarize them in Table \ref{table:coevolve-fm-dm} and elucidate them as follows.

\begin{table*}[!ht]
	\caption{\textbf{Summary of FTL-FM works for co-optimizing FMs and DMs through model-level knowledge transfer of setting \textcircled{3}}. $\bigcirc$ denotes that a reference work intends to protect data against privacy attacks using certain protection methods presented in the "How To Protect" columns. SA: Secure Aggregation. } 
	\centering
\scriptsize
\bgroup
\def\arraystretch{1.3}
  \setlength{\tabcolsep}{3.2pt}
	\begin{tabular}{c|c||c|c|c|c|c|c|c|c}
	    \hline

    	\multirow{3}{*}{\shortstack{Transfer\\Method\\Category}} & \multirow{3}{*}{\shortstack{Reference}} & \multicolumn{4}{c|}{\shortstack{What To Protect}} & \multicolumn{2}{c|}{\shortstack{Exchanged Information}} & \multicolumn{2}{c}{How To Protect}  \\
 
        \cline{3-10}
        
		 &  & \multicolumn{2}{c|}{\shortstack{Client(s)}} & \multicolumn{2}{c|}{\shortstack{Server(s)}} & \multirow{2}{*}{\shortstack{Client(s)\\To Server(s)}} & \multirow{2}{*}{\shortstack{Server(s)\\To Client(s)}} & \multirow{2}{*}{ Client} & \multirow{2}{*}{\shortstack{Server}}\\
		\cline{3-6}

		~ & ~ & \tiny \multirow{1}{*}{$\mathcal{D}^C$} & \tiny \multirow{1}{*}{$\mathcal{W}^C$} & \tiny \multirow{1}{*}{$\mathcal{D}^S$} & \tiny \multirow{1}{*}{$\mathcal{W}^S_{\text{FM}}$} & ~ & ~ & ~ & ~ \\
	  \hline
        \hline
       \multirow{2}{*}{\shortstack{Homo-Model\\Transfer }} & \multirow{2}{*}{\shortstack{FedCoLLM~\cite{fan2023fate}}} & \multirow{2}{*}{\shortstack{$\bigcirc$}} & & & & \multirow{2}{*}{\shortstack{$\mathcal{W}^C$}} & \multirow{2}{*}{\shortstack{Global model\\enhanced by $\mathcal{W}^S_{\text{FM}}$}} &  \multirow{2}{*}{\shortstack{SA}} &  \\
    ~ & & & & & & & & &   \\
          
        \hline
	
       \multirow{2}{*}{\shortstack{Hetero-Model\\Transfer}} & \multirow{2}{*}{\shortstack{CrossLM~\cite{deng2023crosslm}}} & & & & & \multirow{2}{*}{\shortstack{$\mathcal{W}^C$}} & \multirow{2}{*}{\shortstack{$\mathcal{W}^C$ enhanced by $\mathcal{W}^S_{\text{FM}}$}} &  &  \\
         ~ & & & & & & & & &   \\

    \hline
	\end{tabular}
\egroup
\label{table:coevolve-fm-dm}
\end{table*}

\begin{figure*}[t!]
    \centering
\includegraphics[width=0.99\linewidth]{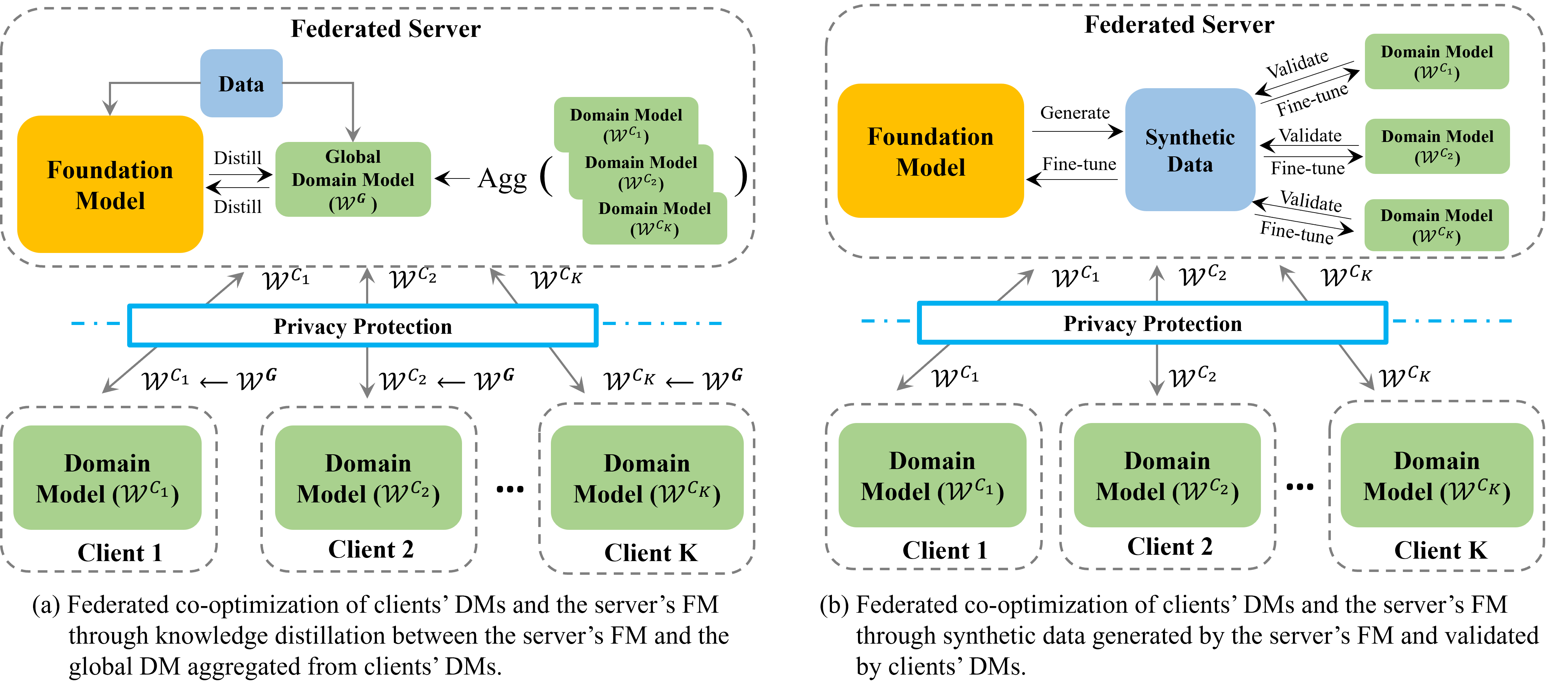}
    \caption{Illustration of co-optimizing server's FM and clients' DMs through transferring model-level knowledge among participating parties (i.e., server and clients). (a) federated co-optimization is achieved through distilling knowledge between the server's FM and the global DM aggregated from clients' DMs. (b) federated co-optimization is achieved through fine-tuning the server's FM and clients' DMs using the synthetic data generated by the server's FM and validated by clients' DMs.}
    \label{fig:co_tuning}
\end{figure*}

FedCoLLM (Federated Co-tuning LLM) was proposed in FATE-LLM~\cite{fan2023fate} (Federated Co-tuning LLM). It co-optimizes the server's FM and clients' DMs simultaneously through mutual knowledge distillation. Figure \ref{fig:co_tuning}(a) illustrates the workflow of FedCoLLM. In FedCoLLM, clients' DMs are initialized with an off-the-shelf FM (i.e., LLaMA-7B~\cite{touvron2023llama}). In each communication round, clients fine-tune their DMs using local data and then send them to the server with secure aggregation~\cite{Bonawitz2016secagg}. The server, in turn, distills the knowledge mutually between its hosted FM (i.e., LLaMa-65B) and the global DM aggregated from clients' local DMs. Subsequently, the server dispatches the global DM to all clients for further training. FedCoLLM adheres to the objective formulated in Eq.(\ref{eq:setting_3}) with $K>1$, $\sum^K_{k=1}p^{C_k}=1$, $p^S=0$, and $\eta>0$. The specific form of the loss $\ell^{C_k}(\mathcal{W}^S_{\text{FM}}, \mathcal{W}^{C_k};d)$ for co-optimizing the server's $\mathcal{W}^S_{\text{FM}}$ and the global DM $\mathcal{W}^{G}$ shared by all clients is formulated in Eq.(\ref{eq:fedcotuning}).

CrossLM was proposed by ~\citet{deng2023crosslm}. It optimizes the server's FM and clients' DMs using synthetic data generated by the server's FM and validated by clients' DMs, as illustrated in Figure \ref{fig:co_tuning}(b). Specifically, the DM on each client is randomly determined as BERT-Base or DistilBERT. In each communication round, clients train their DMs using local data and then send them to the server asynchronously. The server, in turn, generates synthetic data and leverages received DMs to validate the quality of synthetic data. Subsequently, the server utilizes validation feedback from DMs and the validated synthetic data to fine-tune the server's FM. Next, the server uses the validated synthetic data to fine-tune each client's DM and dispatches the fine-tuned DMs to corresponding clients for further training. This concerted effort is aimed at co-optimizing both the FM and DMs. CrossLM adhere to the objective formulated in Eq.(\ref{eq:setting_3}) with $K>1$, $\sum^K_{k=1}p^{C_k}=1$, $p^S=1$, and $\eta=0$. The specific form of the tasks loss $\ell^{C_k}(\mathcal{W}^S_{\text{FM}}, \mathcal{W}^{C_k};d)$ for co-optimizing the server's $\mathcal{W}^S_{\text{FM}}$ and client $k$'s $\mathcal{W}^{C_k}, k=1,\dots,K$ is formulated in Eq.(\ref{eq:fedcotuning}).


In addition to the model transfer methods, the representation transfer method proposed by CreamFL~\cite{yu2023multimodal} (illustrated in Figure \ref{fig:clients_to_server_llm}(a)) also has the potential to achieve co-optimization since representations can convey general knowledge and contextual information of a model and thus can help improve the generalization capability of both FMs and DMs. 

\subsection{Optimize Domain-Specific FMs through model-level knowledge transfer} \label{sec:domain-fm}

A considerable amount of FTL-FM research has been proposed in the second scenario, which essentially falls under the purview of horizontal federated learning~\cite{mcmahan2017communication}. FTL-FM works in this scenario achieve knowledge transfer typically through either homo-model transfer or hetero-model transfer. The homoe-model transfer requires all clients to initialize their DMs with the same FM and conduct knowledge transfer by sharing their full DMs with the server for aggregation. On the other hand, hetero-model transfer enables clients to initialize their DMs with FMs having different architectures and conduct knowledge transfer by sharing partial or proxies of their DMs. We summarize the two categories of FTL-FM works in Table \ref{table:optim-ds-fm} and elucidate them as follows.

\begin{table*}[!ht]
	\caption{\textbf{Summary of FTL-FM works for optimizing domain-specific FMs of setting \textcircled{3}}. $\bigcirc$ ($\bigcirc^\ddagger$) denotes that a reference work intends to protect data or model against privacy (backdoor) attacks using certain protection methods presented in the "How To Protect" columns. HE: Homomorphic Encryption; FE: Freeze Embedding; DP: Differential Privacy; NC: Norm Clipping; PEFT: Parameter-Efficient Fine-Tuning; SA: Secure Aggregation; SCR: Scrubbing; OTR: One Time Pad; TEE: Trusted Execution Environment. } 
	\centering
\scriptsize
\bgroup
\def\arraystretch{1.3}
  \setlength{\tabcolsep}{3.2pt}
	\begin{tabular}{c|c||c|c|c|c|c|c|c|c}
	    \hline

    	\multirow{3}{*}{\shortstack{Transfer\\Method\\Category}} & \multirow{3}{*}{\shortstack{Reference}} & \multicolumn{4}{c|}{\shortstack{What To Protect}} & \multicolumn{2}{c|}{\shortstack{Exchanged Information}} & \multicolumn{2}{c}{How To Protect}  \\
 
        \cline{3-10}
        
		 &  & \multicolumn{2}{c|}{\shortstack{Client(s)}} & \multicolumn{2}{c|}{\shortstack{Server(s)}} & \multirow{2}{*}{\shortstack{Client(s)\\To Server(s)}} & \multirow{2}{*}{\shortstack{Server(s)\\To Client(s)}} & \multirow{2}{*}{ Client} & \multirow{2}{*}{\shortstack{Server}}\\
		\cline{3-6}

		~ & ~ & \tiny \multirow{1}{*}{$\mathcal{D}^C$} & \tiny \multirow{1}{*}{$\mathcal{W}^C$} & \tiny \multirow{1}{*}{$\mathcal{D}^S$} & \tiny \multirow{1}{*}{$\mathcal{W}^S_{\text{FM}}$} & ~ & ~ & ~ & ~ \\
	  \hline
        \hline
       \multirow{13}{*}{\shortstack{Homo-\\Model\\Transfer }} &  \multirow{2}{*}{\shortstack{FedHE~\cite{jin2023fedhe}}} & \multirow{2}{*}{\shortstack{$\bigcirc$}} & & & & \multirow{2}{*}{\shortstack{Partially HE-\\encrypted $\mathcal{W}^C$}} & \multirow{2}{*}{\shortstack{Global model}} & \multirow{2}{*}{\shortstack{HE}} & \\
          & & & & & & & & \\
       \cline{2-10}
       ~ & \multirow{1}{*}{\shortstack{FL4ASR~\cite{azam2023fl4asr}}} & \multirow{1}{*}{\shortstack{$\bigcirc$}} & & & & \multirow{1}{*}{\shortstack{DP-trained $\mathcal{W}^C$}} & \multirow{1}{*}{\shortstack{Global model}} & \multirow{1}{*}{\shortstack{DP}} & \\
       \cline{2-10}
	 ~ &  \multirow{1}{*}{\shortstack{FILM~\cite{gupta2022recovering}}} & $\bigcirc$ & & & & Gradients of $\mathcal{W}^C$ & Global model & FE &   \\
       \cline{2-10}
        ~ & LAMP~\cite{balunovic2022lamp} & $\bigcirc$ & & & & Gradients of $\mathcal{W}^C$ & Global model & DP & \\

        \cline{2-10}
	   ~ &  \multirow{2}{*}{\shortstack{Decepticons\cite{fowl2023decepticons},\\Panning\cite{chu2023panning}}} & \multirow{2}{*}{\shortstack{$\bigcirc$}} & & & & \multirow{2}{*}{\shortstack{Gradients of $\mathcal{W}^C$}} & \multirow{2}{*}{\shortstack{Maliciously updated\\global model}} & \multirow{2}{*}{\shortstack{DP}} &  \\
        & & & & & & & & \\


       \cline{2-10}
        ~ & \multirow{2}{*}{\shortstack{FLTrojan~\cite{rashid2023fltrojan}}} & \multirow{2}{*}{$\bigcirc$} & & &  & \multirow{2}{*}{\shortstack{Maliciously updated \\$\mathcal{W}^C$ and benign $\mathcal{W}^C$}}  & \multirow{2}{*}{\shortstack{Global model}} &  \multirow{2}{*}{\shortstack{DP \& \\SCR}} &   \\
        & & & & & & & & \\
         
       \cline{2-10}
        ~ & Neurotoxin~\cite{zhang2022neurotoxin} & & & & \multirow{1}{*}{$\bigcirc^\ddagger$} & Backdoored $\mathcal{W}^C$ & Global model & & \multirow{1}{*}{NC \& DP} \\
       \cline{2-10}
        ~ & RE-GE~\cite{yoo2022backdoor} & & & & $\bigcirc^\ddagger$ & Backdoored $\mathcal{W}^C$ & Global model & & \multirow{1}{*}{NC \& DP} \\
       \cline{2-10}
       
        ~ & FedNLP~\cite{lin2022fednlp} & & & & & Updates of DMs & Global model & & \\
      \cline{2-10}
         

    \cline{2-10}
    
        ~ & \multirow{1}{*}{\shortstack{FedLLM-on-Edge~\cite{woi2023fededge}}} & & & & & \multirow{1}{*}{\shortstack{$\mathcal{W}^C$}} & \multirow{1}{*}{\shortstack{Global model}} & & \\

        \hline
        
        \multirow{22}{*}{\shortstack{Hetero-\\Model\\Transfer}} & PEU+LoRA~\cite{xu2023lvnlm} & $\bigcirc$ & & & & LoRA model & Global LoRA model  & DP &  \\
    \cline{2-10}
       ~ & DP-LoRA~\cite{liu2023dplora} & $\bigcirc$ & & & & LoRA model & Global LoRA model & DP &  \\
     \cline{2-10}
	  ~ & \multirow{2}{*}{~\citet{huang2024fast}} & \multirow{2}{*}{$\bigcirc$} & \multirow{2}{*}{$\bigcirc$} & & & \multirow{2}{*}{\shortstack{ LoRA \& embedding\\of P-Tuning v2}} & \multirow{2}{*}{\shortstack{Global LoRA model \&\\
    embedding of P-Tuning v2}} & \multirow{2}{*}{\shortstack{OTP \& \\TEE}} &  \multirow{2}{*}{TEE} \\
      ~ & & & & & & & & &   \\
      \cline{2-10}
	  ~ & FedPETuning~\cite{zhang2023fedpetuning} & $\bigcirc$ & & & & PEFT model & Global PEFT model  & PEFT &  \\
    \cline{2-10}
	  ~ & FedPrompt~\cite{Zhao2023fedprompt} & $\bigcirc$ & & &  & Prompt model & Global prompt model & DP &   \\
       \cline{2-10}
	  ~ & \multirow{2}{*}{\shortstack{FedPPT~\cite{Zhao2023fedprompt}}} &  & & & \multirow{2}{*}{\shortstack{$\bigcirc^\ddagger$}} & \multirow{2}{*}{\shortstack{Backdoored \\prompt model}} & \multirow{2}{*}{\shortstack{Global prompt model}} & & \multirow{2}{*}{\shortstack{}}  \\
         ~ & & & & & & & & &   \\
         
      \cline{2-10}
	  ~ & FedSplitBERT\cite{lit2022fedbertsplit} & & & & & \multirow{1}{*}{\shortstack{Portion of $\mathcal{W}^C$}} & \multirow{1}{*}{\shortstack{Global model}} & &  \\
   
   \cline{2-10}
	  ~ & FedDAT~\cite{chen2023feddat} & & & & & Dual-adapter & \multirow{1}{*}{\shortstack{Global dual-adapter}} & &  \\
      \cline{2-10}
	  ~ & FedIT~\cite{zhang2023fedit} & & & & & LoRA model & Global LoRA model & &  \\
      \cline{2-10}
	  ~ & FedPepTAO~\cite{yi2023fedpepTtao} & & & & & Prompt model & Global prompt model & &   \\

      \cline{2-10}
	  ~ & FedLPFM~\cite{wu2023fedlpfm} & & & & & Proxy model & Global proxy model & &   \\

     \cline{2-10}
        ~ & \multirow{3}{*}{\shortstack{CLIP2FL~\cite{shi2023clip}}} & & & & & \multirow{3}{*}{\shortstack{CLIP-enhanced \\$\mathcal{W}^C$ and gradients\\ of local classifier}} & \multirow{3}{*}{\shortstack{Global model}} & & \\
        ~ & & & & & & & & &   \\ 
        ~ & & & & & & & & &   \\ 
         \cline{2-10}
	  ~ & \multirow{2}{*}{HePCo~\cite{halbe2023hepco}} & & & & & \multirow{2}{*}{\shortstack{ Prompt model \\and classifier}} & \multirow{2}{*}{\shortstack{ Global prompt model \\and classifier}} & &   \\
        ~ & & & & & & & & &   \\
        \cline{2-10}
	  ~ & \multirow{2}{*}{FedET~\cite{liu2023fedet}} & & & & & \multirow{2}{*}{\shortstack{Lightweight \\ module group}} & \multirow{2}{*}{\shortstack{Global lightweight \\module group}} & &   \\
      ~ & & & & & & & & &   \\
      \cline{2-10}
	  ~ & FeS~\cite{cai2023fedfsl} & & & & & Bias model& \multirow{1}{*}{\shortstack{Global bias model}} & &  \\
      \cline{2-10}
	  ~ & FedAdapter~\cite{cai2023fedadapter} & & & & & Adapter & Global adapter & &  \\
      \cline{2-10}
	  ~ & FwdLLM~\cite{xu2023fwdllm} & & & & & LoRA model & Global LoRA model & &   \\
      \cline{2-10}
	  ~ & \multirow{2}{*}{HLoRA~\cite{cho2023hlora}} & & & & & \multirow{2}{*}{\shortstack{Heterogeneous \\ LoRA model}} & \multirow{2}{*}{Global LoRA model} & &   \\
      ~ & & & & & & & & &   \\

    \hline
	\end{tabular}
\egroup
\label{table:optim-ds-fm}
\end{table*}


Optimizing clients' domain-specific FMs through homo-model transfer corresponds to the objective formulated in Eq.(\ref{eq:setting_3_variant_0}) with $K>1$ and $\sum^K_{k=1}p^{C_k}=1$. The specific form of the task loss $\ell^{C_k}(\mathcal{W}^G_{\text{FM}};d)$ for fine-tuning the global model $\mathcal{W}^G_{\text{FM}}$ shared by all clients is formulated in Eq.(\ref{eq:fedft}). Current research of this category primarily revolves around benchmark design and privacy-preserving mechanisms. We will delve into the discussion of privacy-preserving mechanisms in Section \ref{sec:setting_2_privacy} and examine the works related to benchmark design as follows.

FedNLP~\cite{lin2022fednlp} provided a benchmarking framework for evaluating well-known federated learning algorithms (e.g., FedAvg, FedProx, and FedOPT) on FMs (e.g., BART) for common NLP tasks. \citet{woi2023fededge} conducted an in-depth study on hardware performance optimization for federated fine-tuning large language models on the edge (FedLLM-on-Edge). More specifically, It provided a micro-level hardware benchmark for measuring energy, computational, and communication efficiency when federated fine-tuning the FLAN-T5 model family, ranging from 80M to 3B parameters. 

\begin{figure*}[th!]
    \centering
    \includegraphics[width=0.99\linewidth]{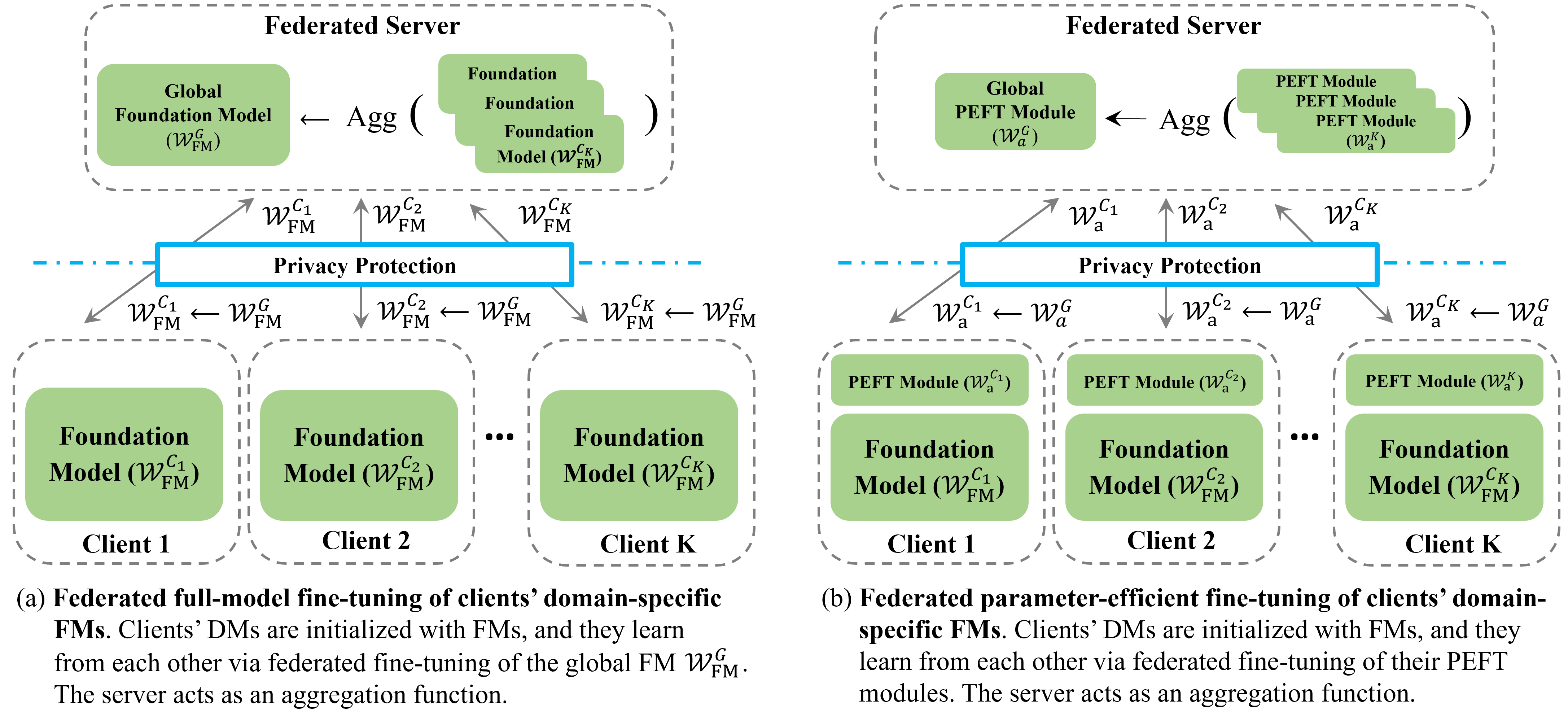}
    \caption{Illustration of optimizing clients' domain-specific FMs. In this scenario, clients' DMs are initialized with FMs, while the server hosts no FM but acts as a model aggregation function. This scenario involves: (a) federated full-model fine-tuning of clients' domain-specific FMs; (b) federated parameter-efficient-finetuning of PEFT modules (e.g., adapter, prompt model, and LoRA model) of clients' domain-specific FMs.}
    \label{fig:fed_peft}
\end{figure*}

One critical driving force behind the adoption of hetero-model transfer is the significant computational and communication costs associated with fine-tuning and sharing clients' full domain-specific FMs. This is mainly due to the large sizes of FMs, as highlighted in the study by ~\citet{cai2023fedadapter}. Current research on hetero-model transfer primarily focuses on the design and development of Parameter-Efficient Fine-Tuning (PEFT) methods, as exemplified in Table \ref{table:optim-ds-fm}.

These PEFT-based FL methods, as illustrated in Figure \ref{fig:fed_peft}, typically involve clients whose DMs are initialized with pre-trained FMs. During training, each client only uploads a PEFT module to the server for aggregation. The PEFT module can be an adapter, LoRA model, or prompt model that contributes to a small portion of the parameters of the original FM, thereby saving computational and communication overheads. These PEFT-based FL methods correspond to the objective formulated in Eq.(\ref{eq:setting_3_variant}) with $K>1$ and $\sum^K_{k=1}p^{C_k}=1$. The specific form of the task loss $\ell^{C_k}(\mathcal{W}^S_{\text{FM}}, \mathcal{W}^{C_k};d)$ for fine-tuning client $k$'s $\mathcal{W}^{C_k}$ with PEFT methods is formulated in Eq.(\ref{eq:fedpeft}).


As for now, few works following Eq.(\ref{eq:setting_3_variant}) explicitly consider protecting data privacy (i.e., $\eta>0$). FedPETuning~\cite{zhang2023fedpetuning}, PEU+LoRA~\cite{xu2023lvnlm}, DP-LoRA~\cite{liu2023dplora}, FedPrompt~\cite{Zhao2023fedprompt}, and the work of \citet{huang2024fast} are representative ones. FedPETuning leverages four PEFT techniques, namely adapter tuning, prefix tuning, LoRA~\cite{hu2021lora}, and BitFit~\cite{ben2022bitfit}, to fine-tune clients' DMs (i.e., $\text{Roberta}_{\text{base}}$) and demonstrated that federated learning combined with LoRA achieved the best privacy-preserving results. DP-LoRA and PEU+LoRA apply LoRA with DP to fine-tune clients' DMs, aiming to achieve better privacy-utility-resource trade-offs than baselines. PEU+LoRA additionally uses Partial Embedding Updates (PEU) to reduce the noise required to protect privacy. \citet{Zhao2023fedprompt} proposed FedPrompt that leverages federated learning combined with soft prompt tuning to fine-tune clients' DMs (e.g., T5~\cite{Raffel2020t5}), aiming to reduce communication cost while maintaining model accuracy. FedPrompt adopts local DP to protect the privacy of clients' data, demonstrating there is a trade-off between privacy and model accuracy. {\blue{\citet{huang2024fast} proposed a method that involves fine-tuning clients' DMs using LoRA and P-Tuning v2. A key consideration in their approach is ensuring the privacy of both clients' data and models. To address this concern, they implemented the fine-tuning process within a Trusted Execution Environment (TEE). Additionally, they employed One Time Pad (OTP) to secure all communicated messages.}}
 
Most methods following Eq.(\ref{eq:setting_3_variant}) focus on improving computation and communication efficiency (i.e., $\eta=0$).~\citet{chen2023feddat} proposed FedDAT, in which each client fine-tunes its local multi-modal model (e.g., VAuLT~\cite{chochlakis2022vault}) by leveraging a Dual-Adapter Teacher (DAT) module to achieve better convergence rate than using conventional PEFT methods. \citet{lit2022fedbertsplit} proposed FedSplitBERT, which splits the BERT encoder into a local part and a global part and shares only a quantized version of the global part with the server for aggregation, aiming to reduce the communication cost with minimal performance loss. FedIT~\cite{zhang2023fedit} was proposed to leverage federated learning combined with LoRA to efficiently utilize clients' diverse instructions stored on local devices to improve the performance of global FM (e.g., Vicuna~\cite{vicuna2023}).
\citet{yi2023fedpepTtao} proposed a parameter-efficient prompt tuning approach with adaptive optimization, named FedPepTAO, to fine-tune clients' large language models (e.g., LLaMA-7B) in a non-IID setting. \citet{halbe2023hepco} proposed HePCo to address a federated class-incremental learning (FCIL) problem. In HePCo, each client is deployed with a ViT model, and each client shares only a prompt model and a classifier (on top of the ViT) with the server to save communication costs. To tackle catastrophic forgetting brought by continual learning and data heterogeneity across clients, the server trains a generator based on current and previous tasks and distills the knowledge from the generator to the global prompt model and classifier at each communication round. {\blue{FedET~\cite{liu2023fedet} also aims to address a FCIL problem. In FedET, each client leverages an enhancer group, a set of lightweight modules, to transfer knowledge to and from the server. Both clients and the server adopt a shared pre-trained Transformer to ensure high precision on various tasks. To address catastrophic forgetting caused by new classes introduced by new tasks and non-i.i.d. class distribution across clients, FedET proposed an enhancer distillation method to modify the imbalance between old and new knowledge and repair the non-i.i.d. problem. ~\citet{wu2023fedlpfm} proposed Fed-LPFM, in which each client uses its local data and a set of private FMs to train a proxy model, and clients' proxy models are shared with and aggregated by the server for federated training. At inference, only the proxy models are used. As a result, the inference latency can be considerably reduced, and the data distribution gap between clients can be mitigated. \citet{shi2023clip} proposed CLIP2FL that leverages an off-the-shelf CLIP model to assist federated learning on heterogeneous and long-tailed data. In the CLIP2FL, the local model of every client consists of a feature extractor and a classifier. During each communication round, each client not only trains its local models using local data but also distills knowledge from a ViT-B/32 model into its local model. The server, on the other side, generates federated features by aggregating classifier gradients and uses these federated features to fine-tune a global classifier for all clients.}}

Besides the aforementioned works, a potentially more demanding and intriguing avenue of research involves empowering edge devices (e.g., smartphones, wearables, and Internet of Things devices) with the knowledge and capabilities of FMs, thereby greatly augmenting the intelligence of systems facilitated by these devices. Take, for instance, the case of smart home systems~\cite{king2023get} enhanced by FMs, which can elevate the quality of life for individuals, or the case of surveillance systems~\cite{yao2019eugene} fortified by FMs, leading to heightened efficiency and accuracy in detecting moving objects such as people and traffic. To achieve these visions, the primary challenge lies in effectively and efficiently adapting FMs to edge devices that have limited memory and computing resources. Federated learning presents a promising solution to this challenge, and several studies have investigated this direction. 

FedAdapter~\cite{cai2023fedadapter} was proposed to improve the computational and communication efficiency of fine-tuning local FMs (e.g., BERT) resided in devices through PEFT (i.e., adapter). To achieve this, it dynamically adjusts the depth and width of adapters during training. FeS~\cite{cai2023fedfsl} was proposed to address the problem of data scarcity in mobile NLP applications by utilizing pseudo-labeling. It employs a curriculum pacing mechanism to gradually speed up the pseudo-labeling speed, a representational filtering mechanism to pseudo-label samples that are most beneficial to learn from, and a co-planing mechanism to control the layer depth and capacity to be trained. Furthermore, FeS fine-tunes only bias parameters of FMs (e.g., RoBERTa-large) while keeping the other parameters frozen to improve training efficiency further. FwdLLM~\cite{xu2023fwdllm} aims to enhance the efficiency of PEFT techniques. It observed that PEFT techniques mainly alleviate network bottlenecks but do not significantly improve training efficiency as they require backpropagation across the entire model to obtain gradients. To address this challenge, FwdLLM employs a backward propagation (BP)-free method to estimate gradients during the forward pass, thereby saving substantial memory and training time. To tackle system heterogeneity, \citet{cho2023hlora} proposed a heterogeneous LoRA (HLoRA) approach that can apply different rank LoRA modules to devices with heterogeneous and constrained computing resources by leveraging zero-padding and truncation for the aggregation.

\subsection{Privacy and Backdoor Attacks}\label{sec:setting_2_privacy}

During the federated learning training or fine-tuning process, clients are subject to data reconstruction attacks initiated by the semi-honest or malicious server. ~\citet{gupta2022recovering} and \citet{balunovic2022lamp} proposed FILM and LAMP attacks that enable a semi-honest server to reconstruct a client's input text from the submitted gradients of the client's FM, and they proposed FWD (freezing word embeddings) and DP-SGD, respectively, to thwart the proposed text reconstruction attacks. \citet{fowl2023decepticons} and \citet{chu2023panning} proposed Decepticons and Panning attacks enabling the malicious server to recover a client's input text by sending a maliciously modified model to clients to capture private information. The two studies suggested leveraging DP-SGD to defend against the proposed malicious attacks. FLTrojan~\cite{rashid2023fltrojan} explored scenarios in which a malicious client can leak the privacy-sensitive data of some other clients in FL. Specifically, a malicious client first identifies the training rounds in which the targeted victim is involved and subsequently maximizes the victim model’s memorization of privacy-sensitive data by manipulating selective weights that are responsible for such memorization.

To study efficient privacy protection mechanisms against privacy attacks mounted by the server, FedHE~\cite{jin2023fedhe} proposed an efficient Homomorphic encryption-based approach for FL that selectively encrypts the most privacy-sensitive model parameters to reduce both computation and communication overheads while providing customizable privacy preservation. FL4ASR~\cite{azam2023fl4asr} investigated the crucial optimization factors, including optimizers, training from a seed model pre-trained centrally or starting from scratch, cohort size, and data heterogeneity, in the context of FL with differential privacy applied to large automatic speech recognition models.

On the other hand, the server may suffer from backdoor attacks mounted by malicious clients. ~\citet{zhang2022neurotoxin} proposed Neurotoxin, a backdoor attack that is performed by injecting backdoors into coordinates that benign clients do not frequently update. ~\citet{yoo2022backdoor} proposed a model poisoning attack conducted by injecting triggers into the embedding of rare words and utilizing gradient ensembling to enhance the poisoning capability. Both two studies leveraged DP and Norm-Clipping to defend against the backdoor attack.

\section{Foundation Model Inference }\label{sec:infer}


After the completion of federated transfer learning, the server or the client may expose its fine-tuned foundation model (FM) through an FM service API for the public to make inferences. We refer to the entity that provides the FM service as the FM provider and the entity that consumes the FM service as the user. In this section, we delve into the privacy threats and potential countermeasures of the FM inference phase, during which the user initiates queries to the FM provider that, in turn, sends back the FM's responses. It is important to note that both the FM provider and the user can potentially act as adversaries during this inference process.

\begin{table*}[htbp] 
\center 

\caption{Summary of adversaries, privacy attacks, protection methods of foundation model inference. EIA: Embedding Inversion Attack; AIA: Attribute Inference Attack; LIA: LLM-assisted Inference Attack; PIP: Privacy-Invasive Prompt; SMPC: Secure Multi-Party Computation; DP: Differential Privacy.}

\footnotesize
\setlength{\tabcolsep}{3.0pt}
\begin{tabular}{ c | c | c | c } 
\hline
   Adversary & Attacking Target & Attack Method & Protection Method \\ 
   \hline \hline
    \multirow{4}{*}{\shortstack{FM Provider}}  & \multirow{4}{*}{\shortstack{Private information conveyed \\ in the prompts of the user}} &  \multirow{4}{*}{\shortstack{EIA \cite{qu2021natural,Song2020infoleak,morris2023text,kugler2021invbert}\\AIA \cite{plant2021cape,li-etal-2022-dont, song2019overlearning, hayet2022invernet,yue2021differential}\\LIA \cite{tong2023infer,staab2023beyond}}} & \multirow{4}{*}{\shortstack{SMPC~\cite{li2023mpcformer, dong2023puma,liu2023llms,zheng2023primer,hou2023ciphergpt, hao2022iron, ding2023east}\\DP-based perturbation~\cite{tong2023infer}\\DP-based ensemble~\cite{hong2023dpopt}\\Anonymization~\cite{staab2023beyond}}} \\
   & &  &  \\
    & &  &  \\
   & &  &  \\
   \hline
   \multirow{3}{*}{\shortstack{FM User}} & \multirow{3}{*}{\shortstack{Private data used to \\ train or augment the FM}} & \multirow{3}{*}{\shortstack{PIP \\\cite{panda2023dpicl,staab2023beyond,carlini2021extracting}}} & DP-based ensemble~\cite{panda2023dpicl} \\ 
   & &   & Filtering~\cite{perez2022red} \\ 
   & &   & Alignment~\cite{staab2023beyond} \\ 
   \hline
\end{tabular}
\label{tab_threat_model}
\end{table*}

\begin{figure*}[h!]
    \centering
    \includegraphics[width=0.98\linewidth]{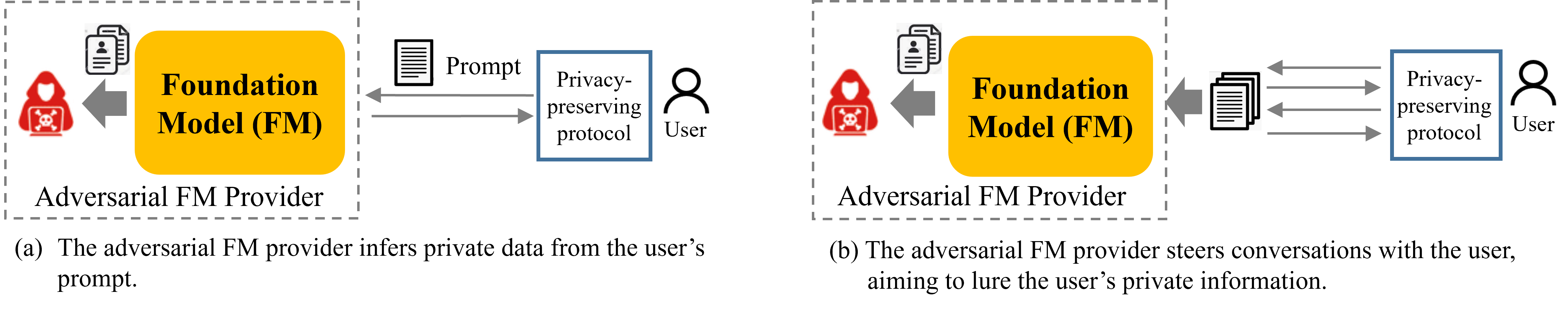}
    \caption{The foundation model inference setting, in which the FM provider is the adversary who aims to infer the user's private data from the prompts submitted by the user.}
    \label{fig:fm_infer_server_attacker}
\end{figure*}

When FM providers are adversaries, they may attempt to infer the private information of the user by investigating the prompts sent by the user with the assistance of the FM. The adversarial FM provider can infer privacy information from the prompt of a single query (illustrated in Figure \ref{fig:fm_infer_server_attacker}(a))~\cite{tong2023infer}, or it can steer a seemingly benign conversation with the user to lure the user into sending out prompts that can be exploited to learn private and sensitive information (illustrated in Figure \ref{fig:fm_infer_server_attacker}(b))~\cite{staab2023beyond}. Various attacking methods have been proposed in these scenarios. For example, FM providers can carry out embedding inversion attacks to reconstruct the original input text based on the provided embedding~\cite{qu2021natural,Song2020infoleak,morris2023text,kugler2021invbert}. They can also employ attribute inference attacks to deduce specific attributes such as race, gender, and age from the given embedding or text~\cite{plant2021cape,li-etal-2022-dont, song2019overlearning, hayet2022invernet,yue2021differential}. The FM provider can also leverage its LLM to recover the original input from its perturbed version~\cite{tong2023infer, staab2023beyond}.

In literature, most of the privacy-preserving methods that aim to protect the user's data privacy during inference focus on designing efficient Secure Multi-Party Computation (SMPC) methods~\cite{li2023mpcformer, dong2023puma,liu2023llms,zheng2023primer,hou2023ciphergpt, hao2022iron, ding2023east}. Given that SMPC comes at a high cost in terms of computational resources and communication, these methods primarily focus on optimizing the efficiency of FM architectures, predominantly Transformer-based neural networks, or SMPC protocols. We will comprehensively review these SMPC-based methods in Section \ref{sec:privacy}. In a recent study, \citet{tong2023infer} proposed InferDPT, which leverages DP-guaranteed text perturbation on the user's prompts to prevent the adversary FM provider from inferring the user's private data. Specifically, in the InferDPT framework, the user first randomly perturbs tokens of its prompt with semantically similar alternatives. Subsequently, the user sends the perturbed prompt to an LLM (i.e., GPT-4), which in turn generates a corresponding response and sends the response back to the user. Finally, the user employs a local pre-trained language model, such as Vicuna-7B, to generate the final response based on the original prompt and the response received from the LLM. DP-OPT~\cite{hong2023dpopt} adopts Deep Language Network (DLN~\cite{sordoni2023dln}) guided by a local FM to optimize prompts automatically. During the optimization, DP-OPT leverages a differentially private ensemble method to generate prompts to protect prompts' privacy. ~\citet{staab2023beyond} investigated leveraging text anonymization to protect the user's data privacy, and it empirically demonstrated that anonymization is insufficient for protecting the user's data privacy. 

\begin{figure*}[h!]
    \centering
\includegraphics[width=0.99\linewidth]{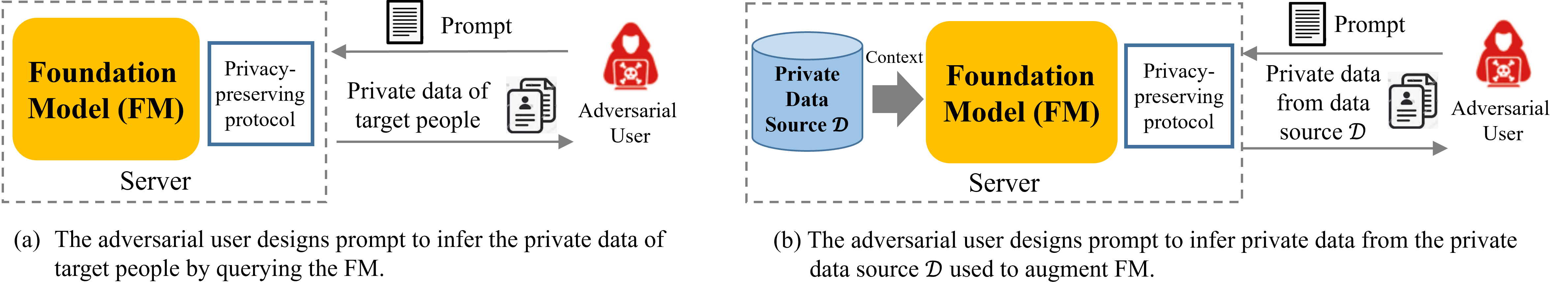}
    \caption{The foundation model inference setting, in which the user is the adversary who aims to design prompts and leverage these prompts to infer the private information of other people.}
    \label{fig:fm_infer_client_attacker}
\end{figure*}

When users are adversaries, their goal is to extract the private information of some targeted people from the FM(illustrated in Figure \ref{fig:fm_infer_client_attacker}(a))~\cite{carlini2021extracting, staab2023beyond} or from the private data sources used to augment the FM(illustrated in Figure \ref{fig:fm_infer_client_attacker}(b))~\cite{panda2023dpicl} through designing privacy-invasive prompts.

For defending against privacy-invasive prompts, \citet{panda2023dpicl} proposed a DP-ICL approach that allows the LLM to generate differentially private responses through a noisy consensus among an ensemble of responses. ~\citet{perez2022red} proposed to remove sensitive text from the response of the FM before it is sent to users, aiming to prevent the FM's response from leaking private information. \citet{staab2023beyond} suggested that model alignment can be used to thwart privacy-invasive prompting. Nonetheless, they also demonstrated that current LLMs are not aligned against privacy-invasive prompts because much of the current alignment research focuses primarily on preventing the LLM from generating harmful and offensive content. Thus, leveraging alignment to protect the FM provider's privacy during inference presents a promising avenue for further research.


\section{Improving Efficiency}\label{sec:efficiency}

Foundation models have become the fundamental infrastructure that drives AI applications across various industries. Nevertheless, adapting large-scale FMs to domain-specific AI applications incurs prohibitive costs. To address this challenge, a series of research works have been focusing on designing and developing efficient fine-tuning and knowledge transfer methods to adapt FMs to downstream tasks or augment FMs with domain-specific knowledge. In this section, we review parameter-efficient fine-tuning and efficient knowledge transfer methods, which play an important role in grounding FMs.

\subsection{Parameter-Efficient Fine-Tuning}


The straightforward way to adapt general-purpose foundation models (FMs) to downstream tasks is to fine-tune all the model parameters (full fine-tuning). However, This is prohibitively expensive for many domain-specific tasks. To mitigate this issue, Parameter-Efficient Fine-Tuning (PEFT) methods \cite{houlsby2019parameter}\cite{he2021towards} are proposed to efficiently adapt FMs to specific domains or tasks. To this end, PEFT methods fine-tune only a small amount of model parameters while freezing the rest model parameters.

Various PEFT methods have been proposed. Specifically, adapter tuning~\cite{houlsby2019parameter} inserts small modules called adapters to each layer of an FM and only adapters are trained during the fine-tuning stage. Inspired by the success of prompting methods that control FM through textual prompts, prompt tuning~\cite{lester2021power} and prefix tuning~\cite{li2021prefix} prepend additional $k$ tunable prefix tokens to the input or hidden layers and only train these tunable tokens on domain-specific tasks. LoRA~\cite{hu2021lora} freezes the FM weights and injects trainable low-rank decomposition matrices into each layer of the Transformer. In general, PEFT has several advantages over full fine-tuning, including:

\begin{itemize}
    \item Improved efficiency: PEFT, by only updating a small number of parameters, can significantly reduce the training time and computational resources required to achieve the same model performance as full fine-tuning.

    \item Better generalization: PEFT can generalize better to new tasks or domains because PEFT retains pre-trained knowledge and prevents the fine-tuning process from drastically changing the model's initial representations. 

    \item Better performance: PEFT can often achieve similar or even better performance than full model fine-tuning because PEFT helps avoid overfitting.

\end{itemize}

While PEFT techniques can alleviate network bottlenecks to some extent, they fail to substantially enhance training efficiency as they still necessitate backpropagation across the entire model to acquire gradients. Recognizing this limitation, recent endeavors have emerged to tackle this predicament. For instance, \citet{xu2023fwdllm} proposed a groundbreaking approach that eliminates the need for backward propagation, thereby saving both time and space for training. Similarly, \citet{sung2022lst} introduced a ladder side-tuning method that circumvents the requirement of backpropagation throughout the entire model by confining it solely to the side network and ladder.
 

\subsection{Efficient Knowledge Transfer}

Although PEFT significantly boosts the computational cost of training FMs, it typically requires a vast amount of storage to deploy FMs with a colossal parameter size, which prevents FMs from being deployed in resource-constrained environments, such as mobile phones. Knowledge transfer methods such as distillation and pruning have emerged as promising solutions to alleviate this challenge. However, traditional knowledge transfer techniques~\cite{gou2021knowledge}  alone may not be sufficient to accurately satisfy the flexible requirements of downstream clients in the era of FMs. For example, downstream clients are likely equipped with different computing powers and storage environments, and thus, they can afford quite different resources to transfer the knowledge from the upstream FM server. Advanced knowledge distillation has become an intensive research area to alleviate the computation and storage burdens of FM clients.

\begin{figure*}[!ht]
    \centering
    \includegraphics[width=0.88\linewidth]{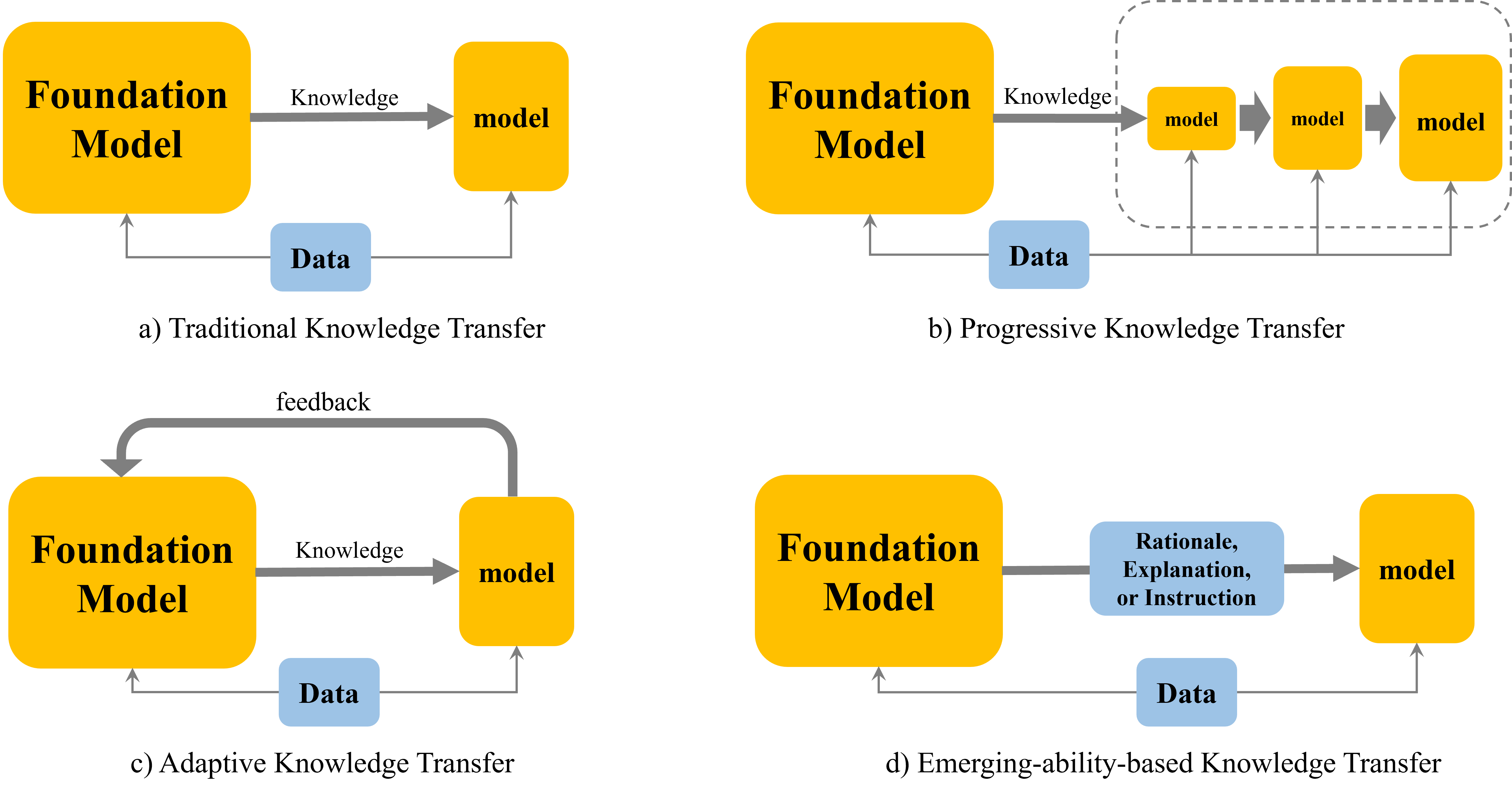}
    \caption{The high-level overview of different categories of knowledge transfer methods.}
    \label{fig:comp_distill}
\end{figure*}

In this section, we review these advanced knowledge transfer methods, and we categorize them into \textit{progressive knowledge transfer}, \textit{adaptive knowledge transfer}, and \textit{emergent-ability-based knowledge transfer} based on how they transfer the knowledge from the teacher model to the student model. 


\subsubsection{Progressive Knowledge Transfer} Progressive Knowledge Transfer (PKT) methods transfer the knowledge of a teacher model (e.g., an FM) into a student model (e.g., a DM) in an iterative manner. More specifically, Instead of transferring all the knowledge from a teacher model to a student model at one training stage, PKT breaks the transferring procedure down into multiple stages. Each stage involves training a student model to mimic the behavior of the teacher model at a specific level of model complexity, as illustrated in Figure \ref{fig:comp_distill}(b). The advantage of PKT is that it allows for more efficient and effective knowledge transfer from the server's teacher model to the downstream client's student model while satisfying the flexible resource constraints of clients compared to traditional knowledge distillation methods.

HomoDistil~\cite{liang2023homodistil} performs distillation with iterative pruning on $\text{BERT}_{\text{base}}$. At each training iteration, HomoDistil prunes a set of least important parameters and simultaneously distills the pruned student, aiming to maintain a small discrepancy between the pruned student and the teacher model. HomoDistil repeats such a procedure in each iteration to maintain the small discrepancy through training, which encourages an effective knowledge transfer. BERT-of-Theseus~\cite{xu2020bert} divides the original $\text{BERT}_{\text{base}}$ into several modules and aims to build their compact substitutes. To this end, BERT-of-Theseus randomly replaces the original modules with their substitutes with a probability $p$ and then fine-tunes the compact modules to mimic the behavior of the original modules. BERT-of-Theseus progressively increases the probability $p$ of replacement during training, which encourages transferring knowledge steadily from the original model to a smaller model. B-DISTIL~\cite{dennis2023bdistil} aims to distill a large pre-trained model onto an ensemble of smaller and low-latency models. The resulting ensemble of smaller models forms a decomposition of the original model such that more accurate prediction can be obtained with more models in the ensemble but at the cost of more training and inference time. A major advantage of B-DISTIL is that B-DISTIL can control the trade-offs between computation cost and accuracy according to specific demands of downstream clients.

\subsubsection{Adaptive Knowledge Transfer} Adaptive knowledge transfer (AKT) methods transfer knowledge from a teacher to a student through an adaptation procedure in which the transferred knowledge is selected based on the student's data and task, as illustrated in Figure \ref{fig:comp_distill}(c). This adaptability enables the student model to capture the teacher model's knowledge better and perform knowledge transfer more efficiently.

~\citet{yu2023selective,hou2023FreD, wang2023can} proposed methods to pre-train, fine-tune, and distill an FM using data that are adapted to downstream clients' private data. More specifically,  ~\citet{hou2023FreD} and \citet{wang2023can} proposed to obtain DMs fine-tuned and distilled, respectively, from an FM based on data that are sampled close to FL clients' private data by exploiting privacy-preserving distribution matching algorithms. ~\citet{yu2023selective} proposed a selective pre-training approach to pre-train an FM using data selected by a domain classifier trained on the downstream client's private data. The three methods aim to obtain pre-trained DMs that can be better adapted to the client's domain-specific tasks. In addition to data adaptation, some other methods guide the teacher model to transfer tailored knowledge that can better improve the student model's performance. ~\citet{wang2022improved} proposed a knowledge selection module (KSM) to transfer the knowledge of appropriate types from the teacher model to the student model when performing knowledge distillation, aiming to improve the performance of the student model more efficiently than knowledge distillation baselines. ~\citet{jiang2023lion} proposed Lion that leverages advanced LLMs, such as ChatGPT~\cite{openai2023gpt4} to identify challenging instructions where the performance of the student model (e.g., LLaMA~\cite{touvron2023llama}) falls short during the process of knowledge distillation, aiming to boost the student model’s proficiency.

\subsubsection{LLM-generated Knowledge Transfer} LLM-generated Knowledge Transfer (LKT) typically fine-tunes a pre-trained student model with input-response pairs in which the responses are augmented by an LLM, which serves as the teacher model. 

Current research has explored utilizing various emergent ability knowledge (e.g., reasoning, rationale, and instruction) to optimize the student model. Approaches such as Fine-tune-CoT~\cite{ho2023large}, CoT Prompting~\cite{magister2023teaching}, and MT-COT~\cite{li2022mtcot} employ reasoning explanations derived from LLM to refine the student model. Distilling Step-by-Step~\cite{hsieh2023distilling} and Sci-CoT~\cite{ma2023sci} utilize rationales generated by LLM to fine-tune the student model (e.g., T5). PaD~\cite{zhu2023pad} leverages reasoning programs to optimize the student model. We refer the reader to Section \ref{sec:llm_kd}, where we review these works from the perspective of federated transfer learning.

\section{Preserving Privacy}\label{sec:privacy}

During federated transfer learning, participating parties transfer the knowledge of their private data and models from each other, aiming to ground FMs. Adversary parties in the loop can investigate transferred knowledge to infer other parties' private information. Therefore, we need ways to protect transferred knowledge to mitigate the chance that adversaries can infer private information. We summarize data privacy protection methods adopted in FTL-FM works in Table \ref{tab:privacy_protection}.

\begin{table*}[ht!]
\centering
\footnotesize
\caption{Summary of data privacy protection methods used in FTL-FM works.}
\begin{tabular}{m{1.8cm}<{\centering}|m{3.0cm}<{\centering}||m{5.5cm}<{\centering}|m{2.0cm}}
\hline

\multirow{2}{*}{\shortstack{Category}} & \multirow{2}{*}{\shortstack{Specific Method}}  &
\multirow{2}{*}{\shortstack{Description}}  &
\multirow{2}{*}{\shortstack{Representative  \\ Work}}  \\
~ & ~ & ~ & ~ \\
\hline
\hline
\hline

\multirow{6}*{\shortstack{Differential \\ Privacy (DP) }} & \shortstack{DP-SGD}  & DP-SGD aims to train private deep learning models and utilizes privacy amplification to obtain a strong DP guarantee &  \cite{abadi2016dpsgd,yu2022differentially,li2022ghostclip,li2023privacy,shi2022jft,bu2022dptitfif,hou2023FreD}  \\
\cline{2-4}
   & DP-FTRL  &  DP variant of Follow-The-Regularized-Leader &  \cite{kairouz2021dpftrl,wang2023can,xu2023fedgboard}   \\
\cline{2-4}
 & Selective DP (SDP) & SDP protects only the sensitive tokens
defined by a policy function  &  \cite{shi2022sdp,shi2022jft} \\
\cline{2-4}
   & DP-based text privatization &  Protect the privacy of textual data based on DP &   \cite{Weggenmann2022dpvae,li2023privacy,zhou2023textobfuscator,timour2022dprewrite,Igamberdiev2023dpbart} \\
\hline

\multirow{7}*{\shortstack{Cryptographic \\ Protection}} & \multirow{1}*{\shortstack{Secure Aggregation (SA)}} & SA is applied in FL to present the server from accessing model information of individual client & \cite{fedost} \\
\cline{2-4}
\multicolumn{1}{c|}{~} & Secure Multi-Party Computation (SMPC) & Use SMPC to encrypt information transmitted between FL parties to protect data privacy & \cite{li2023mpcformer, dong2023puma}\\
\cline{2-4}
\multicolumn{1}{c|}{~} & Homomorphic Encryption (HE) & Use HE to encrypt information transmitted between FL parties to protect data privacy & \cite{chen2022thex, jin2023fedhe} \\
\cline{2-4}
\multicolumn{1}{c|}{~} & Hybrid & Hybrid of cryptographic protection methods &  \cite{liu2023llms,zheng2023primer,hou2023ciphergpt, hao2022iron, ding2023east}\\
\hline
\end{tabular}
\label{tab:privacy_protection}
\end{table*}

\textit{Differential Privacy} (DP)~\cite{dwork2014algorithmic} has become a widely used privacy protection mechanism because of its mathematical properties that facilitate the combination of multiple differential privacy mechanisms and the accumulation of privacy budgets. DP has several variants, including DP-SGD~\cite{abadi2016dpsgd}, DP-FTRL~\cite{kairouz2021dpftrl}, Selective-DP (SDP)~\cite{shi2022sdp}, and DP-Rewrite~\cite{timour2022dprewrite}, that have been adopted to protect data privacy in FTL-FM works. DP-SGD (differentially private stochastic gradient descent) was proposed to train private deep learning models. It leverages privacy amplification by sampling or shuffling to achieve a strong DP guarantee. DP-SGD is not practical to be applied in a cross-device FL system because it requires the system to sample clients uniformly at random on each communication round to provide a formal DP guarantee. To address this obstacle, ~\citet{kairouz2021dpftrl} proposed DP-FTRL (differentially private follow-the-regularized-leader) that can achieve privacy-utility-computation trade-offs comparable with DP-SGD but does not depend on
privacy amplification. SDP (Selective DP) aims to improve the privacy-utility trade-off of DP-SGD when applied to natural language applications. It differs from DP-SGD in that SDP considers only partial dimensions of a training sample as sensitive, thereby saving privacy budgets to achieve the same model performance. DP-SGD, DP-FTRL, and SDP are typically applied to make neural network models DP-guaranteed, whereas DP-rewrite aims to make input text (e.g., textual documents) with DP guarantees by perturbing the original text representations. DP-rewrite techniques are convenient for protecting data-level knowledge transferred between parties in FTL-FM. While DP is a convenient tool for protecting data privacy, it often leads to noticeable utility deterioration, which weakens the motivation for adopting DP in utility-critical applications. 

\textit{Cryptographic protection} is another active research area for protecting the privacy of private data. However, they are mainly used in the inference phase to protect both the model parameters and inference data. \citet{chen2022thex} proposed THE-X that utilizes Homomorphic Encryption (HE) to enable BERT inference on encrypted data. THE-X leveraged approximation methods such as polynomials and
linear neural networks to replace nonlinear operations of BERT with addition and multiplication operations that are compatible with HE. THE-X has the risk of leaking data privacy because it exposes intermediate computing results, and it suffers from performance degradation caused by model structure changes. To address these issues, several works have explored Secure Multi-Party Computation (SMPC) technologies to achieve privacy-preserving Transformer inference. \citet{li2023mpcformer} proposed MPCFormer that replaces bottleneck functions in BERT models with MPC-friendly approximation while leveraging knowledge distillation to maintain model utility. \citet{dong2023puma} proposed the PUMA framework that designed high-quality approximations for expensive functions, such as GeLU and Softmax, which significantly reduce the cost of secure inference while preserving the model performance. In addition, PUMA designed secure Embedding and LayerNorm procedures that preserve desired functionality and the Transformer architecture. 
To improve the efficiency of SMPC operators tailored for FMs' non-linear operations, a line of research works~\cite{liu2023llms,zheng2023primer,hou2023ciphergpt, hao2022iron, ding2023east} proposed hybrid schemes combining HE and SMPC to accelerate linear and non-linear operations of FMs.




\section{Future Directions and Opportunities}\label{sec:future}

Having systematically investigated existing FTL-FM works, it is imperative to recognize that, despite the notable efforts and progress that have been made, there remain many open problems. These problems encompass a multitude of pivotal aspects involved in establishing a \textit{trustworthy} FTL-FM system. Herein, we provide a summary of these issues.

\begin{itemize}
    \item \textbf{Utility on FTL-FM.} Efficiently harnessing federated transfer learning to optimize the utility of foundation models (FMs) and domain models (DMs) is a prominent research and development challenge in the field of FTL-FM. At present, a substantial amount of research in FTL-FM has concentrated on adapting FM knowledge to domain-specific models in scenarios involving a single server and client. Therefore, it is necessary to explore effective knowledge transfer methods specifically tailored to other scenarios, such as the co-evolution of FMs and DMs. Furthermore, the existing FTL-FM approaches primarily address relatively straightforward tasks like classification. Hence, there is a need for the development of algorithms that can facilitate the effective transfer of intricate knowledge and capabilities possessed by FMs to tackle more complex tasks, such as reasoning and chain-of-thought, that clients may require. A seldom-explored avenue in FTL-FM research is the transfer and integration of knowledge across different modalities, such as image, speech, text, and sensor data. The development of innovative algorithms capable of real-time knowledge transfer or fusion from diverse modalities holds tremendous potential in domains like autonomous vehicles, healthcare monitoring, smart homes, and industrial automation.
    
    \item \textbf{Privacy on FTL-FM.} Privacy protection on FTL-FM remains an ongoing area of research that necessitates further improvements despite the significant progress made so far. Differential privacy (DP) and its variations have limitations when it comes to handling complex tasks (e.g., reasoning and chain of thought)~\cite{li2023privacysurvey} and may suffer from a considerable loss of utility. In addition, the impact of DP on privacy and utility under various attacks for complex tasks has not been thoroughly investigated yet. Cryptographic protections such as secure multi-party computation (SMPC) and homomorphic encryption (HE) are primarily utilized during the inference phase of FMs and require further advancements to fully support FTL-FM training. Thus, it is crucial to explore more efficient and effective protection mechanisms that can ensure privacy and uphold utility throughout the entire lifecycle of FTL-FM.
    
    \item \textbf{Efficiency on FTL-FM.} FMs typically have a prohibitively large amount of parameters, making them quite challenging to be deployed in resource-constrained clients. To tackle this obstacle, most federated FMs works (FedFMs for short) resort to parameter-efficient fine-tuning (PEFT) techniques such as Adapter and LoRA. However, PEFT techniques only mitigate the network bottleneck but cannot significantly optimize the efficiency of training because they still require backpropagation throughout the entire model to obtain the gradients. Some recent efforts have tried to address this issue. For example, \citet{xu2023fwdllm} proposed a backward propagation-free method to save the time and space of compute gradients, and \citet{sung2022lst} proposed a ladder side-tuning method to backpropagation through the side network and ladder instead of the entire model. Nevertheless, these studies mainly concentrated on comparatively straightforward tasks such as classification, and their efficacy has not yet been examined in intricate tasks such as chain of thought. In addition, although progress has been made~\cite{xu2023fwdllm,fan2023fate,wang2022multimodal,kuang2023federatedscope,yuan2023m4}, methods to effectively and efficiently transfer knowledge between DM and large FMs (e.g., LLaMA, GLM~\cite{du2022glm}, and BLOOM) at large scale have not yet to be thoroughly explored.

    \item \textbf{Fairness on FTL-FM.} 
    Ensuring fairness within the realm of FTL-FM is paramount as it addresses the complexities of learning from decentralized, often heterogeneous data sources while maintaining model generalizability. The fairness of foundation models is a dynamic research field that continues to evolve ~\cite{zhang2023chatgpt,baldini2021your,ramesh2023fairness, gallegos2023bias}. It involves developing equitable FMs by creating methodologies that prevent the perpetuation of biases, ensure balanced representation, and facilitate fair outcomes across all participating entities. Some key theoretical research directions focus on ensuring equitable representation and performance across diverse datasets, which are often skewed or biased due to their heterogeneous sources. This entails devising novel algorithms and fairness metrics that can be seamlessly integrated into the FTL-FM framework, enabling FMs to learn effectively from a wide array of data sources while mitigating bias and ensuring that the benefits of AI systems are shared more evenly among all stakeholders involved. It is also crucial to dig deep into understanding fairness when it comes to models that handle different types of data and to figure out the balance between making models straightforward to understand and ensuring they are fair. Theoretical exploration of fairness in multimodal models involves understanding how biases may manifest uniquely in systems that process and integrate multiple forms of data, such as text, images, and audio. 

    \item \textbf{Robustness on FTL-FM.} Based on our comprehensive investigation into the various scenarios explored by current FTL-FM research, it is foreseeable that the modes of interaction and collaboration between FM servers and DM clients can be more diverse compared to traditional federated learning settings. Consequently, adversaries are likely to have increased opportunities to exploit security vulnerabilities within FTL-FM systems and launch malicious attacks, such as backdoors and Byzantine attacks. As for now, the security vulnerabilities of FTL-FM have been overlooked, with little attention given to the development of defense mechanisms. Therefore, it is of utmost importance to explore potential vulnerabilities of FTL-FM and devise robust defensive mechanisms to defend against possible malicious attacks, thereby ensuring the robustness of FTL-FM systems.

    \item  \textbf{Model Ownership on FTL-FM.} The advancement of FMs brings the flourishing of content creation applications in a variety of fields, including text, music, imagery, video, and 3D media. However, this rapid growth has led to conflicts, notably those related to copyright infringement. The central challenge in protecting FM copyrights lies in the protection of models that have been developed through substantial time, energy, and cost investments.  \citet{kirchenbauer2023watermark} proposed a watermarking technique for LLMs, which involves watermarking the text generated by LLMs by manipulating the available token part of the vocabulary during each token's sampling. Following this work, several studies have explored effective text watermarking in various contexts \cite{kirchenbauer2023reliability,lee2023wrote, zhao2023provable}. However, there is a notable absence of robust solutions specifically for large language models. This gap underscores the urgent need for scholarly attention and further research in this area.
    
    \item \textbf{Trade-offs among multiple objectives.} A trustworthy FTL-FM system typically must take into account multiple critical but conflicting objectives (e.g., utility, privacy, efficiency, and robustness). 
    Achieving a balance among these conflicting objectives presents challenges. For instance, implementing noise to uphold differential privacy may negatively impact the quality of the generated text or even result in model incoherence. 
    How to achieve optimal trade-offs among multiple conflicting objectives such as privacy, utility, efficiency, and fairness is a research direction crucial to trustworthy federated learning. Constrained multi-objective federated learning (CMOFL)~\cite{kang2023cmofl} is a promising approach to address the trade-off problem. It simultaneously optimizes multiple conflicting objectives and can find Pareto optimal solutions that can satisfy the flexible requirements of federated learning parties. Integrating CMOFL into FTL-FM to address various trade-off problems is also worthy of exploring.
\end{itemize}
In addition, there are several other research areas, including interpretability, fairness, and standardization, which are also pivotal to ensuring the trustworthiness of FTL-FM and, thus, warrant further investigation. Furthermore, given that the research on FTL-FM is still in its nascent stages, numerous opportunities are worth exploring.
\begin{itemize}
    
    \item \textbf{Continual Learning on FTL-FM.} Continual Learning is a research area that aims to continuously integrate trained models with new knowledge while mitigating catastrophic forgetting (CF) of previously acquired knowledge~\cite{biesialska2020continual}. Continual learning is crucial for FMs, whose knowledge is static and can be quickly outdated. As for now, FTL-FM works mainly focused on integrating FMs with domain-specific knowledge or adapting the knowledge of FMs to domain models. CF problems are rarely explored in FTL-FM. Therefore, how to reduce CF in pre-trained FMs and DMs during knowledge transfer is worthy of research. In addition, co-evolving FMs and DMs often involve continually fine-tuning of FMs and DMs over time. For example, after DMs learn the general knowledge from FMs, DMs may then be fine-tuned using domain-specific data newly captured in real-world applications. The incremental data captured by DMs can help FMs to continue to evolve. The continual learning algorithms, privacy protection mechanisms, and efficiency-improving strategies for facilitating the co-evolution of  FMs and DMs are research directions all worth exploring.
    
    \item \textbf{Machine Unlearning on FTL-FM.} Machine unlearning \cite{bourtoule2021machine,cao2015towards} refers to the ability of a machine learning model to forget or remove the knowledge it has gained from specific data points or a set of data points. This is particularly important in cases where the data used to train the model is found to be incorrect or biased or if it infringes upon user privacy. Potential future directions for the application of machine unlearning in FTL-FM can be bifurcated into two main aspects. First, the server, which maintains the foundational model, could aim to distill specialized knowledge as opposed to comprehensive knowledge to the client. Consequently, an unlearning algorithm could be leveraged to eliminate any superfluous knowledge. Second, in scenarios where a client requests the removal of sensitive information from the foundational model, the unlearning algorithm could facilitate this process, thereby ensuring data privacy and security.
    
    \item \textbf{Edge AI on FTL-FM.} Edge AI research is fervently pursuing the integration of edge devices (e.g., smartphones, wearables, and Internet of Things (IoT) devices) with the power of FMs. This complex endeavor spans several critical research areas highly relevant to FTL-FM: (1) Data privacy and system security: ensuring the privacy of sensitive information and security of edge systems is of utmost importance in edge computing. Therefore, developing secure federated learning protocols and cutting-edge protection techniques is vital for protecting data privacy and system security. (2) Operational efficiency: reducing the power consumption of AI on edge devices is a key challenge. Synergistic designs of hardware, software, and algorithms are required to enable FMs to operate and learn efficiently and effectively within the limitations of edge environments. (3) Scalability: It is crucial to enable distributed learning across networks of heterogeneous devices. This requires the careful allocation of diverse computational and networking resources and the creation of standardized protocols for edge AI system interoperability. (4) Cloud-edge online cooperation: developing systems that enable FMs to continually learn from streaming data is essential for applications that demand online decision-making, such as autonomous vehicles and medical diagnostics. On the other hand, as FMs in the cloud continue to evolve, it is critical to seamlessly adapt knowledge of FMs to edge devices to ensure the cutting-edge performance of edge AI systems.
\end{itemize}
These research trajectories are essential in harnessing the advantages of FMs in federated learning environments, with the ultimate goal of transforming various industries with intelligent, efficient, and agile AI-powered solutions.

\section{Conclusion}\label{sec:conclusion}

{\blue{

This work aims to study the topic of grounding foundation models through federated transfer learning by answering questions we raised in the introduction. Toward this end, we first propose a general FTL-FM framework, which formulates three typical FTL-FM settings and their corresponding objectives. Under this framework, we formulate representative knowledge transfer approaches, including domain adaptation, federated prompt learning, federated split learning, federated offsite tuning, federated knowledge distillation, federated co-optimizing, and federated parameter-efficiency fine-tuning.

Then, we break down these questions into concrete research issues concerning the grounding foundation model in federated learning settings: (1) What to transfer; (2) How to transfer; (3) What to protect; (4) How to protect; (5) How to attack. Based on the five research issues, we construct a taxonomy to categorize state-of-the-art FTL-FM works. Subsequently, we review these works by investigating how they address the five research issues. Additionally, we provide an overview of efficiency-improving and privacy-preserving methods, emphasizing their importance in FTL-FM. Finally, we discuss the open opportunities and future directions for FTL-FM research based on the comprehensive investigation of existing FTL-FM works.

}}


\bibliographystyle{ACM-Reference-Format}
\bibliography{FL, FTLLM, large_model, NIPSFM, future_direction, FTLLLM-Infer}

\end{document}